\documentclass[11pt]{article}

\usepackage[preprint]{acl}

\usepackage{times}
\usepackage{latexsym}

\usepackage[T1]{fontenc}

\usepackage[utf8]{inputenc}

\usepackage{microtype}

\usepackage{inconsolata}

\usepackage{graphicx}

\usepackage[ruled,vlined,linesnumbered]{algorithm2e}
\usepackage{amsmath}
\usepackage{algorithmic}
\usepackage{graphicx}
\usepackage{subcaption}
\usepackage{booktabs}
\usepackage{multirow}
\usepackage{makecell}

\title{Self-Distillation for Multi-Token Prediction}

\author{
 \textbf{Guoliang Zhao\textsuperscript{1}},
 \textbf{Ruobing Xie\textsuperscript{1}}\thanks{corresponding author},
 \textbf{An Wang\textsuperscript{1}},
 \textbf{Shuaipeng Li\textsuperscript{1}},
 \textbf{Huaibing Xie\textsuperscript{1}},
 \textbf{Xingwu Sun\textsuperscript{1}}
\\
\\
 \textsuperscript{1}Large Language Model Department, Tencent
\\
}

\begin{document}
\maketitle

\begin{abstract}
As Large Language Models (LLMs) scale up, inference efficiency becomes a critical bottleneck. Multi-Token Prediction (MTP) could accelerate LLM inference by predicting multiple future tokens in parallel. However, existing MTP approaches still face two challenges: limited acceptance rates of MTP heads, and difficulties in jointly training multiple MTP heads. Therefore, we propose MTP-D, a simple yet effective self-distillation method with minimal additional training cost, which boosts MTP head acceptance rates (+7.5\%) while maximumly preserving main-head performance. We also introduce a looped extension strategy for MTP-D, enabling effective and economical MTP head extension and further significant inference speedup to 1-head MTP (+220.4\%). Moreover, we systematically explore and validate key insights on the distillation strategies and the potential scalability of MTP through extensive experiments on seven benchmarks.  These results demonstrate that our MTP-D and looped extension strategy effectively enhance MTP-head performance and inference efficiency, facilitating the practical usage of MTP in LLMs.
\end{abstract}
\section{Introduction}
\label{introduction}
Large Language Models (LLMs) have demonstrated strong performance across diverse tasks~\cite{guo2025deepseek, team2025kimi}.
As task complexity and model scale increase, inference efficiency becomes increasingly important. 
However, most LLMs rely on the Next-Token Prediction (NTP) paradigm, which performs autoregressive, token-by-token generation and inherently incurs high latency and computational cost, particularly for long sequences~\cite{mehra2025multi}.

Multi-Token Prediction (MTP) has been proposed as an effective approach to alleviate this inefficiency~\cite{gloeckle2024better}. 
It extends the traditional NTP paradigm by training LLMs with multiple heads, enabling parallel prediction of future tokens. 
It has been widely adopted in industrial LLMs to accelerate inference~\cite{xiaomi2025mimo, qwen3technicalreport}.
In general, higher acceptance rates and greater scalability of MTP heads lead to more substantial inference speedups.

The cascaded MTP architecture of DeepSeek-V3~\cite{liu2024deepseek} effectively improves the performance of MTP heads. 
However, it still faces several challenges for successors: (a) the limited acceptance rates of MTP heads, which could lead to the exponential decline of the cumulative acceptance rate, and thus harm the practical effect of inference speedup.
(b) The difficulty in jointly training multiple main and MTP heads. The seesaw effect makes it challenging to jointly improve all heads, while substantial performance decrease of the main head is unaccepted in practice.

To address these issues, we propose \textbf{MTP-D}, a simple and effective self-distillation method, which adopts a gradient-detached, Top$N$-logits-selected distillation from the main head to MTP heads in pre-training, along with a looped extension strategy via economical continued pre-training.
Specifically, we adopt the Top$N$-selected logits of the main head to guide the training of MTP heads as additional losses, which naturally conforms to the original design intention of MTP with minimal harm to the main head and marginal training cost.
Moreover, we propose a looped MTP extension strategy, which takes trained MTP heads as a group, orderly extends them group by group as new MTP heads' initialization, and updates them via continue pre-training.
By leveraging intra-group correlations among MTP heads and the distributional consistency induced by distillation, this strategy efficiently extends MTP heads via less tokens, achieving substantial inference speedups.

Experimental results demonstrate that our MTP-D with 4 heads achieves a 7.5\% increase in MTP head acceptance rates with comparable main-head performance, corresponding to a 22.9\% speedup. 
Moreover, our looped extension strategy enables cost-efficient expansion of MTP heads from 4 up to 16, yielding further 35.1\% speedup. 
In addition, we uncover and validate several interesting insights regarding the scalability of MTP. 
Overall, our key contributions are as follows:
\begin{enumerate}
    \item We propose MTP-D, a novel self-distillation framework for improving MTP head acceptance rates with comparable main-head performance and less additional cost.
    \item We introduce a looped extension strategy, enabling cost-efficient extension of trained MTP heads via continue pre-training.
    \item Extensive experiments demonstrate that our method significantly increases MTP head acceptance rates and speedups, shedding light to the practical usage of MTP.
\end{enumerate}
\section{Preliminary}
\label{preliminary}

Next-Token Prediction is constrained by its training paradigm, which substantially limits both the sample efficiency during pretraining and the inference efficiency of  LLMs. Recent studies show that Multi-Token Prediction alleviates these limitations by employing $N$ independent output heads to simultaneously predict the next $N$ tokens, thereby providing richer supervisory signals and significantly improving sample efficiency \cite{qi2020prophetnet}. Furthermore, when combined with speculative decoding, MTP can dramatically accelerate LLM inference \cite{leviathan2023fast, chen2023accelerating, gloeckle2024better, li2024eagle}. 

DeepSeek-V3~\cite{liu2024deepseek} advances MTP by adopting a sequential, cascaded architecture that explicitly models inter-token dependencies while preserving the complete causal chain and autoregressive nature. This architecture has been widely integrated into prominent industrial LLMs, such as MiMo, GLM, LongCat, and Qwen3-Next~\cite{gloeckle2024better, xiaomi2025mimo, zeng2025glm, team2025longcat, qwen3technicalreport}. Specifically, the loss function of DeepSeek MTP is formulated as follows:
\begin{equation}
\begin{aligned}
\mathcal{L}^{\text{CE}}_{\mathrm{mtp}}
&= \sum_{k=1}^{K} \alpha_k \, \mathcal{L}_{\mathrm{mtp_k}}^{\text{CE}} \\
&= \sum_{k=1}^{K} \alpha_k \,
\mathrm{CE}\!\left(
\hat{\mathbf{P}}_{k+1:T+1}^{k},\,
\mathbf{t}_{k+1:T+1}
\right)
\end{aligned}
\label{eq:mtp_loss}
\end{equation}
where $\mathrm{CE}(\cdot)$ represents the cross-entropy loss, $\alpha_k$ denotes the weight coefficient of the $k$-th MTP $\mathrm{CE}$ loss term, and $\mathbf{t}$ and $\hat{\mathbf{P}}$ denote the ground-truth token sequence and its corresponding predicted probability distribution, respectively. 
\section{Method}
\label{methods}

Multi-Token Prediction has been widely proven effective for accelerating LLM inference, which is primarily predicated on the acceptance rate of the speculative tokens generated by MTP heads. We introduce MTP-D, which effectively aligns the top logit distributions of MTP heads with the main head via distillation in pre-training, further enhanced by a looped strategy for economical MTP head extension via continued pre-training.

\begin{figure*}[h]
    \centering
    \includegraphics[width=0.99\linewidth]{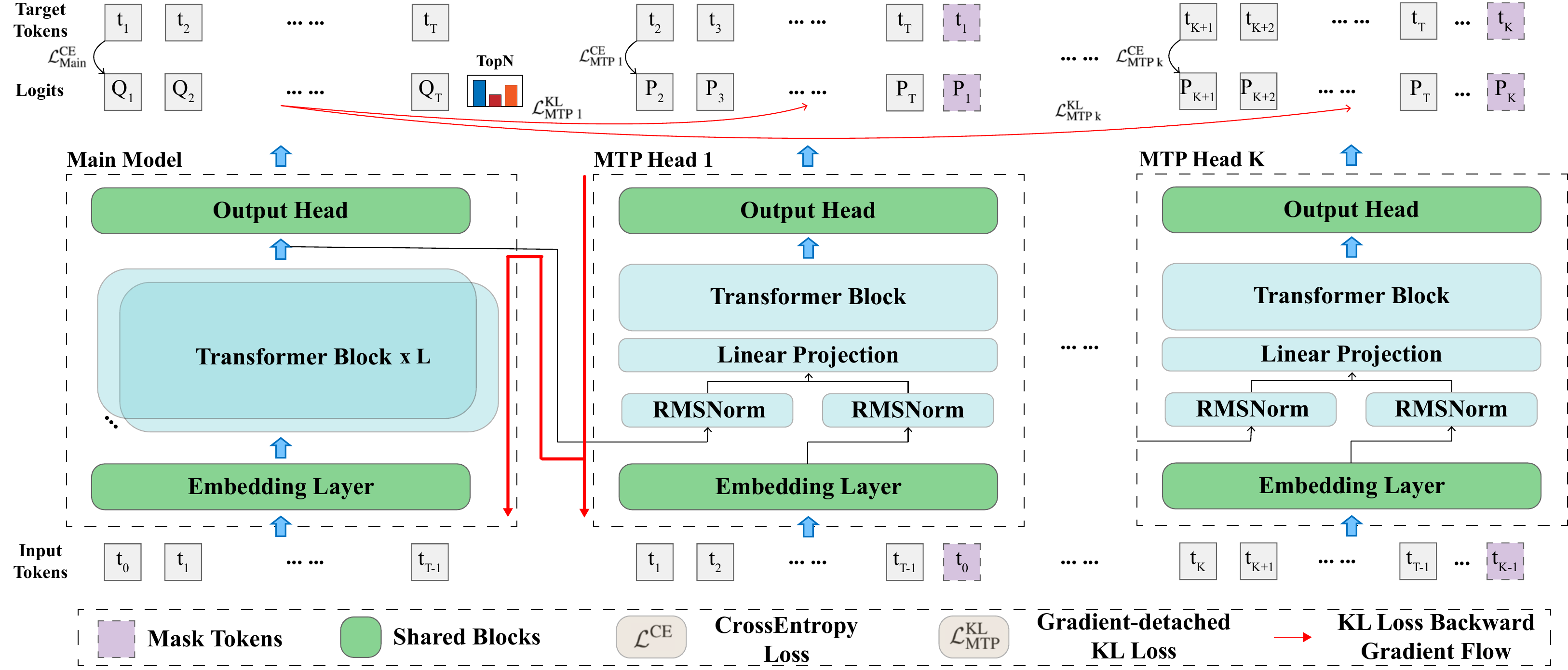}
    \caption{Overview of the gradient-detached, Top$N$-logits-selected self-distillation method.}
    \label{fig:topk_distillation}
\end{figure*}
\subsection{Existing Issues of MTP}
Currently, practical MTP often faces the following challenges: 
(a) \emph{Limited acceptance rates of MTP heads.} A persistent performance gap between MTP heads and the main head restricts acceptance rates and thus limits inference acceleration. As shown in Figure~\ref{fig:illustration_of_loss}, MTP heads incur substantially higher losses during pre-training, with losses increasing as the MTP index grows. The cumulative acceptance rate will rapidly drop to unacceptable value even with moderate single head acceptance rates.
(b) \emph{Difficulty in jointly training multiple MTP heads.} Adding more MTP heads introduces extra loss terms and hyperparameters, which can hinder main-head optimization and complicate large-scale pre-training. Most practical LLMs have fewer than 1-4 MTP heads in training~\cite{xiaomi2025mimo}.

\subsection{Self-Distillation for MTP in Pre-Training}
\label{selfdistillation}
To enhance the MTP heads' performance without compromising the main head's performance, as shown in Figure~\ref{fig:topk_distillation}, we propose a gradient-detached, Top$N$-logits-selected self-distillation method. It introduces an additional distillation supervision signal to align the logit distributions of the MTP heads toward those of the main head, which enforce the consistency of logit distributions between the MTP heads and the main head, thereby significantly imporving the performance of the MTP heads.

Specifically, based on the MTP loss $\mathcal{L}^{\text{CE}}_{\mathrm{mtp}}$ in Eq.~\eqref{eq:mtp_loss}, we introduce a unidirectional Kullback-Leibler (KL) divergence loss \cite{kullback1951information} for self-distillation from the Top$N$ logits of the main head to the corresponding indices of the MTP heads, formulated as follows:
\begin{equation}
\begin{aligned}
\mathcal{L}^{\text{KL}}_{\mathrm{mtp}}
&= \sum_{k=1}^{K} \beta_k \, \mathcal{L}_{\mathrm{mtp_k}}^{\text{KL}} \\
&= \sum_{k=1}^{K} \beta_k \,
\mathrm{KL}\!\left(
\tilde{\mathbf{P}}_{k+1:T+1}^{k},\,
\text{sg}(\tilde{\mathbf{Q}})_{k+1:T+1}
\right)
\end{aligned}
\label{eq:mtp_loss_KL}
\end{equation}
where,
\begin{equation}
\mathcal{I}_{t}^{\text{N}} = \text{TopK}\left(\hat{\mathbf{Q}}_{t}, N\right)
\label{eq:mtp_KL_loss_TopN}
\end{equation}
\begin{equation}
\tilde{\mathbf{Q}}_{k+1:T+1} = \sigma\left(\hat{\mathbf{Q}}_{k+1:T+1}\left[..., \mathcal{I}_{t}^{N}\right]\right)
\label{eq:mtp_KL_loss_Q}
\end{equation}
\begin{equation}
\tilde{\mathbf{P}}_{k+1:T+1}^{k} = \text{log}\left(\sigma\left(\hat{\mathbf{P}}_{k+1:T+1}^{k}\left[..., \mathcal{I}_{t}^{N}\right]\right)\right)
\label{eq:mtp_KL_loss_P}
\end{equation}
Here, $\mathcal{I}_{t}^{N}$ denotes the set of indices corresponding to the Top$N$ elements of the main head logits $\hat{\mathbf{Q}}_{t}$ along the vocabulary dimension $V$ for the $t$-th token. $\sigma(\cdot)$ represents the softmax function, and $\log(\sigma(\cdot))$ corresponds to the log-softmax function. $\mathrm{KL}(\cdot)$ denotes the KL divergence loss, and $\beta_k$ indicates the weighting coefficient of the $k$-th MTP KL loss term. Finally, $\mathrm{sg}(\cdot)$ denotes the stop-gradient operation.
The core of our method lies in gradient-detached self-distillation and Top$N$-selected logits, as described below.

\textbf{Gradient-detached self-distillation.}
In practical settings, we aim to improve the performance of MTP heads while avoiding significant degradation of the main head.
To minimize the influence of self-distillation on the main head, we apply a stop-gradient operation to the main head logits $\hat{\mathbf{Q}}$ in $\mathcal{L}^{\text{KL}}_{\mathrm{mtp}_k}$, effectively preventing gradients from propagating back through $\hat{\mathbf{Q}}$. As illustrated by the red path in Figure~\ref{fig:topk_distillation}, $\mathcal{L}^{\text{KL}}_{\mathrm{mtp}_k}$ propagates gradients exclusively through $\hat{\mathbf{P}}$ during the backward phase, which is identical to the backward path of the MTP cross-entropy loss $\mathcal{L}^{\text{CE}}_{\mathrm{mtp}_k}$. Consequently, with an appropriate configuration of the weighting coefficients $\alpha_k$ for $\mathcal{L}^{\text{CE}}_{\mathrm{mtp}_k}$ and $\beta_k$ for $\mathcal{L}^{\text{KL}}_{\mathrm{mtp}_k}$, our method  could achieve maximal performance of the MTP heads while maintaining comparable performance of the main head.

\textbf{Top$N$-selected logits.}  
Modern LLMs often use extremely large vocabularies (e.g., 122{,}880 in our setting), making full-vocabulary self-distillation loss $\mathcal{L}^{\text{KL}}_{\mathrm{mtp}_k}$ computationally expensive.  
As Figure~\ref{fig:topN_probs} shows, main-head logits follow a long-tailed distribution after softmax, with most token probabilities near zero. Directly distilling all logits leads to redundancy, high memory usage, numerical instability, and weak supervision from low-probability tokens.  
Analyses and ablations indicate that selecting Top$N=10{,}000$ tokens ensures efficient and stable self-distillation. Details are in Appendix~\ref{appendix_logitprob}.

\textbf{Other explorations.}
We further investigated several potential self-distillation strategies for MTP, including: 
(1) different ensemble strategies of the main head and MTP heads as teachers; 
(2) variants of the KL loss function, including forward, reverse, and hybrid KL; 
(3) different Top$N$ logits selection strategies, including the Top$N$ of $\hat{\mathbf{Q}}_{t}$ and the union of the Top$N$ from $\hat{\mathbf{Q}}_{t}$ and $\hat{\mathbf{P}}_{t}$; 
and (4) dynamic adjustment of $\alpha_k$ and $\beta_k$ with training steps. 
Subsequent experiments provide a detailed analysis of the effectiveness of each strategy in Section 4.

\begin{figure*}[!h]
    \centering
    \includegraphics[width=0.8\linewidth]{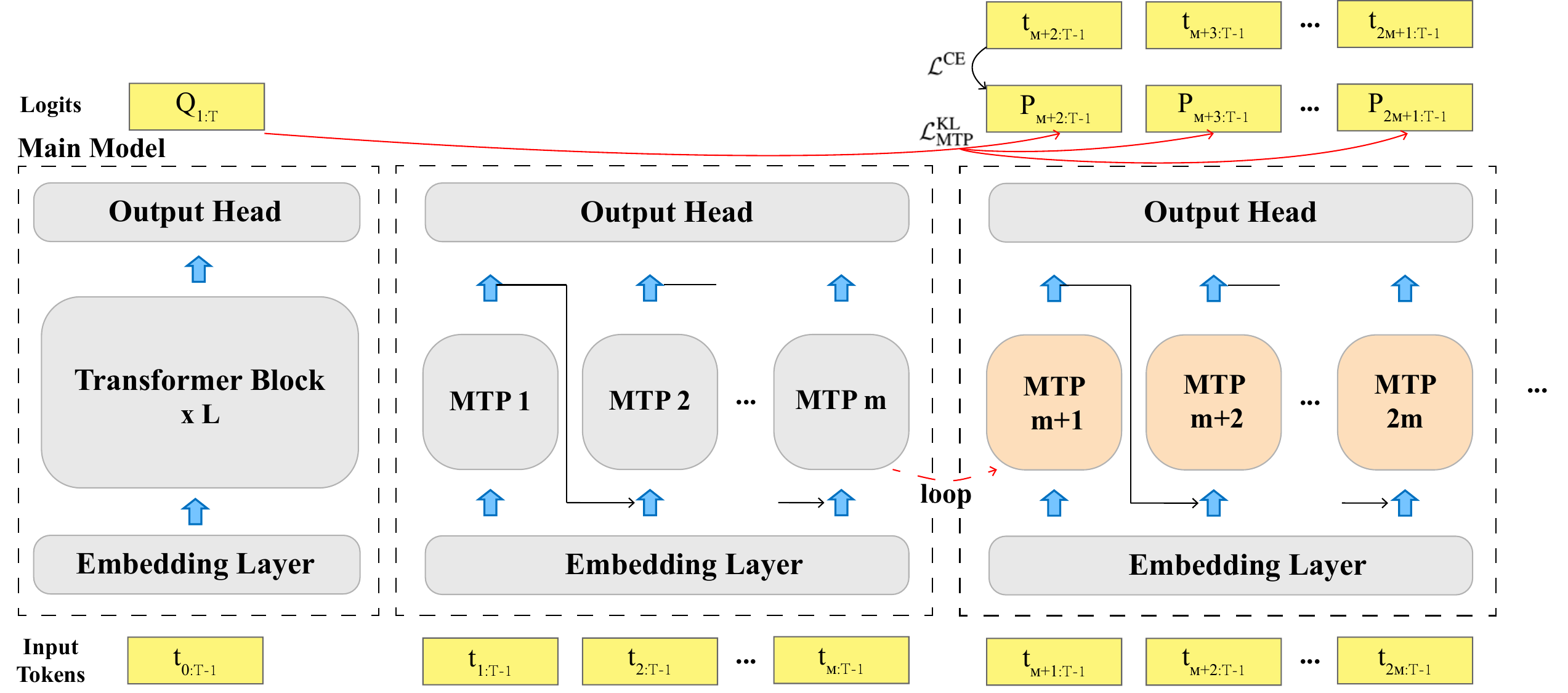}
    \caption{Illustration of the training strategy for looped extension of MTP-D. The gray blocks represent the frozen main model and the trained MTP heads from 1 to $m$. The weights of the MTP heads from 1 to $m$ are copied to initialize the MTP heads from $m{+}1$ to $2m$. The orange blocks denote the trainable MTP heads.}
    \label{fig:loop_model}
\end{figure*}

\textbf{Final strategy.}
The final loss for the MTP heads consists of two components: $\mathcal{L}^{\text{CE}}_{\text{mtp}}$ in Eq. \ref{eq:mtp_loss} and $\mathcal{L}^{\text{KL}}_{\text{mtp}}$ in Eq. \ref{eq:mtp_loss_KL}. $\mathcal{L}^{\text{CE}}_{\text{mtp}}$ aligns the MTP heads with the ground-truth tokens, ensuring the fundamental correctness, particularly during the early stages of pre-training. In contrast, $\mathcal{L}^{\text{KL}}_{\text{mtp}}$ enables the MTP heads to acquire higher-level semantic knowledge from the main head through knowledge distillation, while constraining their probability distributions to be consistent with that of the main head. Together, these two loss terms jointly ensure a stable and effective pre-training: $\mathcal{L}_{\text{mtp}} = \mathcal{L}^{\text{CE}}_{\text{mtp}} + \mathcal{L}^{\text{KL}}_{\text{mtp}}$.

\subsection{Looped MTP Head Extension in Continue Pre-Training}
\label{looped}

Our MTP-D achieves higher acceptance rates via distillation, thereby establishing a solid foundation for scaling up the number of MTP heads. However, this scaling introduces additional losses and hyper-parameters, which may interfere with the optimization of the main head due to conflicts.

From Figure~\ref{fig:topk_distillation}, we observe that the DeepSeek MTP architecture inherently exhibits strong structural consistency and input-output similarity, which provides a natural foundation for scaling up MTP heads via looped extension. As shown in Figure~\ref{fig:loop_model}, our proposed ``\textbf{looped extension}'' denotes an operation in which a previously trained set of $m$ MTP heads is used to initialize the next set of $m$ heads, followed by continued pre-training on the extended MTP heads. Iteratively repeating this operation allows for a gradual expansion of MTP heads.

\textbf{Training-Free Explorations.}
To validate this insight, we conduct a set of pilot experiments under a training-free setting, where both the DeepSeek MTP and our MTP-D are looped up to 8 MTP heads. As illustrated in Figure~\ref{fig:accrate_loop_trainfree}, several key observations can be made as follows:

\begin{figure}[!h]
\begin{subfigure}{0.49\textwidth}
    \centering
    \includegraphics[width=\linewidth]{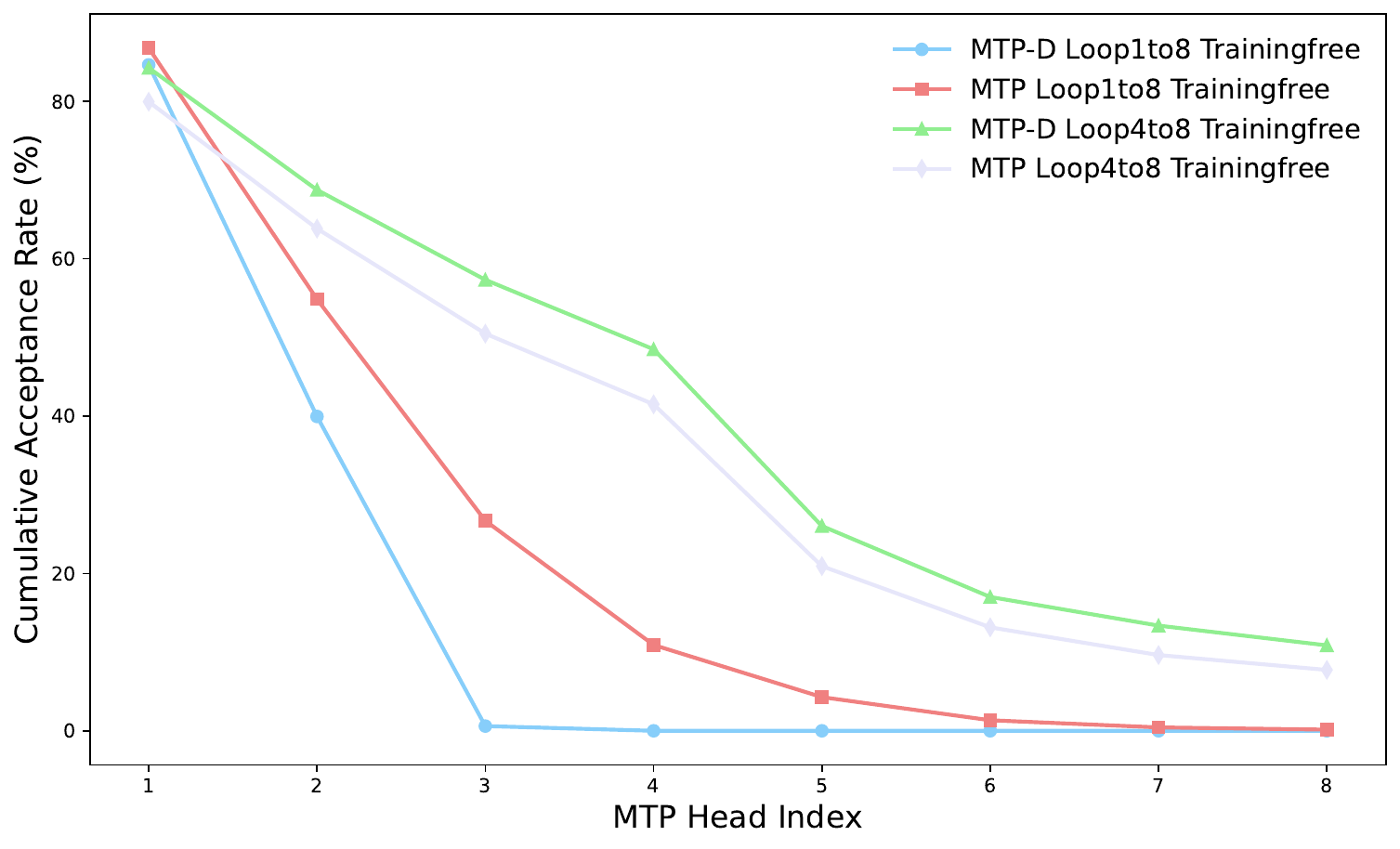}
    \subcaption{Cumulative acceptance rates for AGIEval-en.}
\end{subfigure}
\hfill
\begin{subfigure}{0.49\textwidth}
    \centering
    \includegraphics[width=\linewidth]{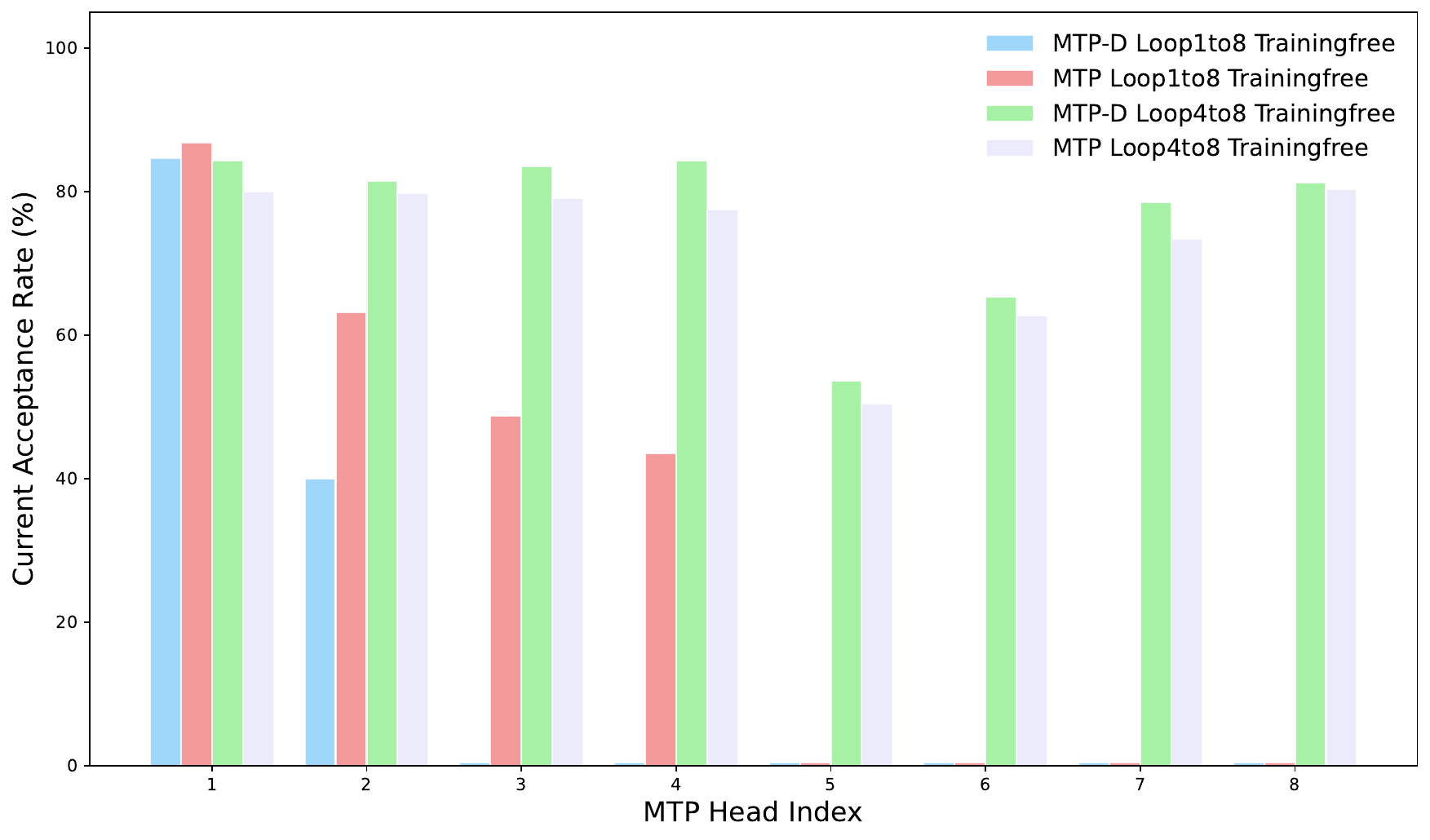}
    \subcaption{Acceptance rates for AGIEval-en.}
\end{subfigure}
\caption{Acceptance rate and cumulative acceptance rate of MTP heads for different models under training-free looped extension. Using the AGIEval-en benchmark as an example, (a) shows the cumulative acceptance rate as the loop is extended to 8 MTP heads for different models, while (b) presents the acceptance rate of each MTP head.}
\label{fig:accrate_loop_trainfree}
\end{figure}

First, MTP heads at loop connection points exhibit a noticeable drop in acceptance rate, though still at an acceptable level. As the number of MTP heads increases, their acceptance rates gradually recover and approach those of the previously trained heads. These observations suggest that the cascaded architecture of DeepSeek MTP, along with its structural consistency and input-output similarity, inherently supports scalability.

Compared to DeepSeek MTP, MTP-D exhibit substantially better scalability. Under the 1-to-8 loop setting, the cumulative acceptance rate of MTP drops to 0.6\% at MTP head~3, whereas MTP-D maintains 26.70\%, enabling further scaling. This improvement stems from enhanced consistency of output distributions between MTP heads and the main head, which significantly boosts scalability.
Detailed analysis of the training-free looped MTP is in Appendix~\ref{appendix_loop_trainfree}.

\textbf{Looped Extension of MTP-D.} The looped extension inherently preserves the correlations among intra-group MTP heads. As shown in Figure~\ref{fig:loop_model}, continued pre-training is performed using the same self-distillation strategy, which largely maintains consistency between the output distributions of all heads. Additionally, to avoid degrading the performance of the main head and to reduce training cost, both the main model and the previously trained MTP heads are kept frozen.

\section{Experiments}
\label{results}
\subsection{Experimental Setup}

\textbf{Model configuration and training details.} 
Our MTP-D was implemented on both 2B Dense and N10BA1B MoE LLMs. All experiments were conducted on the FineWeb-Edu-350BT dataset~\cite{penedo2024fineweb}, where MTP-D pre-training used 350B tokens and looped extension continued pre-training used 70B tokens. Primary experiments were run on 256 NVIDIA H20 GPUs. More details are provided in Appendix~\ref{appendix_configdetails}.

\textbf{Evaluation benchmarks.} We evaluate our MTP-D on a diverse set of widely used pretraining benchmarks spanning multiple domains, including AGIEval-en \cite{zhong2024agieval} for human-centric standardized examinations; GSM8K \cite{cobbe2021training} and MATH \cite{hendrycks2021measuring} for mathematical reasoning; NaturalQuestions \cite{kwiatkowski2019natural} for knowledge-intensive question answering; SimpleQA \cite{wei2024measuring} for short fact-seeking queries; SuperGPQA \cite{du2025supergpqa} for graduate-level knowledge and reasoning; and TriviaQA \cite{joshi2017triviaqa} for reading comprehension.

\textbf{Speculative decoding and metrics.}
We adopt a main-head-constrained speculative decoding strategy \cite{cai2024medusa, xia2023speculative, xia2024unlocking}, which ensures that the inference results of the main model are exactly identical to those obtained with MTP heads, as detailed in Algorithm~\ref{alg:mtp_spec_refined}. 

Specifically, we report the following metrics: (1) \emph{Accuracy}, defined as the average accuracy of the main head across benchmarks; (2) \emph{Acceptance Rate (AR)}, defined as the percentage of draft tokens generated by each MTP head that are accepted during verification, relative to the total number of tokens verified for that head; (3) \emph{Cumulative Acceptance Rate (CAR)}, defined as the percentage of draft tokens accepted for each MTP head relative to the total number of tokens generated by that head; and (4) \emph{Speedup Ratio}, defined as the relative inference speedup compared to the baseline model, namely the DeepSeek MTP with a single MTP head. Details  are provided in Appendix~\ref{appendix_decoding}.

\begin{table*}[h]
\centering
\small
\setlength{\tabcolsep}{2.4pt}
\begin{tabular}{c l l c c c c c c c c}
\toprule
& & & &
\multicolumn{1}{c}{\textbf{General}} &
\multicolumn{2}{c}{\textbf{Math}} &
\multicolumn{3}{c}{\textbf{Knowledge}} &
\multicolumn{1}{c}{\textbf{STEM}} \\
\cmidrule(lr){5-5}
\cmidrule(lr){6-7}
\cmidrule(lr){8-10}
\cmidrule(lr){11-11}
\textbf{K} & \textbf{Model} & \shortstack{\textbf{Method} \\ \textbf{(MTP)}} & \textbf{H} &
\shortstack{\textbf{AGIEval} \\ \textbf{en}} &
\shortstack{\textbf{GSM8K}} &
\textbf{MATH} &
\shortstack{\textbf{Natural} \\ \textbf{Questions}} &
\shortstack{\textbf{Simple} \\ \textbf{QA}} &
\textbf{TriviaQA} &
\shortstack{\textbf{Super} \\ \textbf{GPQA}} \\
\midrule
\multirow{4}{*}{1} 
  & \multirow{2}{*}{\shortstack{2B \\ Dense}} & MTP & 1 
  & 85.75
  & 84.86
  & 85.66
  & 90.91 
  & 94.28
  & 82.32 
  & 85.10 \\
    &                          & MTP-D & 1 
  & \textbf{88.98}
  & \textbf{88.54}
  & \textbf{89.58}
  & \textbf{94.40}
  & \textbf{94.30}
  & \textbf{86.78} 
  & \textbf{87.93} \\
\cmidrule(lr){2-11}
    & \multirow{2}{*}{\shortstack{A1B \\ MoE}} & MTP & 1
  & 85.41
  & 85.24
  & 85.25
  & 91.34
  & 90.65
  & 80.18
  & 82.51 \\

  &                          & MTP-D & 1
  & \textbf{90.98}
  & \textbf{89.36}
  & \textbf{87.62}
  & \textbf{93.80}
  & \textbf{93.99}
  & \textbf{86.27}
  & \textbf{87.38} \\
\midrule
\multirow{16}{*}{4} 
  & \multirow{8}{*}{\shortstack{2B \\ Dense}} & \multirow{4}{*}{MTP} & 1
  & 81.88 / 81.88
  & 90.32 / 90.32
  & 81.27 / 81.27
  & 89.23 / 89.23
  & 85.17 / 85.17
  & 82.09 / 82.09
  & 82.62 / 82.62 \\
  &                          &          & 2
  & 66.81 / 81.59
  & 82.47 / 91.31
  & 64.63 / 79.52
  & 82.55 / 92.52
  & 68.07 / 79.91
  & 66.84 / 81.42
  & 68.26 / 82.62 \\
  &                          &          & 3
  & 54.28 / 81.24
  & 74.40 / 90.21
  & 50.83 / 78.65
  & 77.14 / 93.45
  & 55.74 / 81.89
  & 53.39 / 79.89
  & 56.65 / 82.99 \\
  &                          &          & 4
  & 45.47 / 83.77
  & 68.52 / 92.09
  & 40.27 / 79.23
  & \textbf{73.51} / 95.30
  & 45.60 / 81.82
  & 43.75 / 81.96
  & 48.62 / 85.83 \\
\cmidrule(lr){3-11}
    &                          & \multirow{4}{*}{MTP-D} & 1
  & \textbf{85.86} / 85.86
  & \textbf{91.69} / 91.69
  & \textbf{84.73} / 84.73
  & \textbf{90.16} / 90.16
  & \textbf{85.20} / 85.20
  & \textbf{84.71} / 84.71
  & \textbf{86.20} / 86.20 \\
  &                          &             & 2
  & 71.96 / 83.81
  & 84.63 / 92.30
  & 69.46 / 81.98
  & 81.15 / 90.01
  & 70.97 / 83.29
  & 71.16 / 84.00
  & 71.96 / 83.48 \\
  &                          &             & 3
  & 61.25 / 85.12
  & 78.33 / 92.55
  & 56.75 / 81.70
  & 74.71 / 92.06
  & 57.57 / 81.12
  & 58.57 / 82.31
  & 62.57 / 86.95 \\
  &                          &             & 4
  & \textbf{52.96} / 86.46
  & \textbf{72.76} / 92.89
  & \textbf{46.42} / 81.81
  & 71.22 / 95.34
  & \textbf{46.27} / 80.38
  & \textbf{48.52} / 82.85
  & \textbf{54.98} / 87.94 \\

\cmidrule(lr){2-11}
    & \multirow{8}{*}{\shortstack{A1B \\ MoE}} & \multirow{4}{*}{MTP} & 1
  & 82.38 / 82.38
  & 89.63 / 89.63
  & 80.54 / 80.54
  & 82.81 / 82.81
  & 83.64 / 83.64
  & 78.06 / 78.06
  & 80.81 / 80.81 \\
  &                          &          & 2
  & 68.11 / 82.67
  & 81.39 / 90.80
  & 63.11 / 78.35
  & 74.54 / 90.02
  & 67.29 / 80.45
  & 61.40 / 78.66
  & 64.48 / 79.79 \\
  &                          &          & 3
  & 57.26 / 84.09
  & 74.31 / 91.30
  & 48.58 / 76.98
  & 68.99 / 92.56
  & 52.80 / 78.47
  & 47.66 / 77.62
  & 53.34 / 82.72 \\
  &                          &          & 4
  & 48.82 / 85.27
  & 68.22 / 91.81
  & 37.85 / 77.92
  & 65.31 / 94.67
  & 41.87 / 79.28
  & 37.45 / 78.63
  & 45.22 / 84.78 \\
\cmidrule(lr){3-11}
    &                          & \multirow{4}{*}{MTP-D} & 1
  & \textbf{85.67} / 85.67
  & \textbf{91.41} / 91.41
  & \textbf{82.22} / 82.22
  & \textbf{86.98} / 86.98
  & \textbf{86.40} / 86.40
  & \textbf{82.46} / 82.46
  & \textbf{85.92} / 85.92 \\
  &                          &             & 2
  & 71.93 / 83.97
  & 83.84 / 91.72
  & 65.10 / 79.18
  & 79.38 / 91.26
  & 73.20 / 84.72
  & 68.42 / 82.98
  & 72.79 / 84.72 \\
  &                          &             & 3
  & 61.55 / 85.58
  & 76.97 / 91.82
  & 49.75 / 76.42
  & 73.24 / 92.26
  & 60.36 / 82.45
  & 55.04 / 80.45
  & 62.88 / 86.38 \\
  &                          &             & 4
  & \textbf{52.47} / 85.25
  & \textbf{71.47} / 92.86
  & \textbf{37.89} / 76.16
  & \textbf{68.05} / 92.92
  & \textbf{49.18} / 81.48
  & \textbf{44.51} / 80.91
  & \textbf{54.28} / 86.44 \\
\bottomrule
\end{tabular}
\caption{Cumulative acceptance rate and acceptance rate of MTP heads for the 2B Dense and A1B MoE models across multiple benchmarks. Here, $H$ denotes the index of the MTP head. For the $K=1$ group, the CAR is identical to the AR. For the $K=4$ group, each cell reports CAR/AR (\%) values. The best CARs of different MTP heads are highlighted in \textbf{bold}. For 2B Dense (A1B) with 1 head, MTP-D improves the average acceptance rate by ~3.09\% (4.11\%). With 4 heads, the improvement is ~3.91\% (4.73\%) for the fourth head.}
\label{tab:distill_ar_mtp}
\end{table*}

\subsection{Main Experiments}
Across multiple pre-training benchmarks, we conduct a comprehensive evaluation of the DeepSeek MTP (as the main competitor) and our MTP-D on 2B Dense and A1B MoE LLMs. The evaluation metrics include the average main-head accuracy, the acceptance rate and cumulative acceptance rate of different MTP heads, and the speedup ratio.

(a) \emph{\textbf{MTP-D largely improves the acceptance rate of MTP heads while maintaining comparable main-head performance.}}
From Table \ref{tab:distill_ar_mtp}, the results demonstrate that our MTP-D substantially improves the acceptance rates of MTP heads with different dense/MoE backbones and model sizes, verifying the effectiveness of our self-distillation. From Table~\ref{tab:distill_acc_mainhead} that reflects the main-head performance, for $K=1$, MTP-D achieves slightly higher average accuracy (11.68) compared to MTP (11.28), and remains comparable when $K=4$ (10.96 vs. 11.06). 

\begin{figure}[!h]
    \centering
    \includegraphics[width=0.88\linewidth]{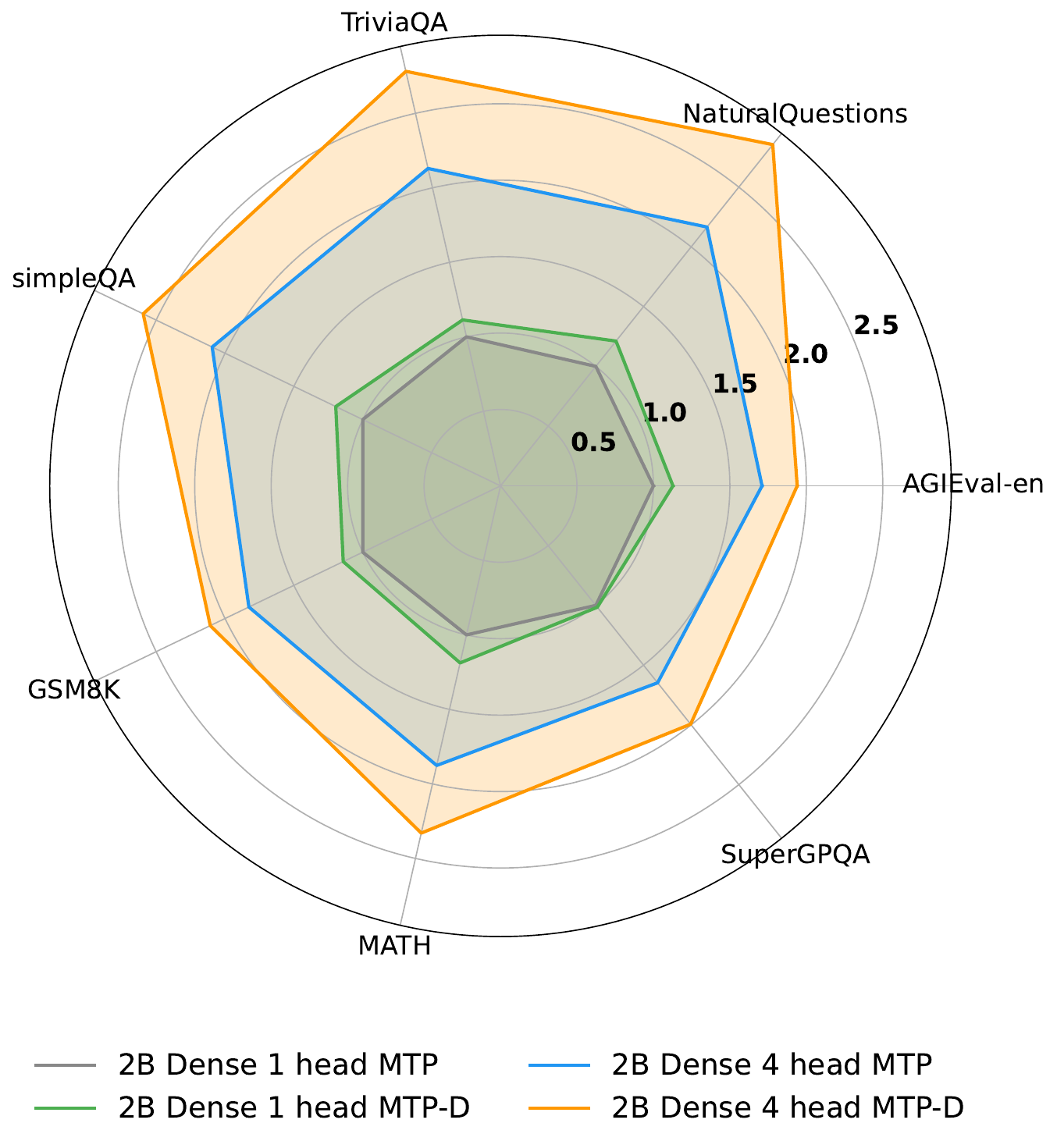}
    \caption{Speedup ratios of the 2B Dense model under different MTP methods and $K$ settings across multiple benchmarks, where the inference speed of 1 head MTP serves as the baseline.}
    \label{fig:2b_dense_speedup}
\end{figure}

(b) \emph{\textbf{MTP-D achieves significantly faster inference speed.}}
Table~\ref{tab:distill_ar_mtp} shows that, for $K=1$, MTP-D improves the (cumulative) acceptance rate by 3.6\% over MTP, corresponding to an approximate 14\% inference speedup. For $K=4$, using the cumulative acceptance rate of the 4-th MTP head as an example, MTP-D yields a 7.5\% increase, translating to a speedup of 22.9\%. Compared to the single-head configuration, the four-head MTP-D achieves a speedup of even up to 107.4\%, highlighting the benefits of scaling up MTP heads while noting that achievable speedup is limited by the cumulative acceptance rate. Comparisons of the speedup ratios are shown in Figures~\ref{fig:2b_dense_speedup} and \ref{fig:a1b_moe_speedup}.

(c) \emph{\textbf{MTP-D functions well with different MTP settings, LLM structures, and model sizes.}}
MTP-D effectively enhances both acceptance rates and inference speed, with performance gains becoming more pronounced as $K$ increases, and that our method is effective across both dense and MoE. In fact, as the size of the main model increases, the performance improvements brought by distillation to the MTP heads become larger, accompanied by correspondingly greater inference speedups.

\begin{figure}[h]
\begin{subfigure}{0.49\textwidth}
    \centering
    \includegraphics[width=\linewidth]{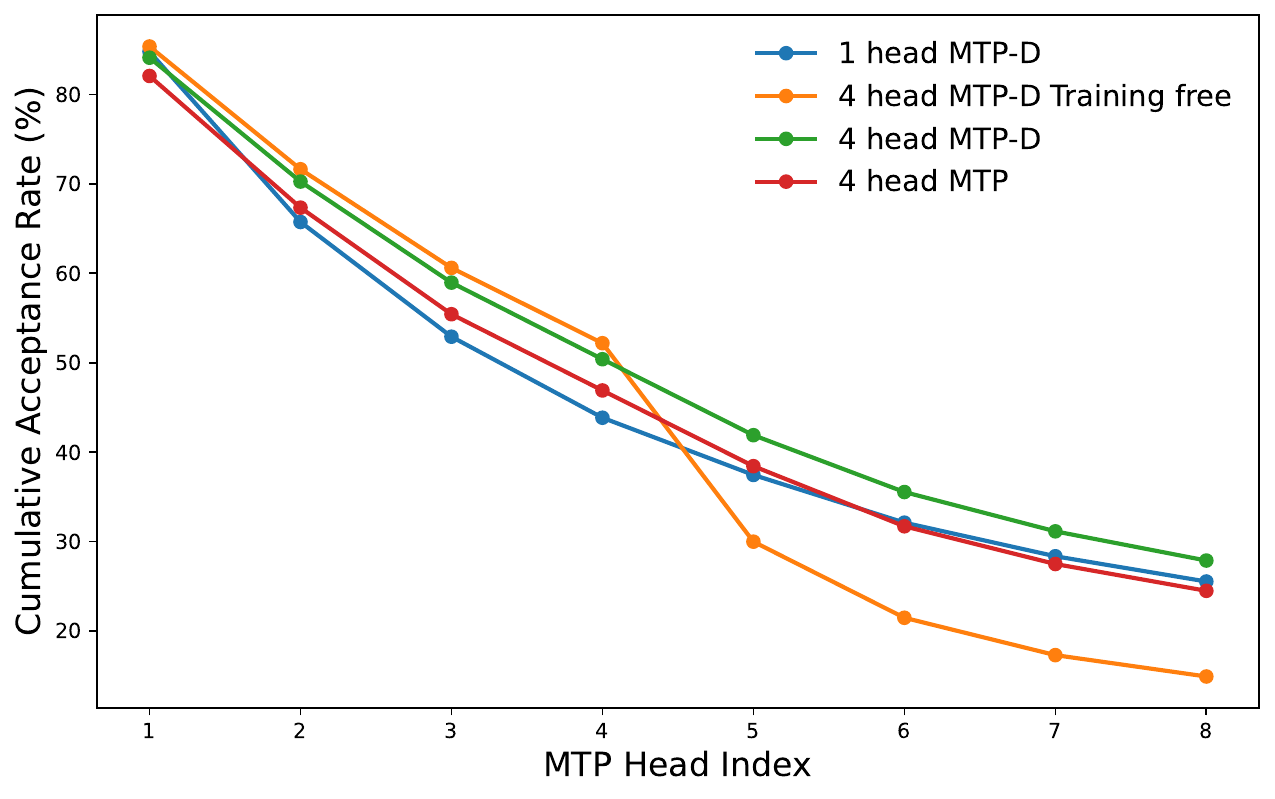}
    \subcaption{Average cumulative acceptance rate across all benchmarks for different looped extension strategies up to 8 loops.}
\end{subfigure}
\hfill
\begin{subfigure}{0.49\textwidth}
    \centering
    \includegraphics[width=\linewidth]{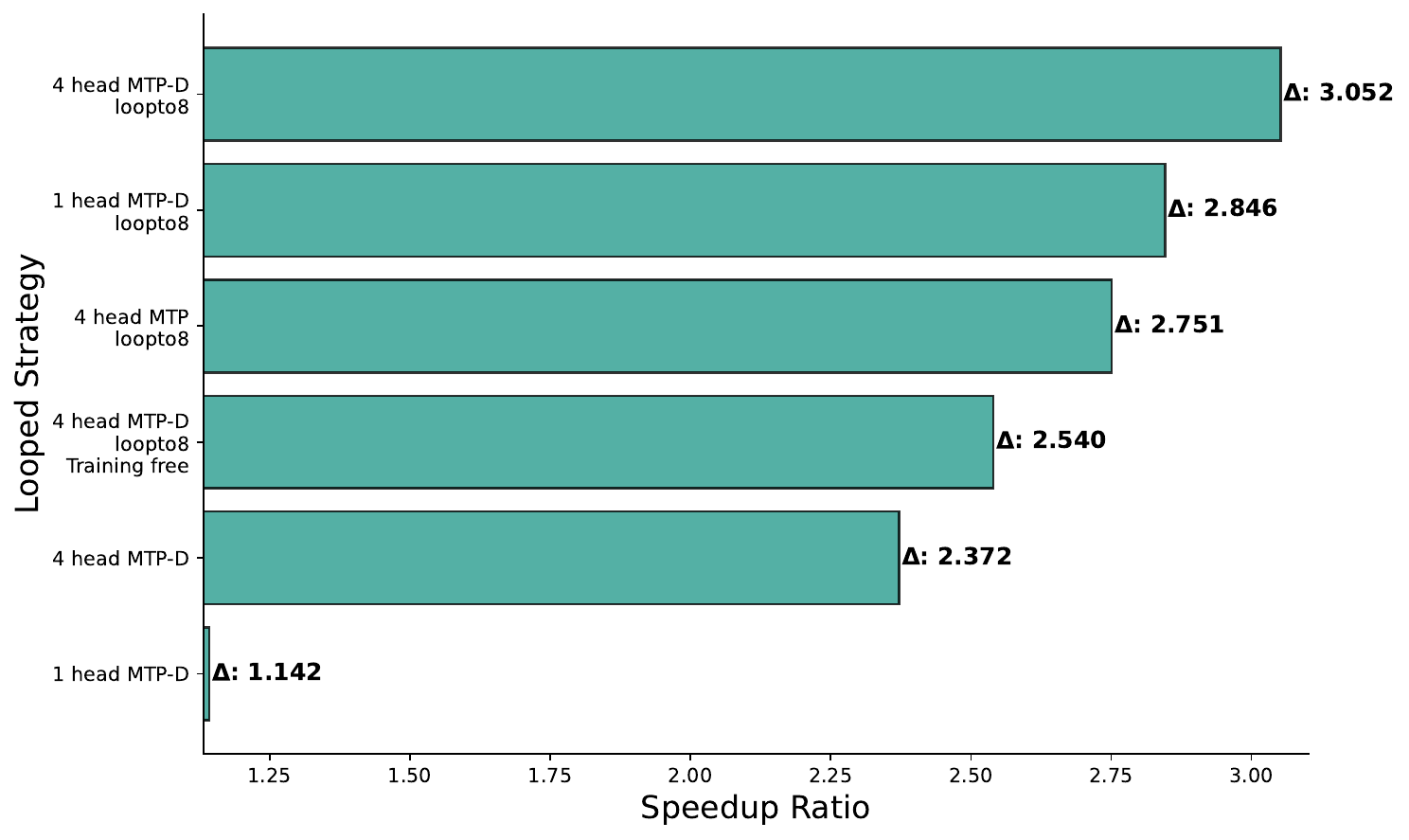}
    \subcaption{Average speedup ratio across all benchmarks for different looped extension strategies up to 8 loops.}
\end{subfigure}
\caption{Comparison of performance across different looped extension strategies with up to 8 loops. The 1-head MTP serves as the baseline.}
\label{fig:loop_zw}
\end{figure}

\begin{table*}
\centering
\small
\setlength{\tabcolsep}{2.4pt}
\begin{tabular}{l c c c c c c c c c}
\toprule
& &
\multicolumn{1}{c}{\textbf{General}} &
\multicolumn{2}{c}{\textbf{Math}} &
\multicolumn{3}{c}{\textbf{Knowledge}} &
\multicolumn{1}{c}{\textbf{STEM}} &
\\
\cmidrule(lr){3-3}
\cmidrule(lr){4-5}
\cmidrule(lr){6-8}
\cmidrule(lr){9-9}
\multicolumn{2}{c}{\textbf{Strategy}} &
\shortstack{\textbf{AGIEval} \\ \textbf{en}} &
\shortstack{\textbf{GSM8K}} &
\textbf{MATH} &
\shortstack{\textbf{Natural} \\ \textbf{Questions}} &
\shortstack{\textbf{Simple} \\ \textbf{QA}} &
\textbf{TriviaQA} &
\shortstack{\textbf{Super} \\ \textbf{GPQA}} &
\textbf{Mean} \\
\midrule
\multicolumn{2}{c}{\textbf{MTP-D}} 
& 19.20 | 88.98 
& 1.74 | 88.54 
& 1.70 | 89.58 
& 13.00 | 94.30 
& 2.22 | 94.30 
& 37.08 | 86.78 
& 6.81 | 87.93 
& \textbf{11.68} | 90.06 \\
\midrule
\multicolumn{2}{c}{no detach} 
& 19.60 | 93.67 
& 1.74 | 93.70 
& 1.50 | 93.78 
& 11.58 | 96.88
& 2.01 | 96.40 
& 30.42 | 92.31 
& 4.45 | 93.61 
& \underline{10.19 | 94.34} \\
\midrule
\multirow{3}{*}{Top$N$} & 10000x2
& 18.18 | 89.80
& 1.44 | 90.12
& 1.40 | 89.84
& 13.24 | 94.19
& 2.41 | 93.74
& 34.72 | 87.33
& 7.53 | 88.53
& 11.27 | 90.51 \\ 
& 1000
& 20.64 | 90.44 
& 1.36 | 88.63 
& 1.50 | 89.52 
& 12.77 | 94.73 
& 2.29 | 94.06 
& 34.86 | 88.75 
& 6.88 | 87.14 
& 11.47 | 90.47 \\
& 1
& 18.91 | 87.55
& 1.82 | 87.00
& 1.65 | 85.76
& 13.38 | 92.06
& 2.24 | 92.56
& 36.11 | 83.11
& 6.16 | 83.72
& 11.47 | 87.39 \\ 
\midrule
\multirow{4}{*}{$\beta_k$} & 1.5
& 20.06 | 91.06
& 1.67 | 88.95
& 1.50 | 89.44
& 14.02 | 94.43
& 1.99 | 94.52
& 37.08 | 88.56
& 5.19 | 88.65
& 11.64 | 90.80 \\ 
& 0.5
& 21.17 | 89.03
& 1.44 | 88.52
& 1.55 | 88.10
& 12.71 | 93.31
& 2.13 | 92.84
& 35.69 | 87.38
& 5.59 | 88.86
& 11.47 | 89.72 \\ 
& 0.3
& 18.68 | 90.20
& 2.12 | 85.89
& 1.75 | 88.19
& 13.10 | 93.99
& 1.92 | 91.05
& 34.72 | 85.50
& 6.91 | 87.02
& \underline{11.31 | 88.83} \\ 
& 0.1
& 18.07 | 88.83
& 1.74 | 84.69
& 1.70 | 85.71
& 12.85 | 92.85
& 2.18 | 93.18
& 34.31 | 84.00
& 6.05 | 84.34
& \underline{10.99 | 87.66} \\ 
\midrule
\multirow{2}{*}{KL} & reverse
& 22.14 | 89.32
& 1.67 | 89.06
& 1.50 | 89.17
& 12.83 | 94.38
& 2.06 | 90.99
& 34.72 | 89.81
& 6.40 | 88.78
& 11.62 | 90.22 \\ 
& hybrid
& 20.12 | 88.20
& 1.67 | 87.69
& 1.55 | 87.84
& 12.60 | 94.31
& 1.92 | 92.44
& 32.36 | 85.67
& 6.73 | 86.30
& \underline{10.99 | 88.92} \\ 

\bottomrule
\end{tabular}
\caption{Ablation results for MTP-D with 1 head across multiple benchmarks. Each cell shows two numbers separated by ``\textbar'' (left: main-head accuracy; right: MTP-head AR). The MTP-D configuration is as follows: \texttt{detach}, Top$N = 10{,}000$, $\beta_k = 1.0$, and the KL function is the forward KL. Underlined entries indicate strategies that significantly reduce main-head performance.}
\label{tab:distill_ablation}
\end{table*}

\subsection{Ablation Study and Model Analysis}
We conduct comprehensive ablation experiments to validate the contribution of each component in our MTP-D, using main-head accuracy, MTP-head acceptance rates, and their associated losses as evaluation metrics.

Taking a single MTP head as an example, we conduct ablation studies on four key strategies of MTP-D: \texttt{detach}, Top$N$, $\beta_k$, and the KL function. Our default configuration is: \texttt{detach}, Top$N = 10{,}000$, $\beta_k = 1.0$, and forward KL. Analysis is based on the loss curves in Fig.~\ref{fig:ablation_loss_compare} and benchmark results in Table~\ref{tab:distill_ablation}.

First, removing \texttt{detach} greatly increases the main-head loss (+0.079) and reduces its performance (-1.49\%), demonstrating that stop-gradient operation on the main-head logits $\hat{\mathbf{Q}}$ effectively prevents gradient backpropagation through $\hat{\mathbf{Q}}$, thereby mitigating interference to the main head. 

Next, we examine Top$N$ in the vocabulary dimension $V$ of $\hat{\mathbf{Q}}$. Due to the long-tail distribution in the vocabulary space (Fig.~\ref{fig:topN_probs}), directly distilling over all logits leads to computational redundancy, excessive memory usage, and numerical instability in logarithmic operations, potentially causing gradient oscillation or vanishing (Fig.~\ref{fig:illustration_of_loss_detach}). As Top$N$ increases, the main-head loss slightly rises, but performance remains largely comparable, whereas the MTP-head loss decreases significantly until Top$N$ exceeds a threshold (1,000), beyond which gains plateau. Here, ``10000$\times$2'' denotes the union of the Top$N$ sets for $\hat{\mathbf{Q}}$ and $\hat{\mathbf{P}}$. Considering the loss factor, we select Top$N = 10{,}000$.  

The $\beta_k$ factor is another key strategy. Increasing $\beta_k$ improves both main-head performance and MTP-head AR, but overly large values significantly increase main-head loss (e.g., $\beta_k = 1.5$, loss +0.028); thus, we choose $\beta_k = 1.0$. Finally, we ablate the KL function (details in Appendix~\ref{appendix_results_distill}) and find that forward KL achieves the best trade-off between loss and evaluation performance.

For models with 4 heads, we also perform ablation experiments on the multi-level distillation strategies and the corresponding $\beta_k$. Details are provided in Appendix~\ref{appendix:4head_ablation}.

\subsection{Results of Looped MTP Extension}
\label{subsec:results_loopeddistillmtp}
In this section, we investigate the loop scalability and upper bounds of models with different numbers of MTP heads, shown in Figures~\ref{fig:loop_zw}. In addition, in Appendix~\ref{appendix_loop_trainfree}, Figures~\ref{fig:looptrainfree_allbench_cumulative} and \ref{fig:looptrainfree_allbench} show CAR and AR for MTP head loops up to 8 under a training-free setting. Figures~\ref{fig:L-distill_loopto8} and \ref{fig:L-distill_loopto16} present CAR for loops up to 8 and 16, respectively, under continued pre-training. Table~\ref{tab:results_loop_speedup} reports the speedup ratios for various looped extension experiments. From these results, we draw the following insights:

\emph{\textbf{Insight 1: Cascaded MTP is inherently scalable due to the structural consistency and input-output similarity.}}
When extending 4 MTP heads in a training-free manner, the acceptance rate drops mainly at the loop junction (e.g., from 80\% to 50\%) but remains acceptable, while subsequent heads within the group gradually recover to levels comparable to trained heads.

\emph{\textbf{Insight 2: MTP-D exhibits superior scalability compared to MTP.}}
In training-free extension from 1 to 8 loops, MTP quickly collapses (e.g., CAR drops to 0.6\% by the 3-rd head on AGIEval-en), while MTP-D maintains much higher CAR (26.70\%). This indicates that distillation improves consistency between MTP heads and the main head, leading to better scalability. Continued pre-training further enhances CAR and speedup across all benchmarks.

\emph{\textbf{Insight 3: Grouped MTP heads exhibit stronger loop scalability than a single MTP head.}}
Compared with single-head extension, grouped MTP heads consistently achieve higher CAR and speedup under looped extension, indicating that intra-group correlations are preserved and effectively enhance continued pre-training.

\emph{\textbf{Insight 4: A limited data size is sufficient for looped MTP extension.}}
Scaling the data size from 70B to 350B results in only marginal gains in CAR and speedup ratios for MTP-D.

\emph{\textbf{Insight 5: MTP heads as distillation teachers with the main head enhance head consistency and loop scalability.}}
An ensemble of main- and MTP-head logits achieves comparable CAR while enhancing main-head performance and loop scalability over standard MTP-D.

\emph{\textbf{Insight 6: MTP-D have more potential up to 16 heads.}} 
In the 4-to-16 MTP-D setting, the 16th MTP head maintains a CAR of 5–10\%. For benchmarks with high acceptance rates (e.g., MATH, SimpleQA, etc), loops up to 16 still provide notable speedup. 
However, due to the cascaded MTP architecture, CAR gradually declines with more heads, ultimately limiting practical scalability.
\section{Related Work}
\label{related_work}

\textbf{MTP Architectures.}
The MTP paradigm has been widely studied in both research and industrial LLMs, as it provides rich supervision, improves sample efficiency, and accelerates inference. The classical MTP architecture was introduced by the Meta team~\cite{gloeckle2024better}, and DeepSeek-V3~\cite{liu2024deepseek} further proposed the cascaded DeepSeek MTP, which has seen broad adoption. Several studies have explored MTP’s mechanisms and applications in LLMs, including adaptation to small models~\cite{aynetdinov2025pre}, strategies for parallel token prediction~\cite{mehra2025multi}, and its impact on relational learning~\cite{zhong2025understanding}. Other works investigate MTP variants to enhance training~\cite{grivas2025fast, mahajan2025beyond} or accelerate inference~\cite{samragh2025your, liu2025mtp, cai2025fastmtp}.

\textbf{Distillation in LLMs.}  
Supervised knowledge distillation~\cite{buciluǎ2006model, hinton2015distilling} is a classical technique successfully applied to auto-regressive models~\cite{sanh2019distilbert} and has become central in industrial LLMs, particularly during post-training~\cite{liu2025did, agarwal2024policy, gu2023minillm, wen2023f}. For instance, DeepSeek R1~\cite{guo2025deepseek} demonstrated that distilling chain-of-thought data enhances reasoning, making distillation a key part of LLM pipelines. It has also been explored in pre-training, with its scaling laws systematically analyzed~\cite{busbridge2025distillation}. Inspired by these advances, we adapt distillation to MTP, where the main head serves as teacher and MTP heads as students, enabling natural self-distillation.

\section{Conclusion}
\label{conclusion}

In this work, we propose MTP-D, a self-distillation framework for MTP in pre-training, along with a looped extension strategy in continued pre-training. It enables significantly better MTP head acceptance rate and thus faster inference speed with different LLM backbones, while maintaining comparable main-head performance and relatively marginal additional training costs.
MTP-D successfully enables up to 8-16 MTP heads with further inference speed gain.
Extensive experiments demonstrate the superiority of our methods. The proposed methods and the associated insights provide valuable guidance for improving the pre-training and inference of future LLMs with MTP, as well as practical inspiration for broader applications of MTP.

\section*{Limitations}
\label{sec:limitations}

This work focuses on self-distillation for multi-token prediction during pre-training. Future work should explore its adaptation to post-training and investigate the scalability of MTP-D during post-training. Due to resource constraints, the theoretical relationship between optimal $\alpha_k$ and $\beta_k$ and the number of MTP heads $K$ has not been sufficiently explored, and our method has not been validated on diverse datasets or ultra-large models. We hope it can be validated in industrial-scale LLM pre-training in the future.

\bibliography{custom}

\newpage
\appendix

\section{The Use of LLMs}
\label{appendix:usage_llm}

In this paper, we leveraged LLMs to support and refine the writing. Specifically, LLMs were used for grammar and spelling correction, as well as polishing linguistic expressions to enhance clarity and readability.
\section{Description of Model Configurations and Training Details}
\label{appendix_configdetails}
Table~\ref{tab:model_configs} summarizes the detailed configurations of the two models used in our experiments, namely 2B Dense and 10B A1B MoE. The 2B Dense LLM comprises 32 decoder layers and $K$ MTP heads. The N10BA1B MoE LLM consists of 22 decoder layers (1 ``af'' layer and 21 ``ae'' layers) and $K$ MTP heads. Both models employ GQA attention and a vocabulary size of 122,880 tokens. The \texttt{mtp\_head\_layers} is set to 1. Following the DeepSeek MTP architecture, the MTP heads share the embedding layer and output head with the main head. Notably, for both dense and MoE models, a single dense layer rather than a MoE layer is adopted as the MTP head, which is consistent with LongCat~\cite{team2025longcat}.

Table~\ref{tab:training_hyperparams} reports the training hyperparameters of MTP-D in both the pre-training and continued pre-training stages. These mainly differ in warmup strategy and data size: in continued pre-training, looped MTP disables learning-rate warmup and is trained on 70B tokens. The MTP loss coefficient $\alpha_k$ is set to 0.3 following DeepSeek-V3~\cite{liu2024deepseek}, and a fine-grained search is performed for $\beta_k$, resulting in $\beta_k=1.0$ for single-head models ($K=1$) and $\beta_k=0.5$ for four-head models ($K=4$).

Specifically, training employed the AdamW optimizer \cite{adam2014method} with parameters $(\beta_1, \beta_2)=(0.9, 0.95)$ and $\epsilon=1\times10^{-8}$. A weight decay of 0.1 and gradient clipping with norm 1.0 were applied. The learning rate followed the Warmup-Stable-Decay (WSD) scheduler \cite{hu2024minicpm}, with a maximum of $3\times10^{-4}$ and a minimum of 0. The sequence length was set to 4096 tokens, and the batch size was 8192 tokens. RMSNorm was used for normalization, and rotary position embedding (ROPE) \cite{su2024roformer} was applied. For MTP-D, warmup steps were set to 1000 with a total data size of 350B tokens, while for looped MTP, warmup was 0 and the data size was 70B tokens. All experiments were conducted on the FineWeb-Edu-350BT dataset, which is a randomly sampled subset of approximately 350 billion tokens from the whole FineWeb-Edu corpus \cite{penedo2024fineweb}. Training was conducted using the TorchTitan pretraining framework \cite{liang2024torchtitan}. Our primary experiments were conducted on 256 NVIDIA H20 GPUs (98 GB) for approximately 30 days.

\begin{table}[h]
\centering
\caption{Detailed description of model configurations.}
\label{tab:model_configs}
\small
\begin{tabular}{lcc}
\toprule
\textbf{Hyperparameter} & \textbf{2B Dense} & \textbf{N10BA1B MoE} \\
\midrule
dim & 2048 & 1536 \\
n\_heads & 16 & 32 \\
n\_kv\_heads & 4 & 4 \\
head\_dim & 128 & 128 \\
embedding\_tie & False & False \\
qk\_norm & True & True \\
ffn\_hidden\_dim & 6144 & 6912 \\
max\_seq\_len & 4096 & 4096 \\
rope\_theta & 10000.0 & 10000.0 \\
norm\_eps & 1e-5 & 1e-5 \\
decoder\_layers & 32 af & af + 21 ae \\
\midrule
expert\_hidden\_dim & - & 768 \\
num\_experts & - & 128 \\
use\_shared\_expert & - & True \\
num\_shared\_experts & - & 1 \\
top\_k & - & 8 \\
use\_grouped\_mm & - & True \\
initializer\_range & - & 0.02 \\
\midrule
num\_mtp\_heads & K & K \\
mtp\_head\_layers & 1 & 1 \\
mtp\_dense\_or\_moe & dense & dense \\
\bottomrule
\end{tabular}
\end{table}

\begin{table}[h]
\centering
\caption{Training hyperparameter details for MTP-D and looped MTP}
\small
\label{tab:training_hyperparams}
\setlength{\tabcolsep}{5pt}
\begin{tabular}{l l l}
\toprule
\textbf{Hyperparameter} & \textbf{MTP-D} & \textbf{Looped MTP} \\
\midrule
Optimizer & \multicolumn{2}{c}{AdamW} \\
Adam $(\beta_1, \beta_2)$ & \multicolumn{2}{c}{(0.9, 0.95)} \\
Adam $\epsilon$ & \multicolumn{2}{c}{$1 \times 10^{-8}$} \\
Weight decay & \multicolumn{2}{c}{0.1} \\
Clip grad norm & \multicolumn{2}{c}{1.0} \\
Max lr & \multicolumn{2}{c}{$3.0 \times 10^{-4}$} \\
Min lr & \multicolumn{2}{c}{0} \\
Lr decay & \multicolumn{2}{c}{Cosine} \\
Decay rate & \multicolumn{2}{c}{10\%} \\
Sequence length & \multicolumn{2}{c}{4096} \\
Batch size & \multicolumn{2}{c}{8192} \\
Normalization & \multicolumn{2}{c}{RMSNorm} \\
Vocabulary size & \multicolumn{2}{c}{122880} \\
Positional encoding & \multicolumn{2}{c}{ROPE} \\
Warmup steps & \textbf{1000} & \textbf{0} \\
Data Size & \textbf{350B} & \textbf{70B} \\
\midrule
$\alpha_k$ & \multicolumn{2}{c}{0.3} \\
\multirow{2}{*}{$\beta_k$} & \multicolumn{2}{c}{1.0 \space for \space $K{=}1$} \\
 & \multicolumn{2}{c}{0.5 \space for \space $K{=}4$} \\
\bottomrule
\end{tabular}
\end{table}

\section{Analysis of Logits Probability Distribution}
\label{appendix_logitprob}
\begin{figure*}[h]
\begin{subfigure}{0.32\textwidth}
    \centering
    \includegraphics[width=\linewidth]{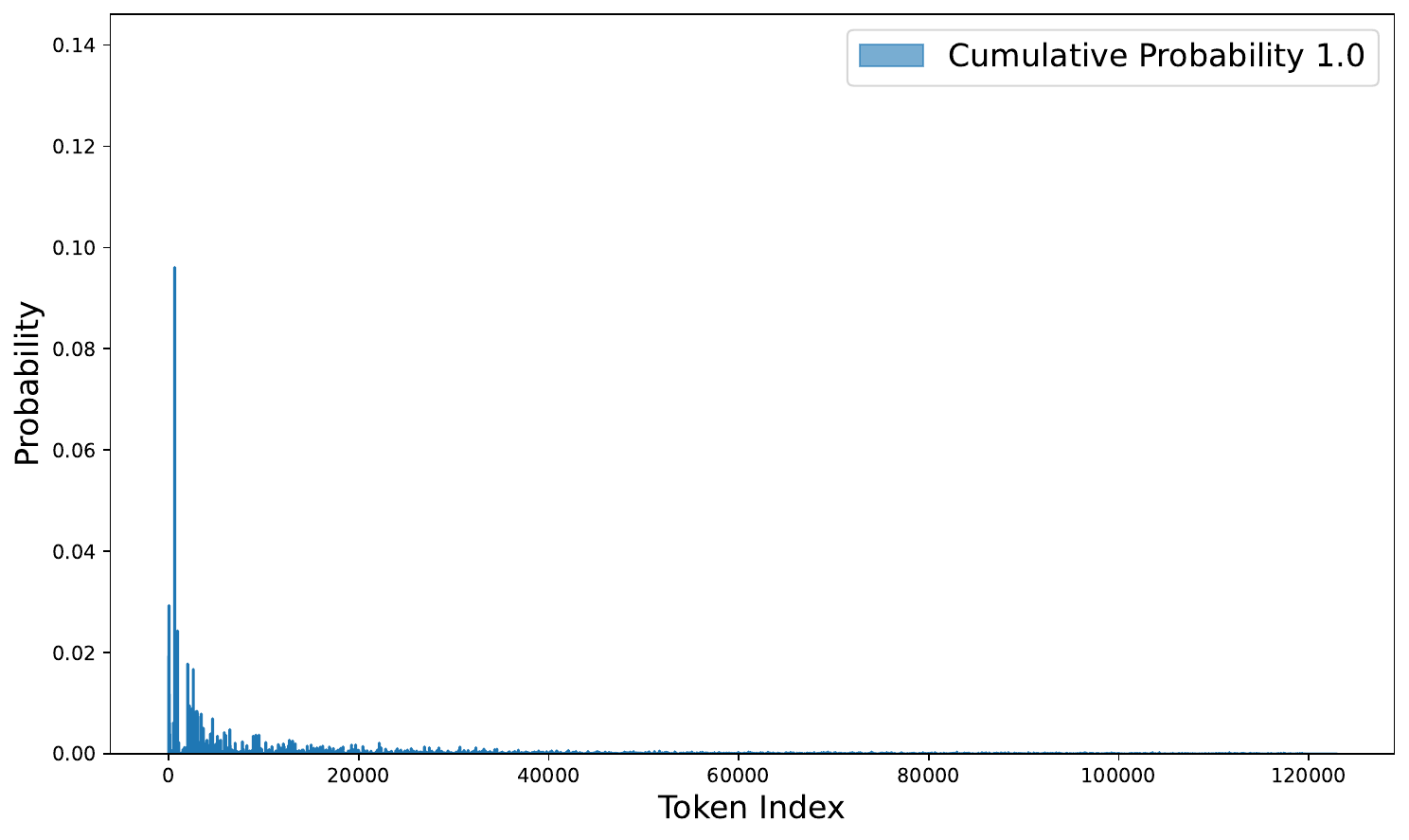}
    \subcaption{All.}
\end{subfigure}
\hfill
\begin{subfigure}{0.32\textwidth}
    \centering
    \includegraphics[width=\linewidth]{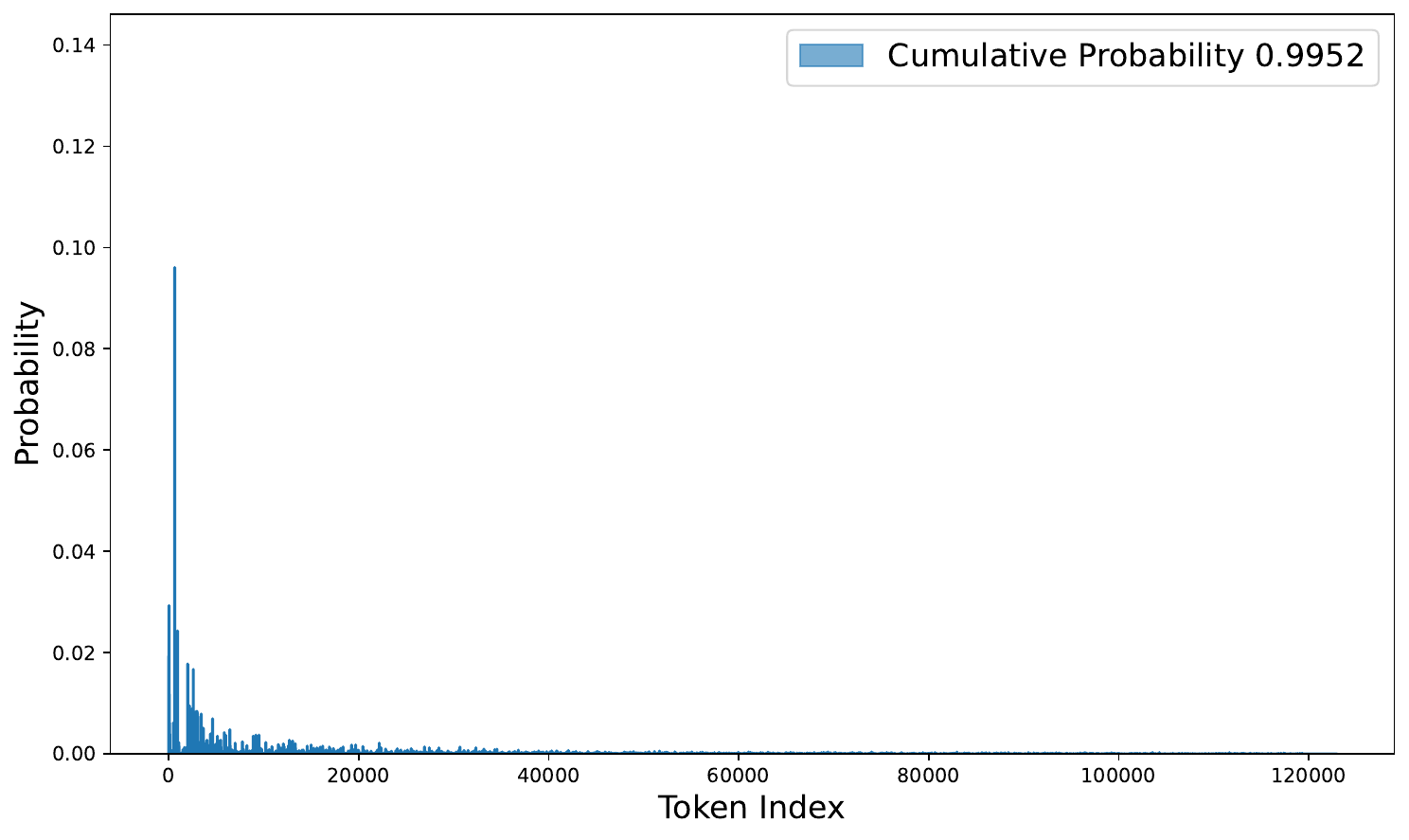}
    \subcaption{Top10000.}
\end{subfigure}
\begin{subfigure}{0.32\textwidth}
    \centering
    \includegraphics[width=\linewidth]{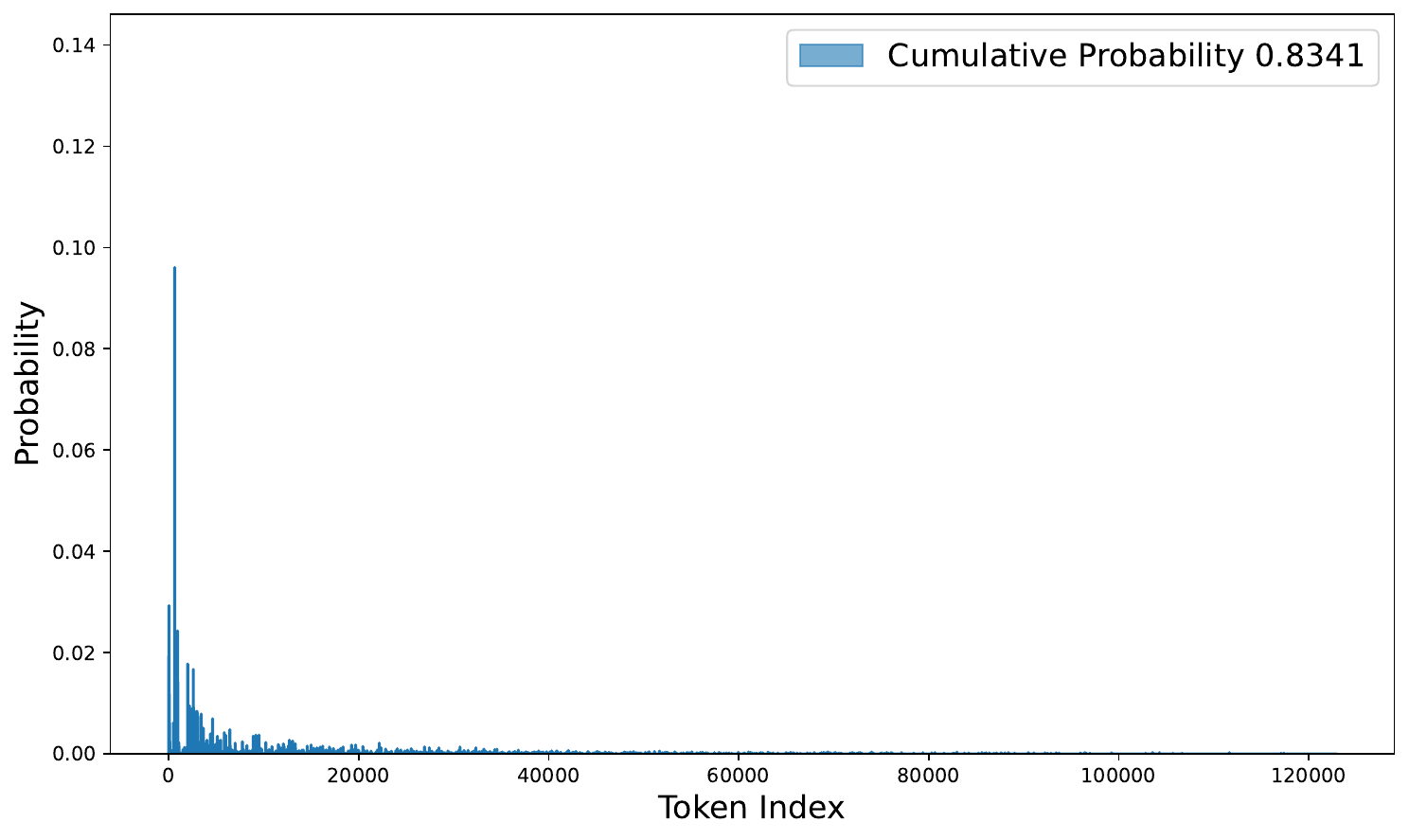}
    \subcaption{Top1000.}
\end{subfigure}
\hfill
\begin{subfigure}{0.32\textwidth}
    \centering
    \includegraphics[width=\linewidth]{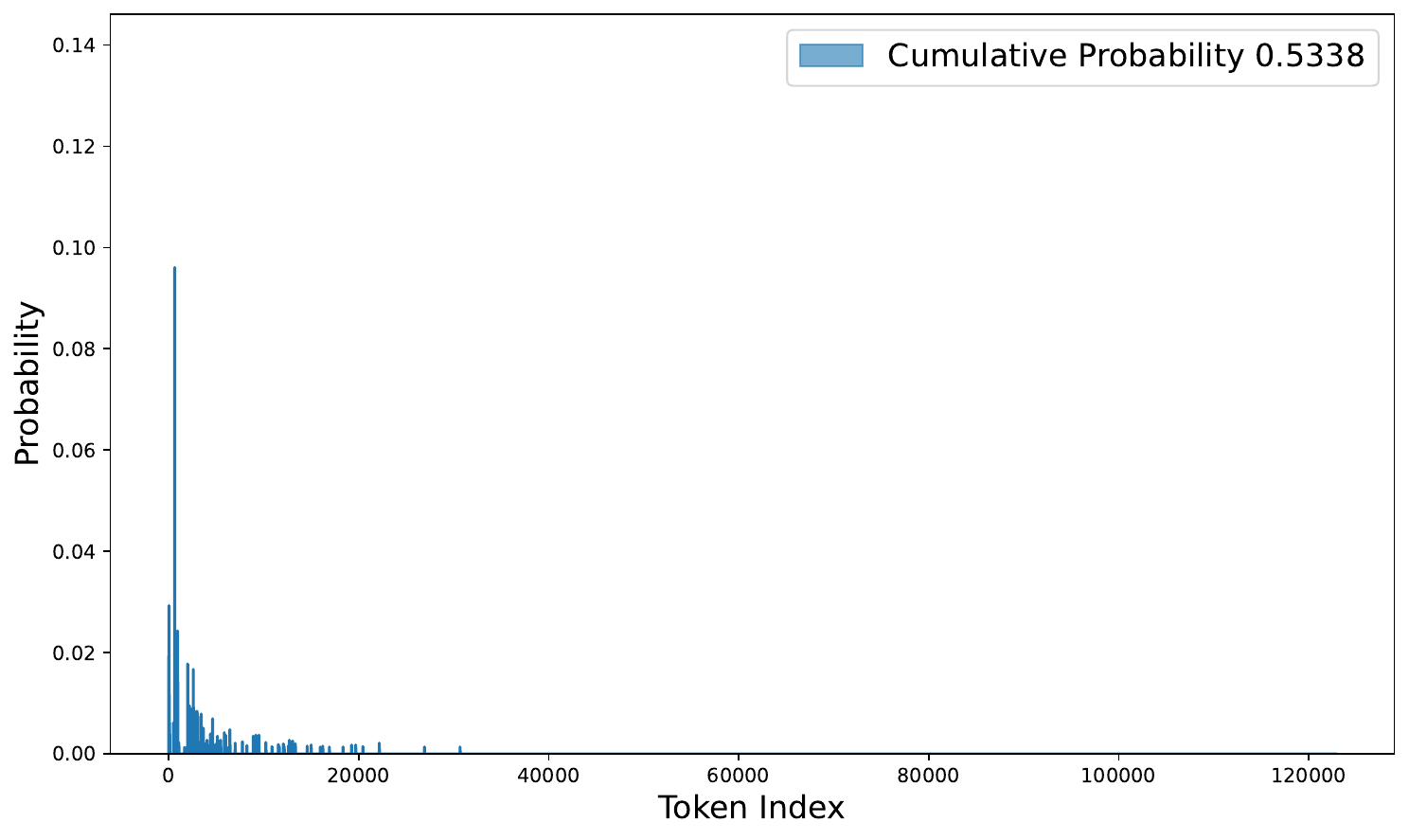}
    \subcaption{Top100.}
\end{subfigure}
\hfill
\begin{subfigure}{0.32\textwidth}
    \centering
    \includegraphics[width=\linewidth]{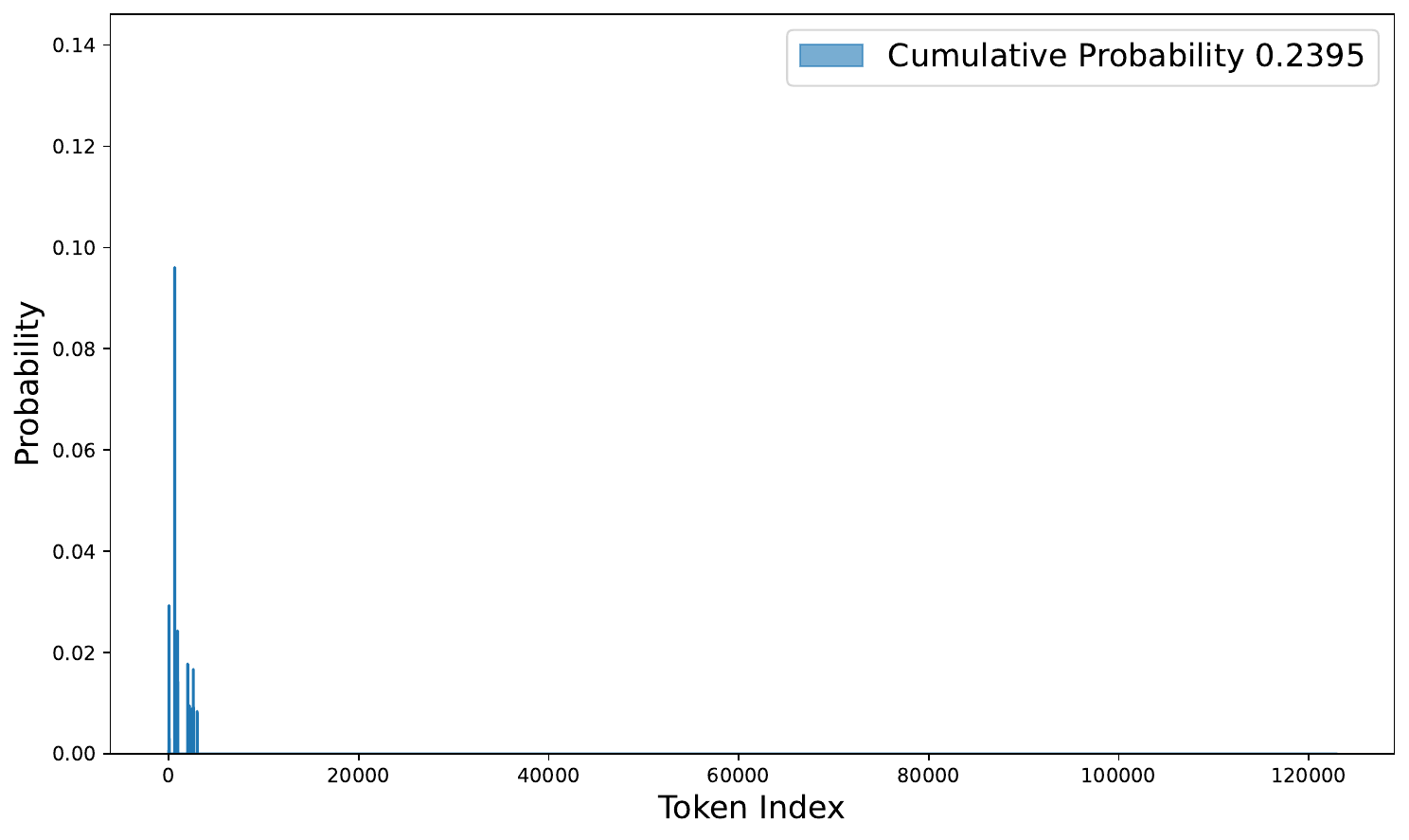}
    \subcaption{Top10.}
\end{subfigure}
\hfill
\begin{subfigure}{0.32\textwidth}
    \centering
    \includegraphics[width=\linewidth]{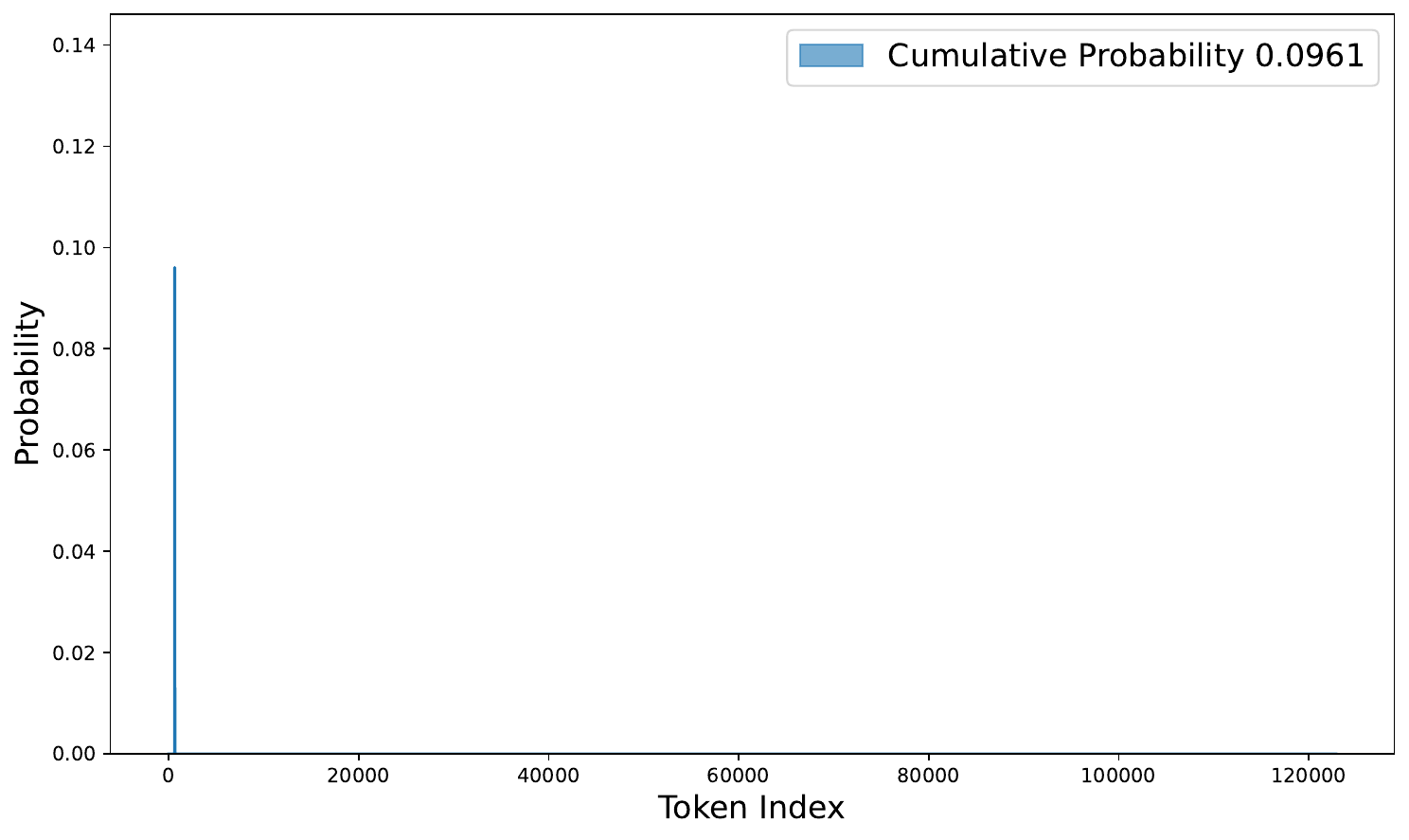}
    \subcaption{Top1.}
\end{subfigure}
\caption{Illustration of the probability distributions of the main head logits for the $t$-th token under different Top$N$ settings.}
\label{fig:topN_probs}
\end{figure*}

As illustrated in Figure~\ref{fig:topN_probs}, the main-head logits for the $t$-th token exhibit a long-tailed distribution after softmax, with most candidate tokens assigned near-zero probabilities. Specifically, the cumulative probability reaches 0.9952 for Top$N=10{,}000$ and 0.8341 for Top$N=1{,}000$. Directly distilling over all vocabulary logits incurs computational redundancy, high memory consumption, and numerical instability, while abundant low-probability signals can hinder effective learning from high-probability ones. Based on these observations, we select Top$N=10{,}000$.

\section{Speculative Decoding}
\label{appendix_decoding}

\begin{figure}[h]
    \centering
    \includegraphics[width=0.99\linewidth]{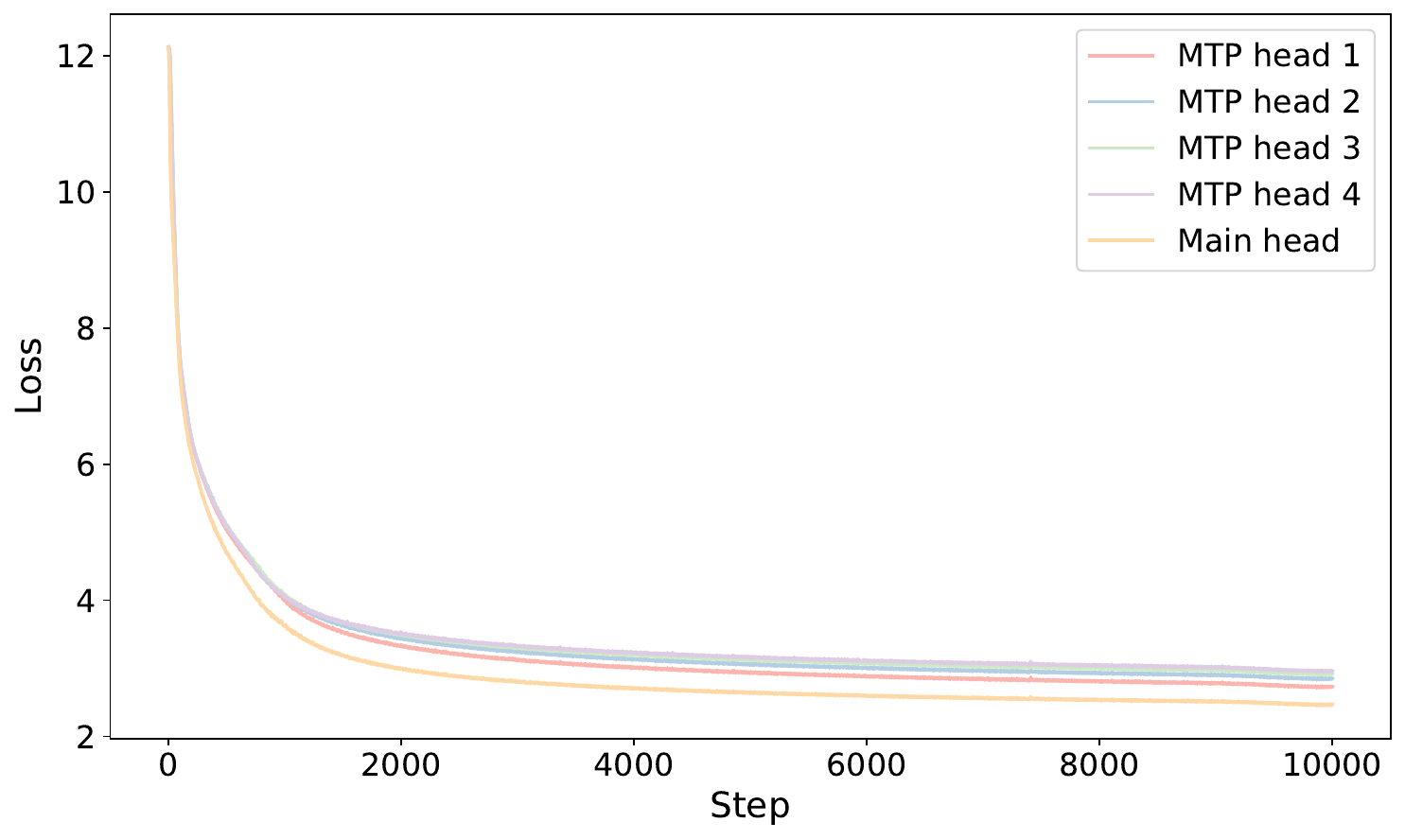}
    \caption{Illustration of the training loss curves for a 2B dense LLM equipped with 4 heads. The final losses of the main head and MTP heads 1 to 4 are 2.47 and [2.73, 2.85, 2.92, 2.96], respectively.}
    \label{fig:illustration_of_loss}
\end{figure}

To ensure that the inference results of the main model are exactly identical to those obtained with MTP heads, we adopt a main-head-constrained speculative decoding strategy. To guarantee comparability of inference times across all experiments, inference is performed entirely locally using a single-batch setup, with greedy decoding employed for all samples, and KV cache is not used during inference. Considering the answer lengths in the pretraining benchmarks and the capabilities of the pretraining models, the maximum generation length for each sample is set to 100 tokens.
\begin{equation}
\mathrm{AR}_j
= \frac{\sum_{s=1}^{S} A_j^{(s)}}{\sum_{s=1}^{S} C^{cmp,(s)}_j},
\qquad j = 1, \dots, K .
\end{equation}
\begin{equation}
\mathrm{CAR}_j
= \frac{\sum_{s=1}^{S} A_j^{(s)}}{\sum_{s=1}^{S} C_{\text{step}}^{(s)}},
\qquad j = 1, \dots, K .
\end{equation}
Here, $S$ denotes the total number of samples in the corresponding benchmark, and $K$ represents the number of MTP heads. $A_j$ is the number of tokens generated by the $j$-th MTP head that are accepted by the main head during verification, $C^{\text{cmp}}_j$ is the total number of tokens from the $j$-th MTP head that are verified by the main head, and $C_{\text{step}}$ represents the total number of tokens generated by each MTP head, which is the same across all MTP heads.

Notably, due to the pre-training evaluation setup, the maximum generation length is limited to 100. The overall inference speedup from the increased acceptance rate is expected to grow with longer generation lengths, which is particularly important for long-sequence reasoning. In future work, we aim to extend this approach to post-training scenarios.

\section{Detailed Analysis for MTP-D}
\label{appendix_results_distill}

\begin{figure}[h]
    \centering
    \includegraphics[width=0.99\linewidth]{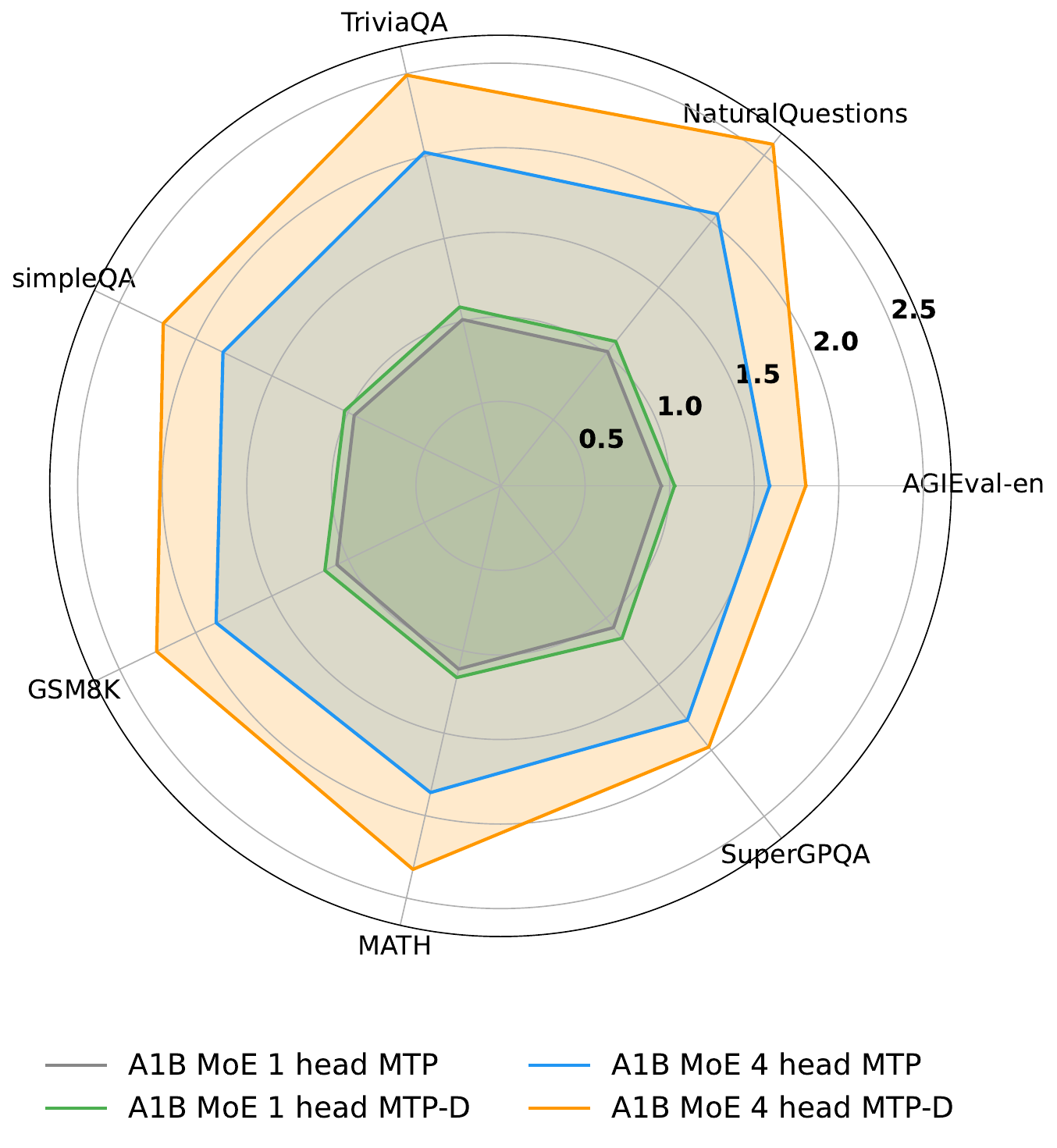}
    \caption{Speedup ratios of the A1B MoE model under different MTP methods and $K$ settings across multiple benchmarks, where the inference speed of 2B Dense DeepSeek MTP with $K=1$ serves as the baseline.}
    \label{fig:a1b_moe_speedup}
\end{figure}

The reverse KL loss, defined as $\mathrm{KL}(q \| p) = \mathbf{E}_{\mathbf{x} \sim q} \left[ \log \frac{q(\mathbf{x})}{p(\mathbf{x})} \right]$, encourages the model distribution $q$ to align with the modes of the target distribution $p$. The loss function for the reverse KL is given as follows:
\begin{equation}
\begin{aligned}
\mathcal{L}^{\text{KL, R}}_{\mathrm{mtp_k}}
&= \mathrm{KL}\!\left(\,
\text{sg}(\bar{\mathbf{Q}})_{k+1:T+1},\bar{\mathbf{P}}_{k+1:T+1}^{k}
\right)
\end{aligned}
\end{equation}
\begin{equation}
\bar{\mathbf{Q}}_{k+1:T+1}^{k} = \text{log}\left(\sigma\left(\hat{\mathbf{Q}}_{k+1:T+1}^{k}\left[...\right]\right)\right)
\label{eq:mtp_KL_loss_Q_R}
\end{equation}
\begin{equation}
\bar{\mathbf{P}}_{k+1:T+1} = \sigma\left(\hat{\mathbf{P}}_{k+1:T+1}\left[...\right]\right)
\label{eq:mtp_KL_loss_P_R}
\end{equation}

The hybrid KL loss is defined as a weighted combination of the forward KL and reverse KL.

\subsection{Details of Main Experiments}
Table~\ref{tab:distill_acc_mainhead} provides the average main-head accuracy across multiple benchmarks for the main experiments. Figure~\ref{fig:a1b_moe_speedup} compares the speedup ratios of different MTP methods using A1B MoE as the backbone.

\begin{table}[h]
\centering
\small
\begin{tabular}{c l c c}
\toprule
\textbf{K} & \textbf{Model} & \textbf{Method (MTP)} & \textbf{Mean} \\
\midrule
\multirow{4}{*}{1} 
& \multirow{2}{*}{2B Dense} & MTP   & 11.28 \\
&                         & MTP-D & 11.68 \\
\cmidrule(lr){3-4}
& \multirow{2}{*}{A1B MoE}  & MTP   & 13.75 \\
&                         & MTP-D & 14.15 \\
\midrule
\multirow{4}{*}{4} 
& \multirow{2}{*}{2B Dense} & MTP   & 11.06 \\
&                         & MTP-D & 10.96 \\
\cmidrule(lr){3-4}
& \multirow{2}{*}{A1B MoE}  & MTP   & 13.80 \\
&                         & MTP-D & 13.72 \\
\bottomrule
\end{tabular}
\caption{Mean accuracy of the main head for the 2B Dense and A1B MoE models across benchmarks using different MTP methods.}
\label{tab:distill_acc_mainhead}
\end{table}

\subsection{Details of Ablation Experiments}
In this subsection, we provide a detailed analysis of the ablation experiments for MTP-D with 1 and 4 MTP heads.

\subsubsection{1 MTP head}
\begin{figure}[h]
\begin{subfigure}{0.49\textwidth}
    \centering
    \includegraphics[width=\linewidth]{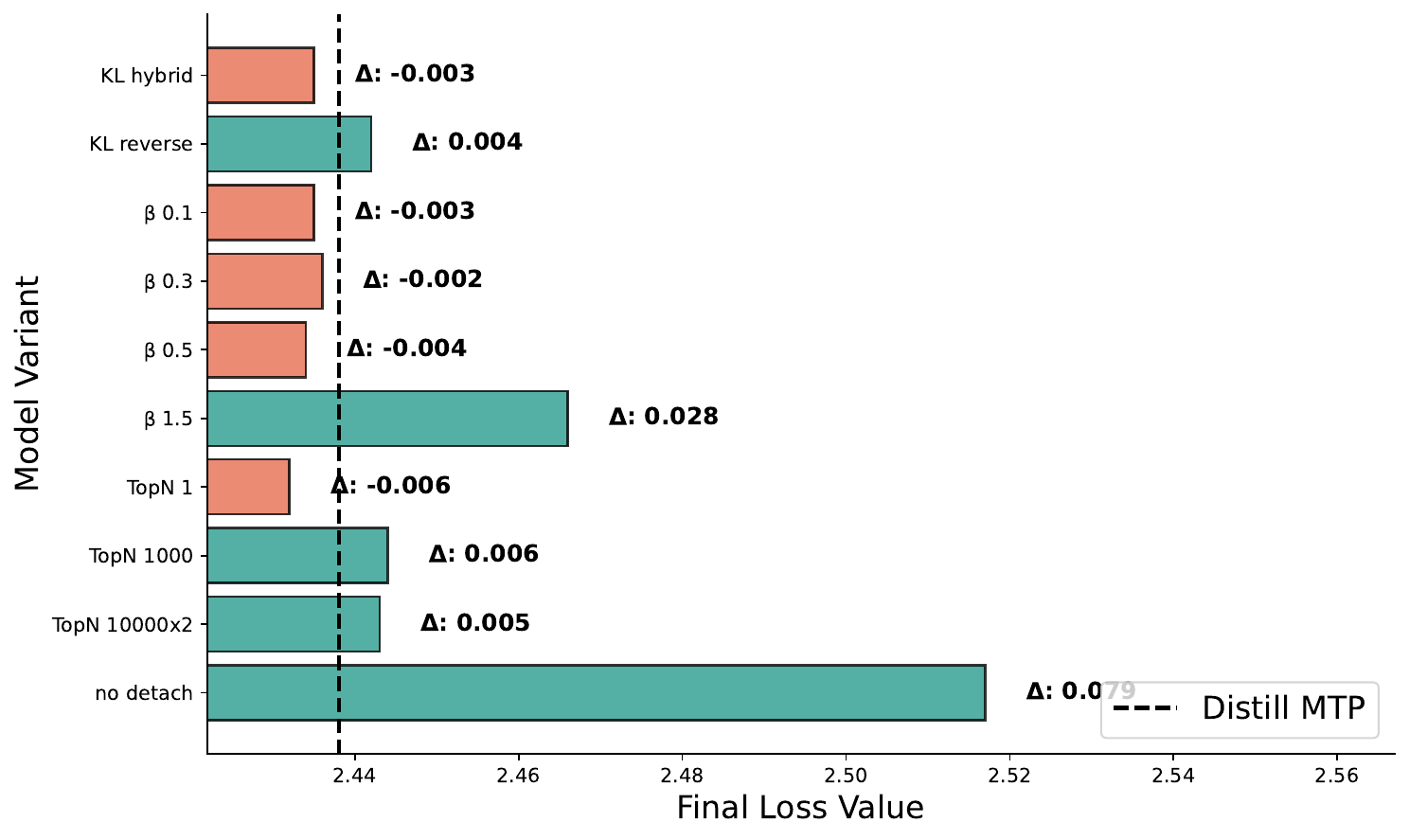}
    \subcaption{Loss comparison of main head.}
\end{subfigure}
\hfill
\begin{subfigure}{0.49\textwidth}
    \centering
    \includegraphics[width=\linewidth]{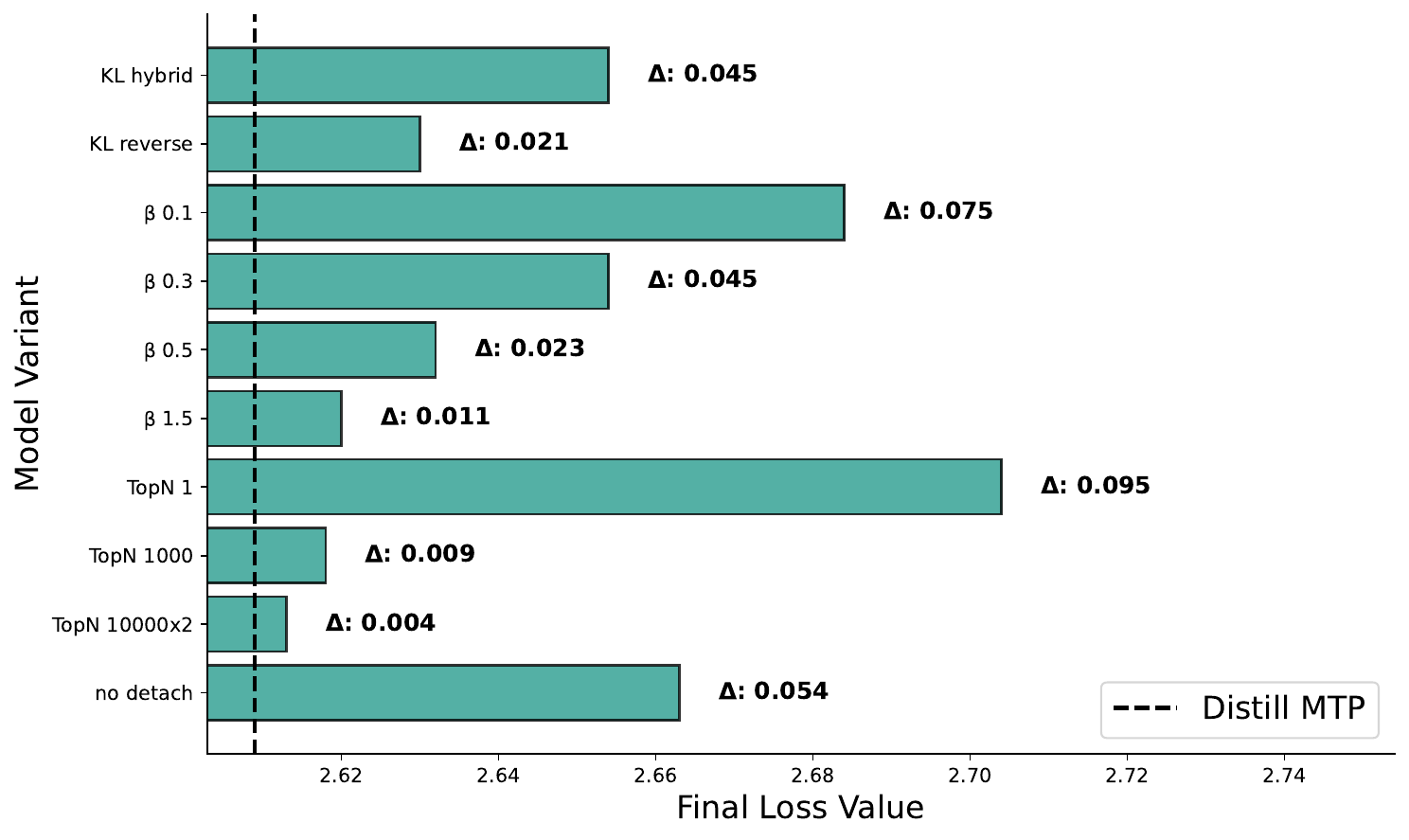}
    \subcaption{Loss comparison of MTP head.}
\end{subfigure}
\caption{Ablation experiments for MTP-D and its variants with loss as metrics. Orange indicates losses better than MTP-D, while green indicates worse performance.}
\label{fig:ablation_loss_compare}
\end{figure}

Loss serves as an important metric for pre-training evaluation. Figure~\ref{fig:ablation_loss_compare} presents the final loss comparison of different variants of MTP-D with 1 MTP head, including both main-head and MTP-head losses. While maintaining comparable main-head loss, our MTP-D achieves a substantially lower MTP-head loss than the other variants.

\begin{figure}[h]
    \centering
    \includegraphics[width=0.99\linewidth]{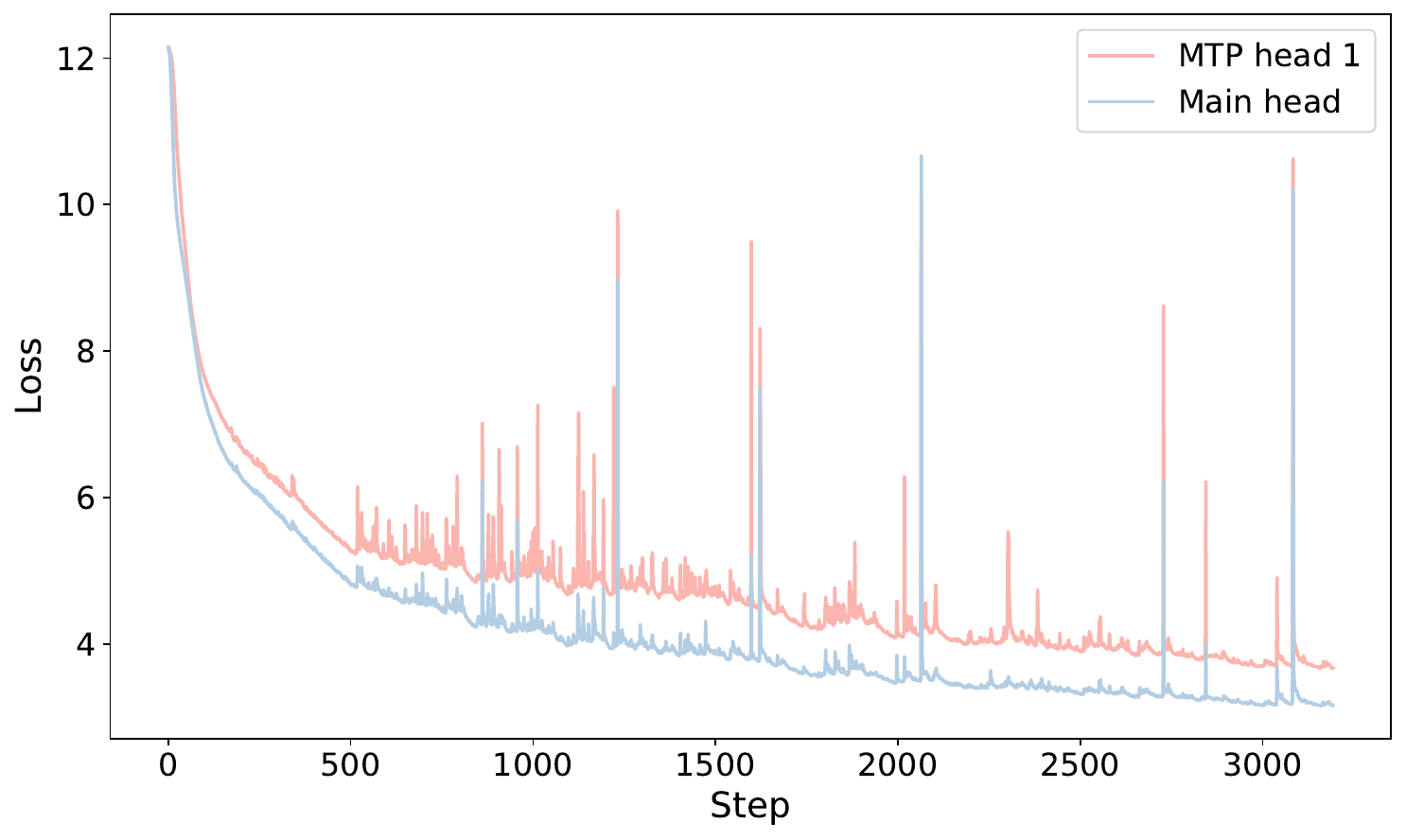}
    \caption{Training loss curves for the 2B Dense model with 1 head using the whole vocabulary.}
    \label{fig:illustration_of_loss_detach}
\end{figure}

In addition, the loss curves for distillation over the entire vocabulary (Top$N$) are shown in Figure~\ref{fig:illustration_of_loss_detach}. Due to the long-tailed distribution of logits across the vocabulary, logarithmic operations can introduce numerical instability, potentially causing gradient oscillation or vanishing.

\subsubsection{4 MTP heads}
\label{appendix:4head_ablation}
We provide additional ablation experiments for MTP-D with 4 heads, analyzing the multi-level distillation strategies and the corresponding $\beta_k$ values.

For MTP-D with 4 heads, the hyperparameters are set as $\beta_k=0.5$ (fixed) with main-to-MTP distillation. We first conduct a hyperparameter study on $\beta_k$. As shown in Table~\ref{tab:distill_ablation_4head}, main-head accuracy initially increases and then decreases with $\beta_k$, reaching its peak at 0.3. However, Table~\ref{tab:distill_ablation_4head_mtp} shows that the CAR and AR of MTP heads are relatively low for $\beta_k=0.3$. We further explore a dynamic training strategy where $\alpha_k$ and $\beta_k$ vary with training steps: for steps $<2000$, $\alpha_k$ decreases linearly from 0.7 to 0.3, and $\beta_k$ increases linearly from 0.1 to 0.5, intending for CrossEntropy loss to dominate early training and KL loss to dominate once the main head has sufficient capability. The results, however, do not meet expectations. 

Additionally, we investigate more complex distillation strategies based on MTP-D: ``ensemble mean'' and ``ensemble split''. In ``ensemble mean'', the KL teacher for the $k$-th MTP head is the weighted average of the logits from MTP$_{0:k-1}$ and the main head. In ``ensemble split'', the KL teacher for the $k$-th MTP head consists of multiple logits from MTP$_{0:k}$ and the main head, with the losses weighted accordingly. The results indicate that using MTP heads as KL distillation teachers provides more diverse supervision signals, improving both the main model performance and the acceptance rates of MTP heads.

\begin{table}[h]
\centering
\small
\begin{tabular}{l c}
\toprule
\textbf{Strategy} & \textbf{Mean} \\
\midrule
MTP-D & 10.96 \\
\midrule
$\beta_k = 0.1$ & 10.93 \\
$\beta_k = 0.3$ & \textbf{11.13} \\
$\beta_k = 1.0$ & \underline{10.62} \\
\midrule
step weights & \underline{9.39} \\
\midrule
ensemble mean & \textbf{11.22} \\
ensemble split & 11.20 \\
\bottomrule
\end{tabular}
\caption{Main-head mean accuracy of ablation experiments for MTP-D with 4 heads.}
\label{tab:distill_ablation_4head}
\end{table}

\begin{table*}[h]
\centering
\small
\setlength{\tabcolsep}{2.4pt}
\begin{tabular}{l c c c c c c c c c}
\toprule
& & &
\multicolumn{1}{c}{\textbf{General}} &
\multicolumn{2}{c}{\textbf{Math}} &
\multicolumn{3}{c}{\textbf{Knowledge}} &
\multicolumn{1}{c}{\textbf{STEM}} \\
\cmidrule(lr){4-4}
\cmidrule(lr){5-6}
\cmidrule(lr){7-9}
\cmidrule(lr){10-10}
\multicolumn{2}{c}{\textbf{Strategy}} &
\textbf{H} &
\shortstack{\textbf{AGIEval} \\ \textbf{en}} &
\shortstack{\textbf{GSM8K}} &
\textbf{MATH} &
\shortstack{\textbf{Natural} \\ \textbf{Questions}} &
\shortstack{\textbf{Simple} \\ \textbf{QA}} &
\textbf{TriviaQA} &
\shortstack{\textbf{Super} \\ \textbf{GPQA}} \\
\midrule

\multirow{4}{*}{\textbf{MTP-D}}& & 1 &
85.86 / 85.86 &
85.20 / 85.20 &
86.20 / 86.20 &
91.69 / 91.69 &
90.16 / 90.16 &
84.73 / 84.73 &
87.77 / 87.77 \\

& & 2 &
71.96 / 83.81 &
70.97 / 83.29 &
71.96 / 83.48 &
84.63 / 92.30 &
81.15 / 90.01 &
69.46 / 81.98 &
71.16 / 84.00 \\

& & 3 &
61.25 / 85.12 &
57.57 / 81.12 &
62.57 / 86.95 &
78.33 / 92.55 &
74.71 / 92.06 &
56.75 / 81.70 &
58.57 / 82.31 \\

& & 4 &
52.96 / 86.46 &
46.27 / 80.38 &
54.98 / 87.94 &
72.76 / 92.89 &
71.22 / 95.34 &
46.42 / 81.81 &
48.52 / 82.85 \\
\midrule
\multirow{4}{*}{\textbf{$\beta_k=0.3$ }}& & 1 &
83.57 / 83.57 &
85.42 / 85.42 &
83.38 / 83.38 &
91.39 / 91.39 &
86.67 / 86.67 &
82.24 / 82.24 &
83.86 / 83.86 \\
& & 2 &
70.44 / 84.30 &
70.25 / 82.24 &
69.96 / 83.91 &
83.90 / 91.81 &
78.91 / 91.04 &
67.02 / 81.49 &
70.05 / 83.54 \\
& & 3 &
60.10 / 85.31 &
57.33 / 81.61 &
60.50 / 86.47 &
75.87 / 90.43 &
72.06 / 91.32 &
52.87 / 78.89 &
57.56 / 82.16 \\
& & 4 &
51.85 / 86.28 &
46.17 / 80.54 &
52.76 / 87.21 &
69.91 / 92.14 &
67.67 / 93.90 &
42.03 / 79.51 &
47.37 / 82.30 \\
\midrule
\multirow{4}{*}{ensemble mean}& & 1 &
86.15 / 86.15 &
86.26 / 86.26 &
87.05 / 87.05 &
90.93 / 90.93 &
86.29 / 86.29 &
84.99 / 84.99 &
84.72 / 84.72 \\
& & 2 &
73.43 / 85.24 &
72.44 / 83.97 &
73.43 / 84.35 &
83.30 / 91.61 &
81.12 / 94.00 &
70.06 / 82.43 &
71.55 / 84.46 \\
& & 3 &
62.84 / 85.58 &
59.91 / 82.71 &
65.00 / 88.52 &
75.86 / 91.07 &
77.21 / 95.17 &
57.07 / 81.46 &
59.47 / 83.11 \\
& & 4 &
54.41 / 86.59 &
49.18 / 82.25 &
58.25 / 89.63 &
69.69 / 91.86 &
73.14 / 94.73 &
46.57 / 81.60 &
49.44 / 83.15 \\

\bottomrule
\end{tabular}
\caption{MTP-head AR and CAR for MTP-D with 4 heads across benchmarks, comparing three groups with similar main-head performance.}
\label{tab:distill_ablation_4head_mtp}
\end{table*}

\section{Detailed Analysis of Looped MTP}
\label{appendix_loop_trainfree}

In this section, we provide a detailed analysis of acceptance rates and cumulative acceptance rates across multiple pre-training benchmarks for looped MTP, considering both training-free and continued pre-training extensions. 

\subsection{Training-Free Looped MTP}
\begin{figure*}[h]
\begin{subfigure}{0.24\textwidth}
    \centering
    \includegraphics[width=\linewidth]{figs/train_free/AGI-Eval-en_cumulative_acceptance_cumulative.pdf}
    \subcaption{AGIEval-en.}
\end{subfigure}
\hfill
\begin{subfigure}{0.24\textwidth}
    \centering
    \includegraphics[width=\linewidth]{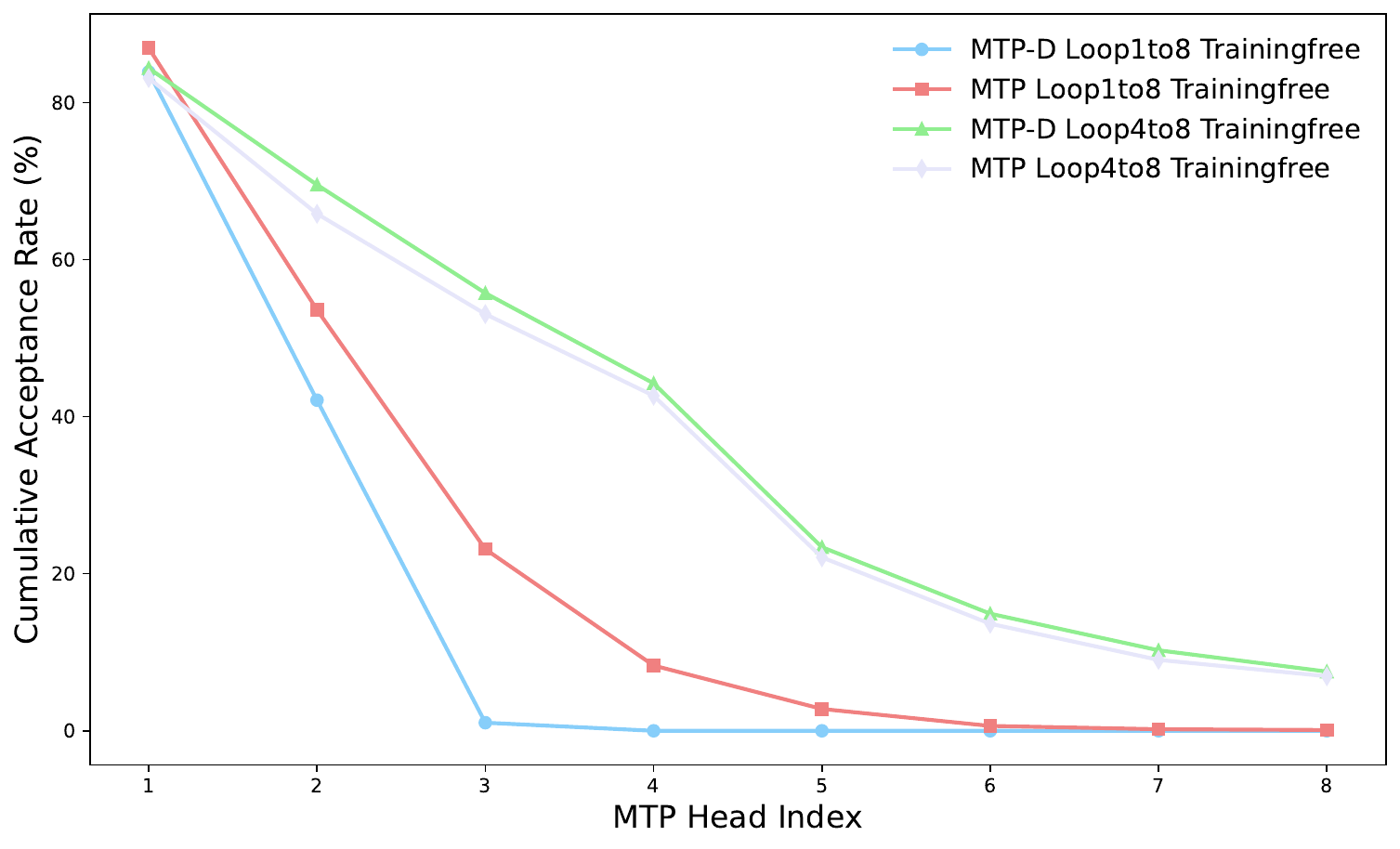}
    \subcaption{GSM8K.}
\end{subfigure}
\hfill
\begin{subfigure}{0.24\textwidth}
    \centering
    \includegraphics[width=\linewidth]{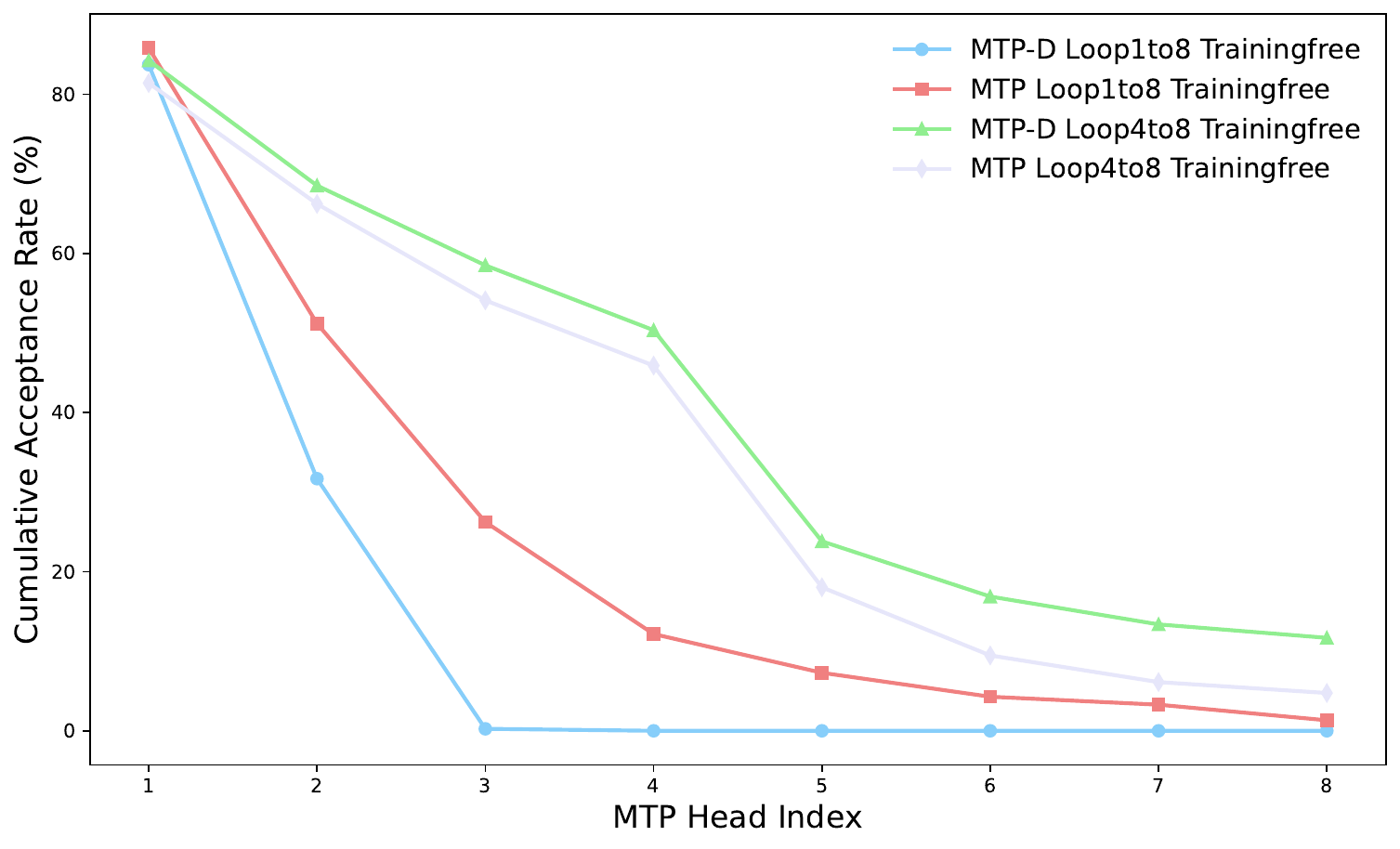}
    \subcaption{MATH.}
\end{subfigure}
\hfill
\begin{subfigure}{0.24\textwidth}
    \centering
    \includegraphics[width=\linewidth]{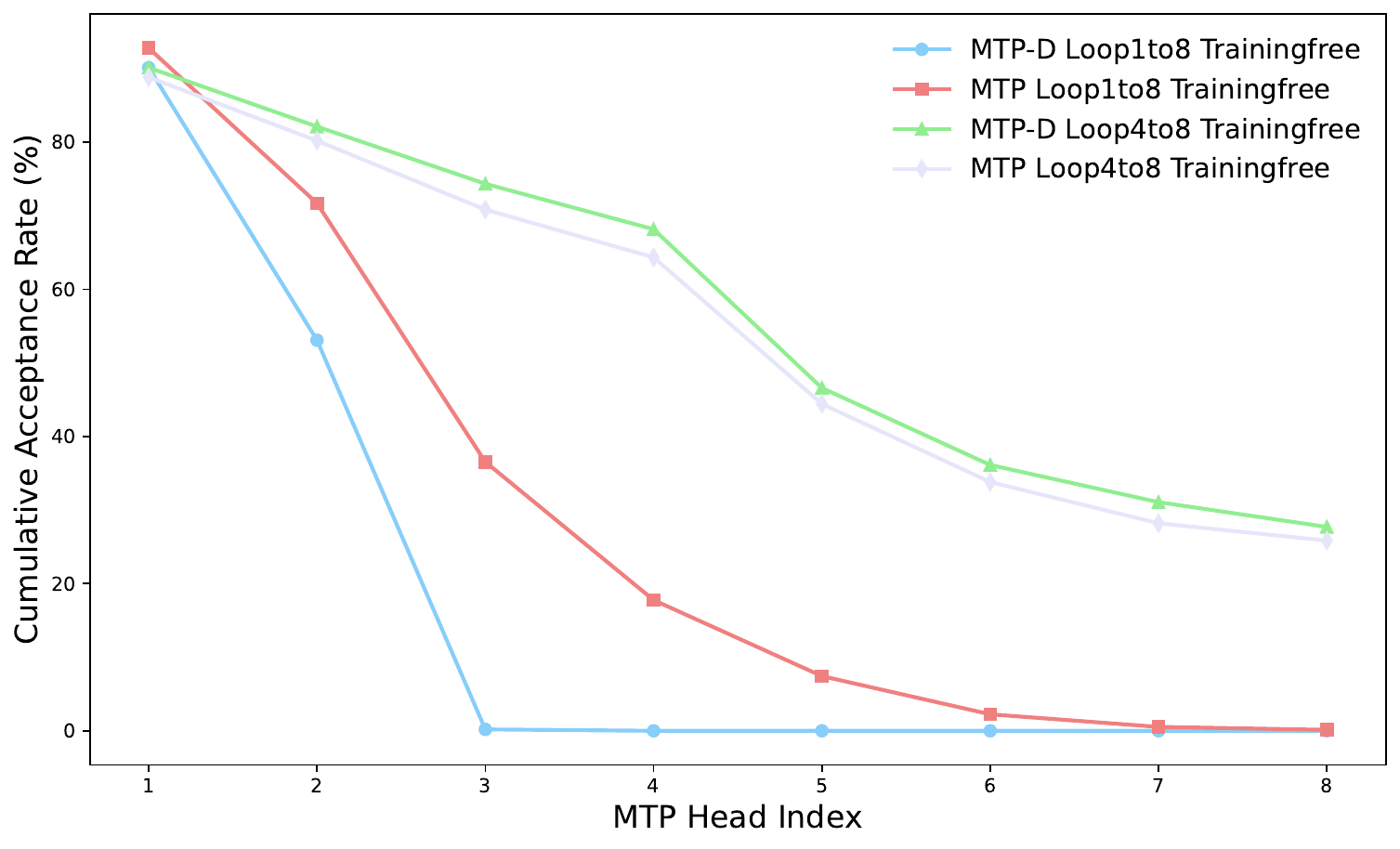}
    \subcaption{NaturalQuestions.}
\end{subfigure}
\hfill
\begin{subfigure}{0.24\textwidth}
    \centering
    \includegraphics[width=\linewidth]{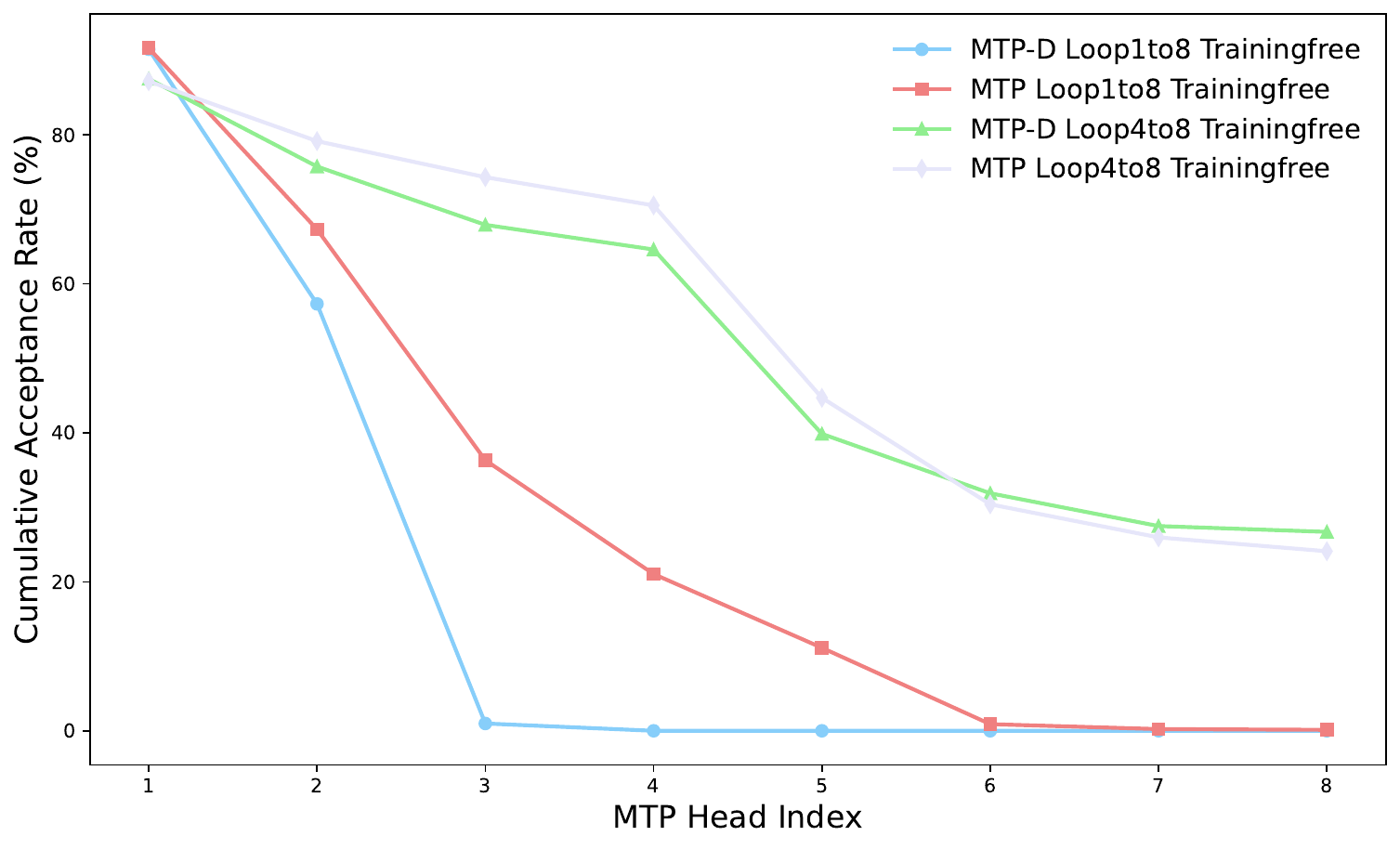}
    \subcaption{SimpleQA.}
\end{subfigure}
\hfill
\begin{subfigure}{0.24\textwidth}
    \centering
    \includegraphics[width=\linewidth]{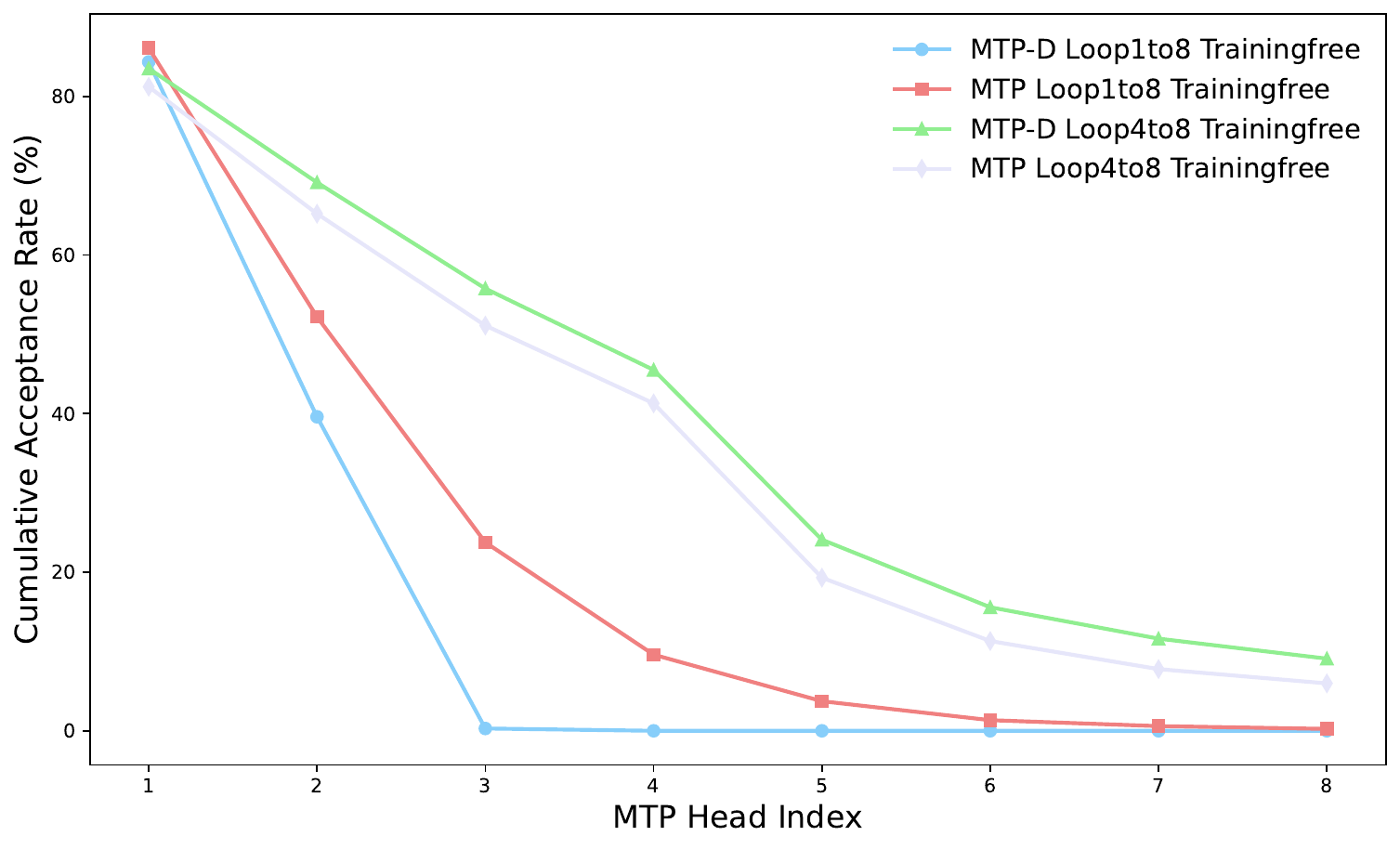}
    \subcaption{SuperGPQA.}
\end{subfigure}
\hfill
\begin{subfigure}{0.24\textwidth}
    \centering
    \includegraphics[width=\linewidth]{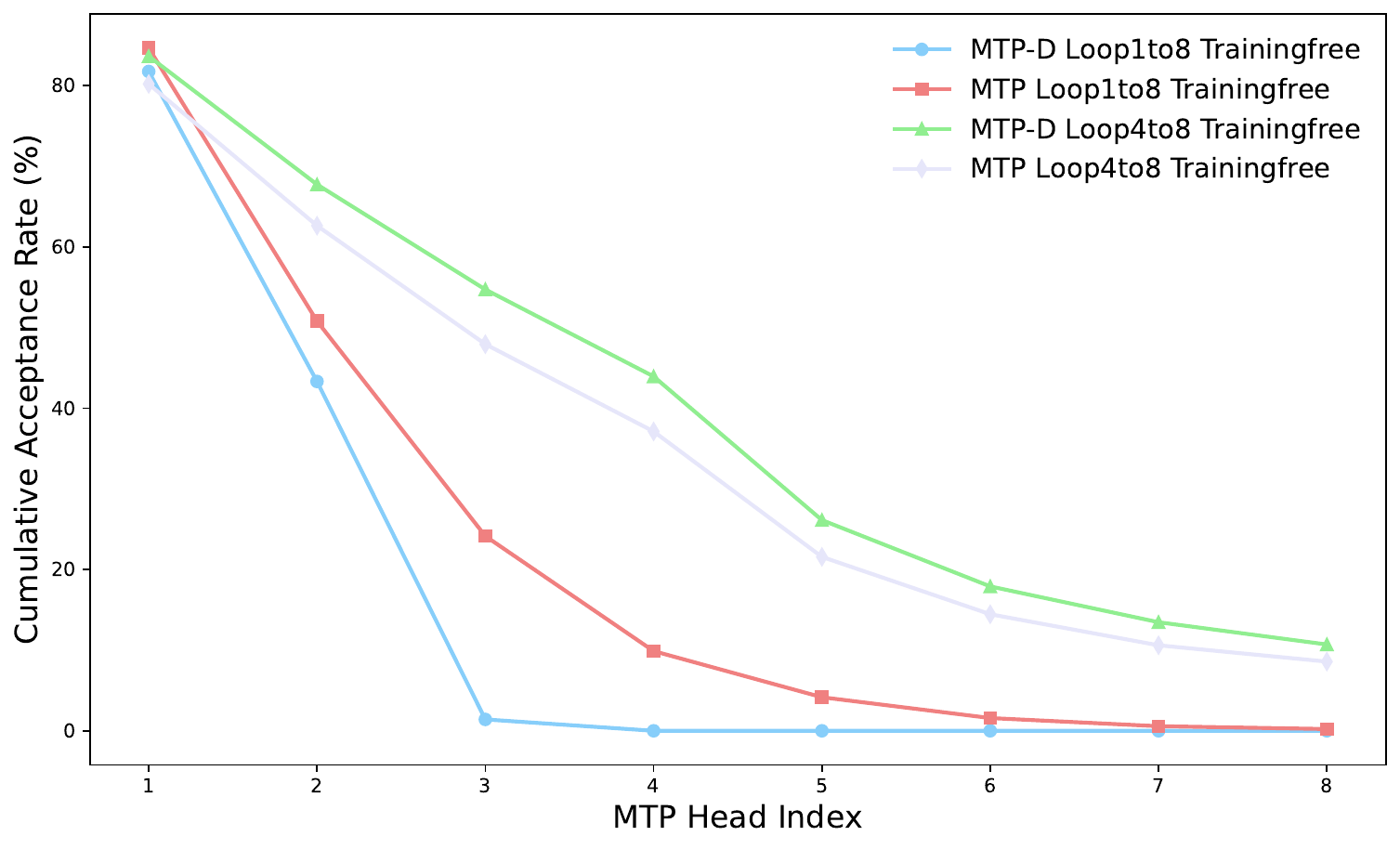}
    \subcaption{TriviaQA.}
\end{subfigure}
\hfill
\caption{Cumulative acceptance rates on multiple pretraining benchmarks with the training-free looped MTP scaled up to 8.}
\label{fig:looptrainfree_allbench_cumulative}
\end{figure*}

\begin{figure*}[h]
\begin{subfigure}{0.24\textwidth}
    \centering
    \includegraphics[width=\linewidth]{figs/train_free/AGI-Eval-en_current_acceptance_bar.pdf}
    \subcaption{AGIEval-en.}
\end{subfigure}
\hfill
\begin{subfigure}{0.24\textwidth}
    \centering
    \includegraphics[width=\linewidth]{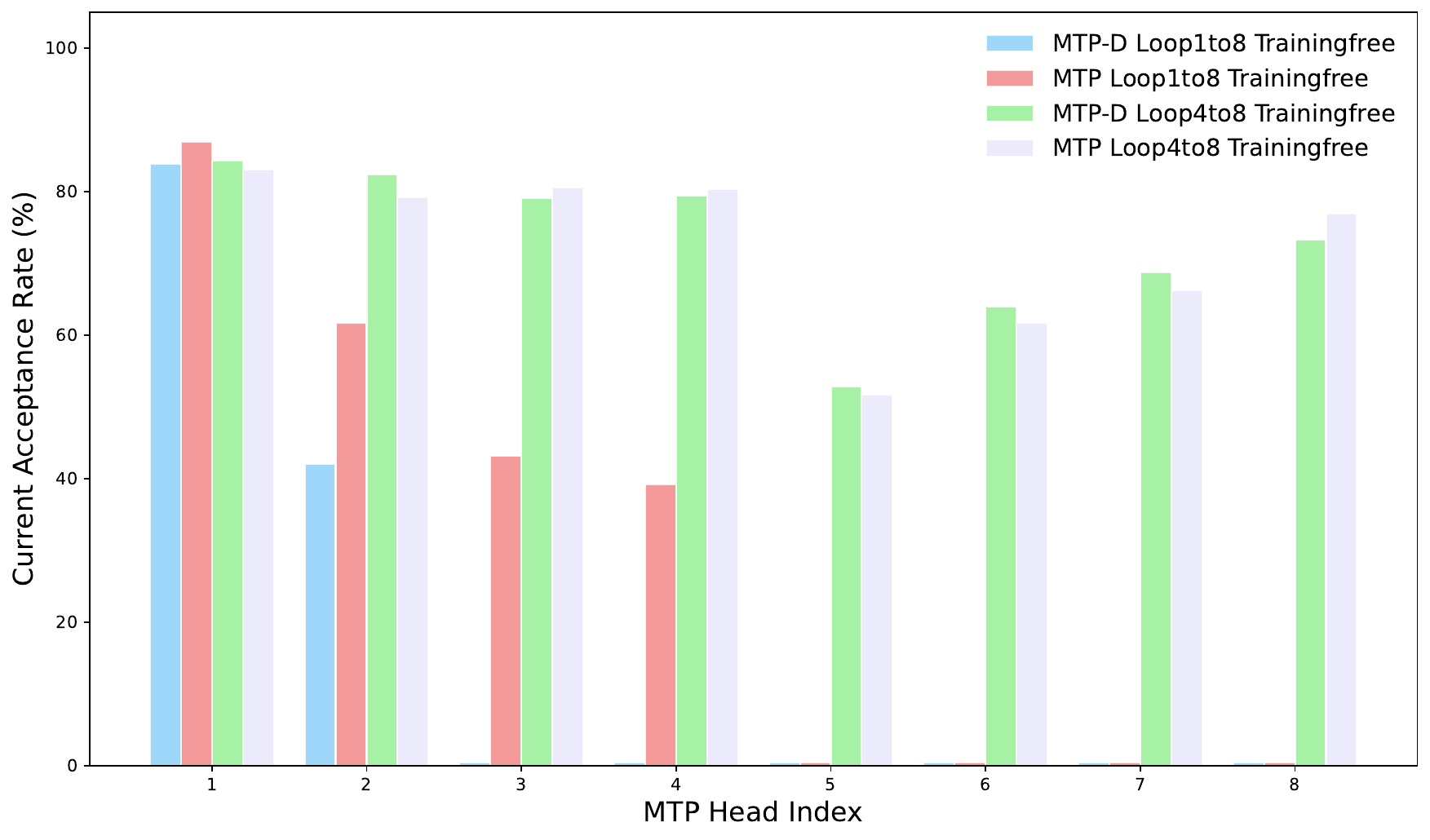}
    \subcaption{GSM8K.}
\end{subfigure}
\hfill
\begin{subfigure}{0.24\textwidth}
    \centering
    \includegraphics[width=\linewidth]{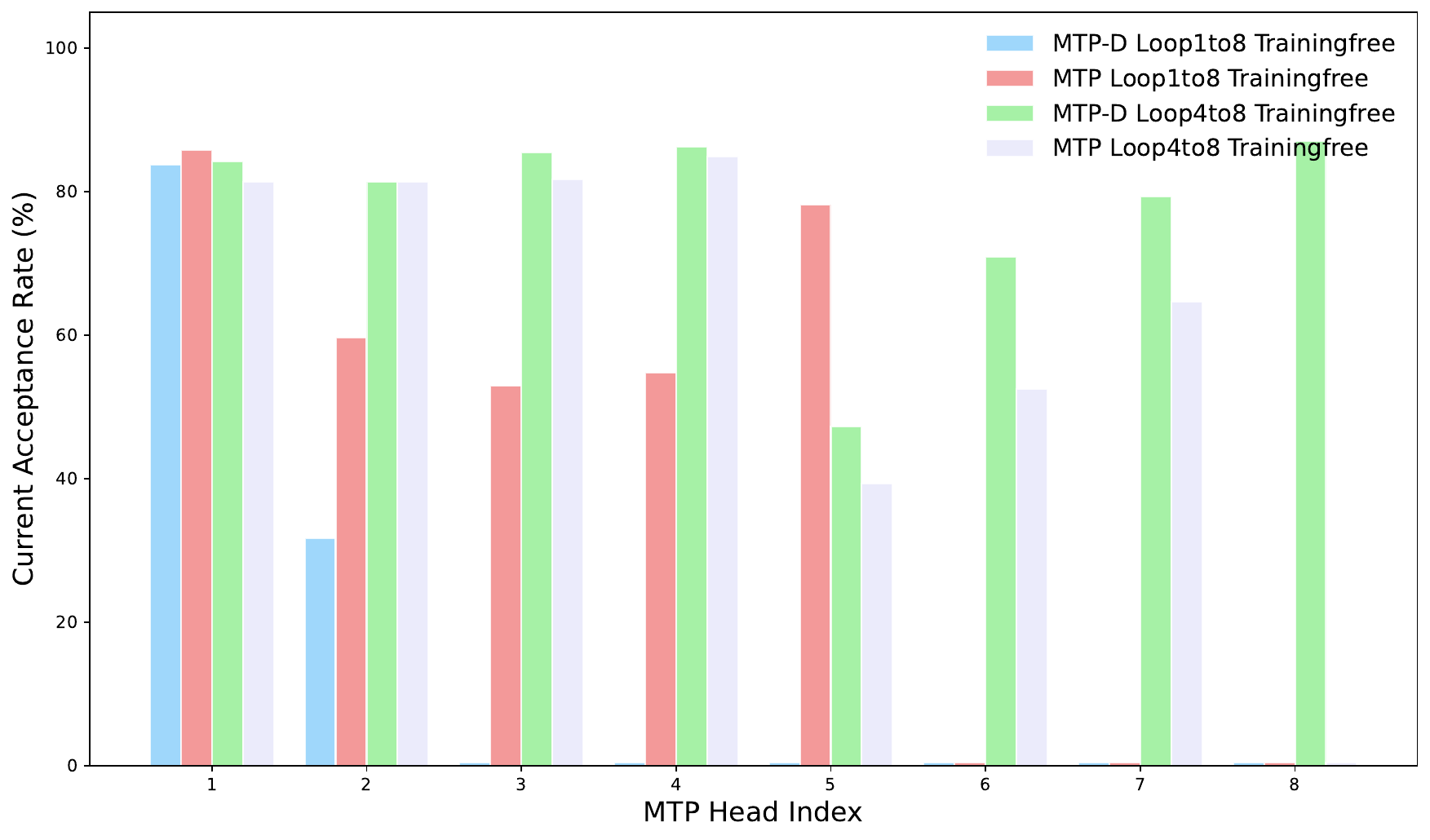}
    \subcaption{MATH.}
\end{subfigure}
\hfill
\begin{subfigure}{0.24\textwidth}
    \centering
    \includegraphics[width=\linewidth]{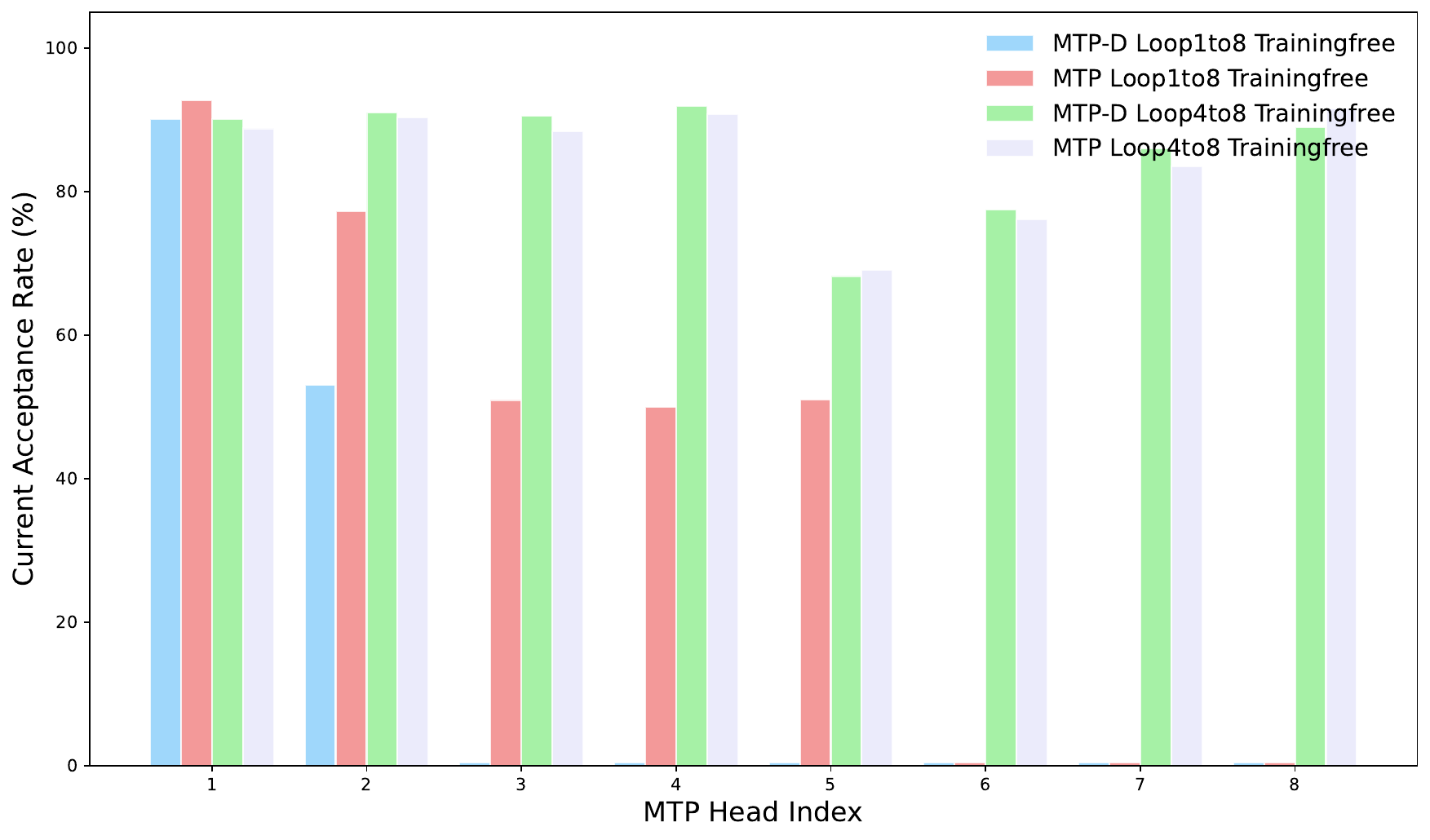}
    \subcaption{NaturalQuestions.}
\end{subfigure}
\hfill
\begin{subfigure}{0.24\textwidth}
    \centering
    \includegraphics[width=\linewidth]{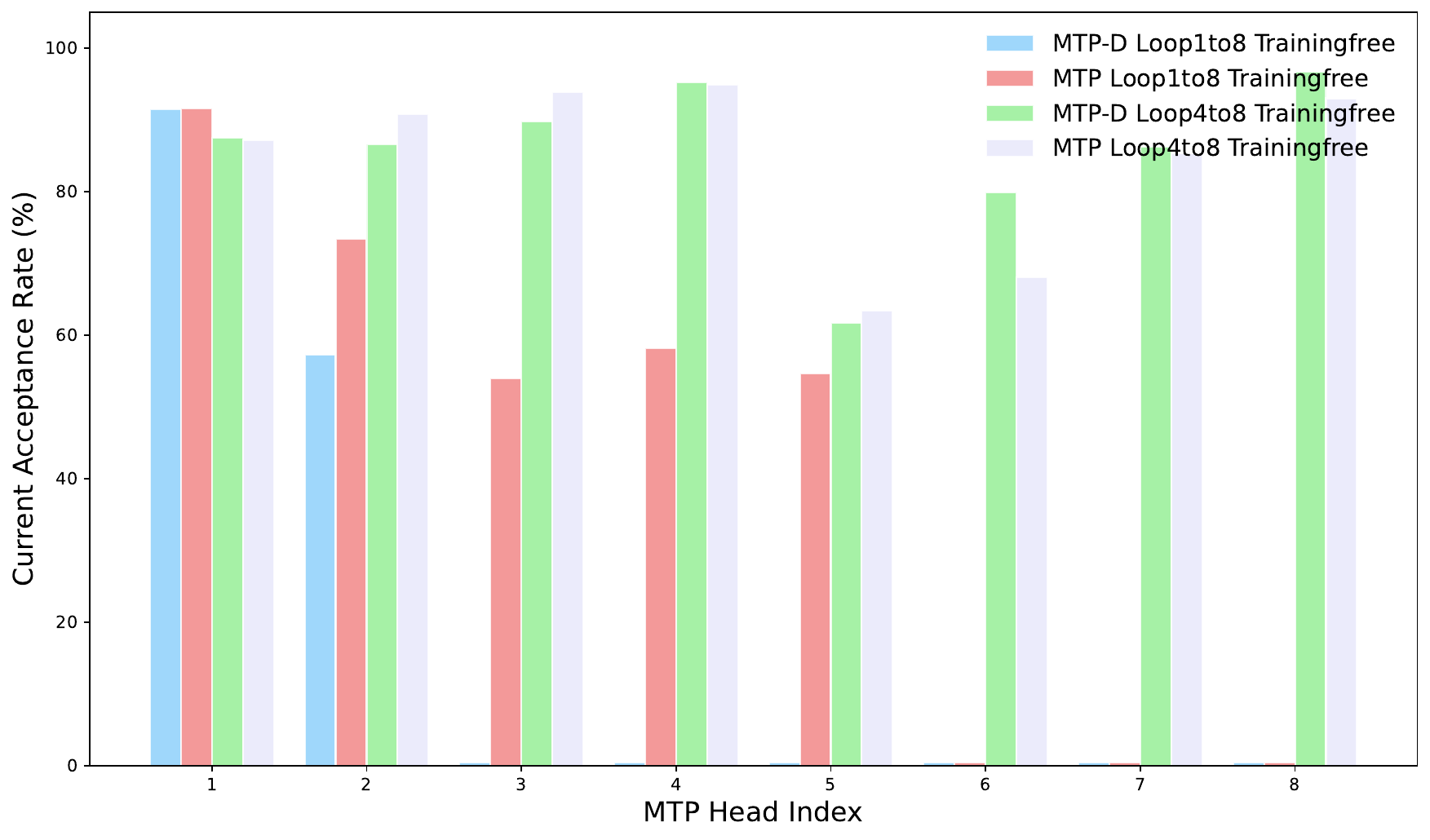}
    \subcaption{SimpleQA.}
\end{subfigure}
\hfill
\begin{subfigure}{0.24\textwidth}
    \centering
    \includegraphics[width=\linewidth]{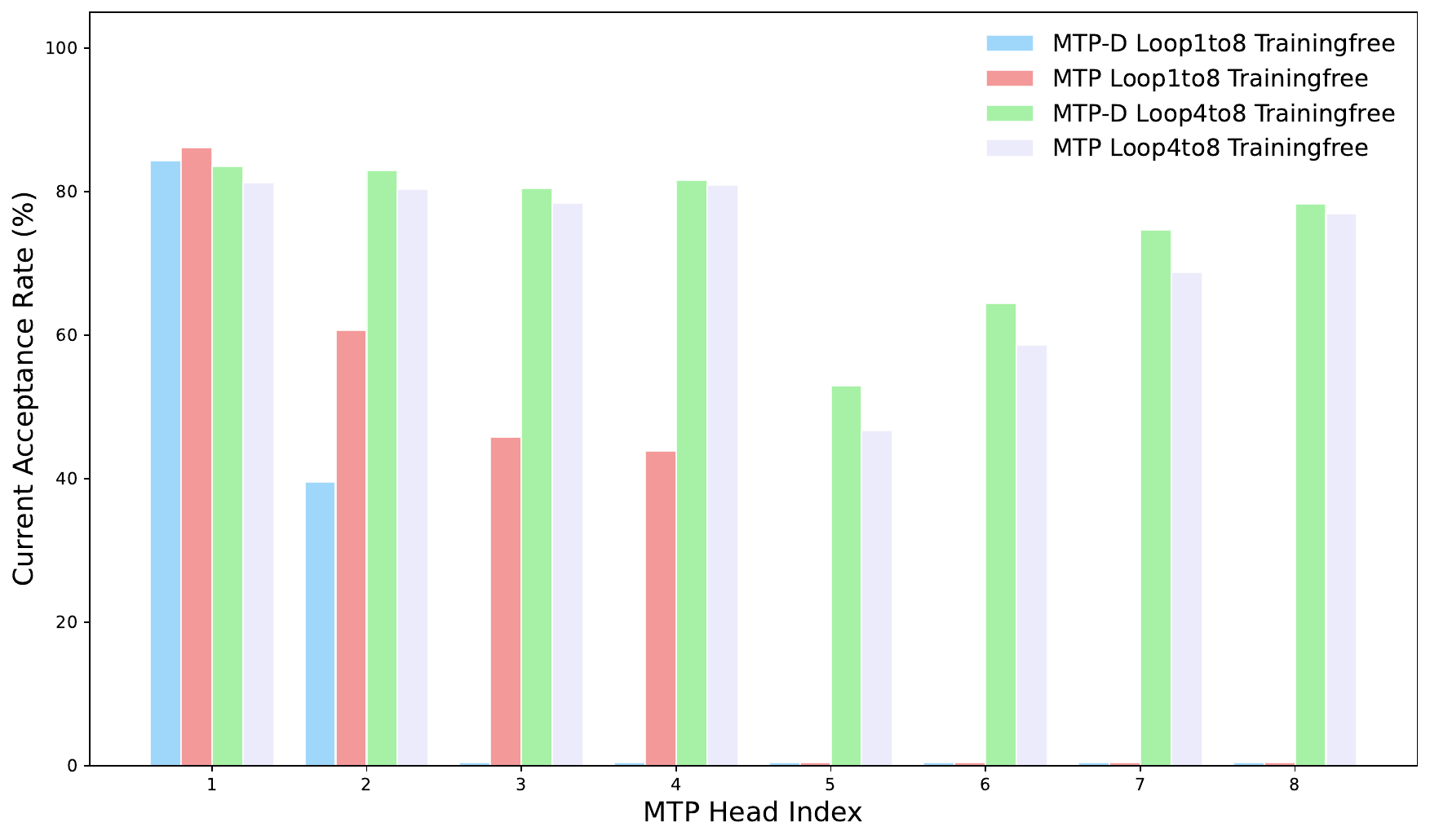}
    \subcaption{SuperGPQA.}
\end{subfigure}
\hfill
\begin{subfigure}{0.24\textwidth}
    \centering
    \includegraphics[width=\linewidth]{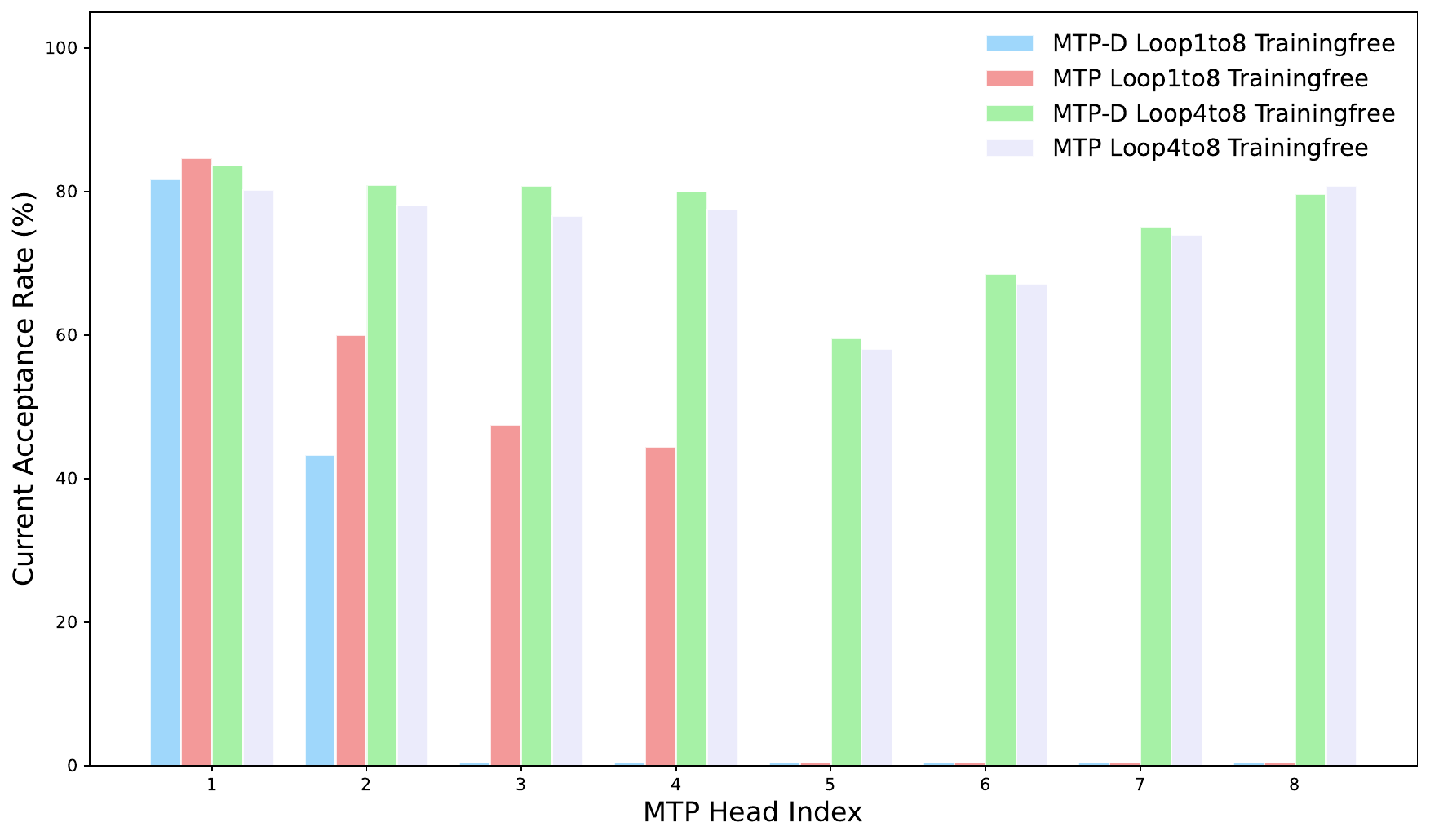}
    \subcaption{TriviaQA.}
\end{subfigure}
\hfill
\caption{Acceptance rates on multiple pretraining benchmarks with the training-free  looped MTP scaled up to 8.}
\label{fig:looptrainfree_allbench}
\end{figure*}

Figures~\ref{fig:looptrainfree_allbench_cumulative} and \ref{fig:looptrainfree_allbench} show CAR and AR for MTP head loops up to 8 under a training-free setting. 

First, MTP heads located at the loop connection points exhibit a noticeable drop in acceptance rate compared to other heads. Taking AGIEval-en for example, the acceptance rate of the 4-to-8 looped MTP drops sharply from approximately 80\% at MTP head~4 to around 50\% at MTP head~5. Despite this degradation, the acceptance rates still remain at an acceptable level. Moreover, as the MTP heads scale up from 5 to 8, their acceptance rates gradually increase and approach those of the trained MTP heads~1 to~4. A similar trend is observed for the 1-to-8 looped MTP. These results indicate that the cascaded architecture of DeepSeek MTP, together with its structural consistency and input-output similarity, inherently supports scalability.

Furthermore, a comparison between our MTP-D and the DeepSeek MTP under the 1-to-8 loop scaling up setting reveals that MTP heads trained with the self-distillation paradigm exhibit substantially superior scalability. As shown in Figure~\ref{fig:accrate_loop_trainfree}(a), the cumulative acceptance rate of the DeepSeek MTP drops to 0.6\% when loop-scaled up to MTP head~3. In contrast, our MTP-D maintains a cumulative acceptance rate of 26.70\% at MTP head~3, enabling it to be further scaled up to a larger number of MTP heads. These results demonstrate that our proposed MTP-D significantly enhances the consistency of output distributions across MTP heads with main head, thereby endowing our MTP-D with markedly improved scalability. 

\subsection{Continued Pre-trained Looped MTP}
\begin{figure*}[h]
\begin{subfigure}{0.24\textwidth}
    \centering
    \includegraphics[width=\linewidth]{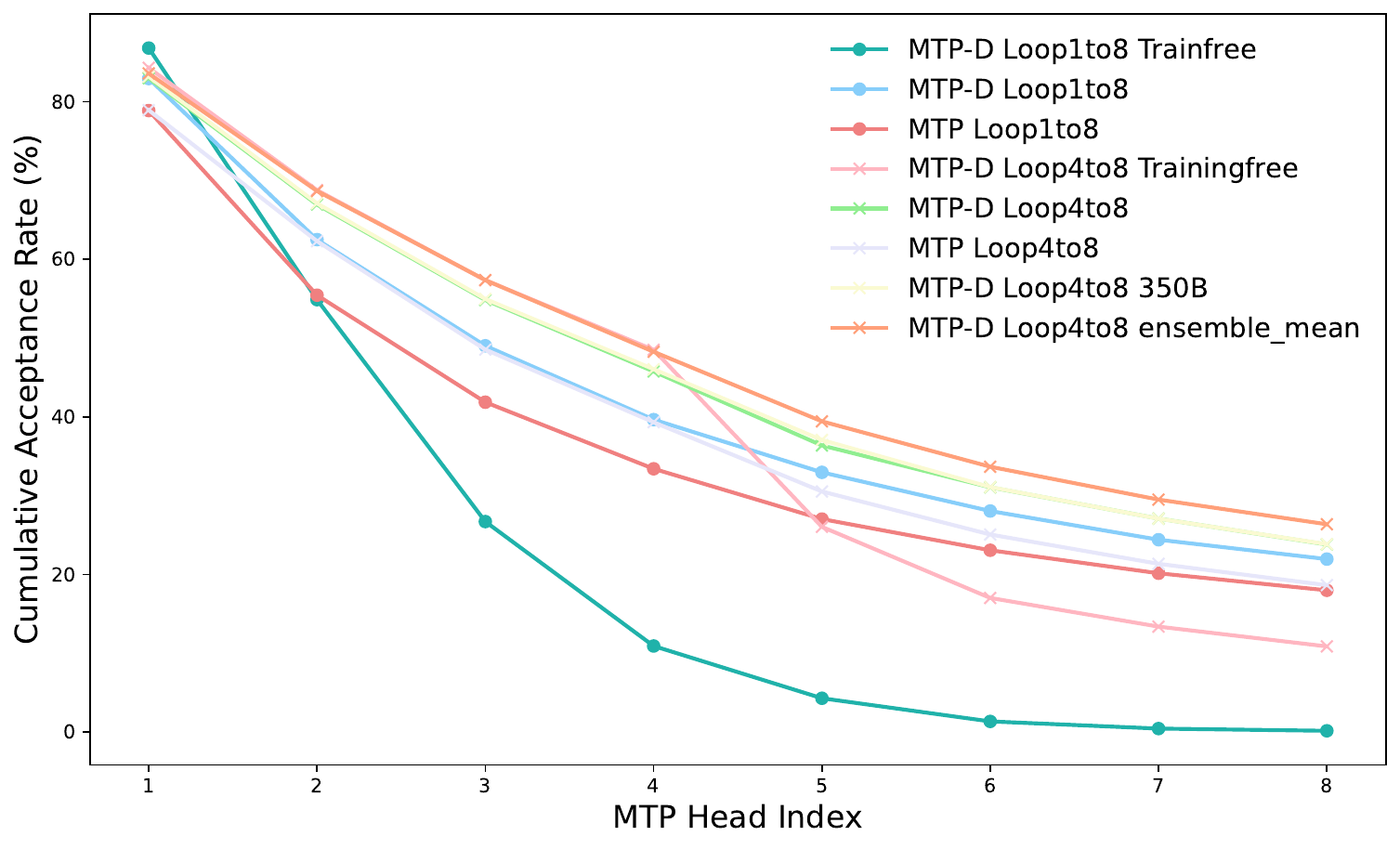}
    \subcaption{AGIEval-en.}
\end{subfigure}
\hfill
\begin{subfigure}{0.24\textwidth}
    \centering
    \includegraphics[width=\linewidth]{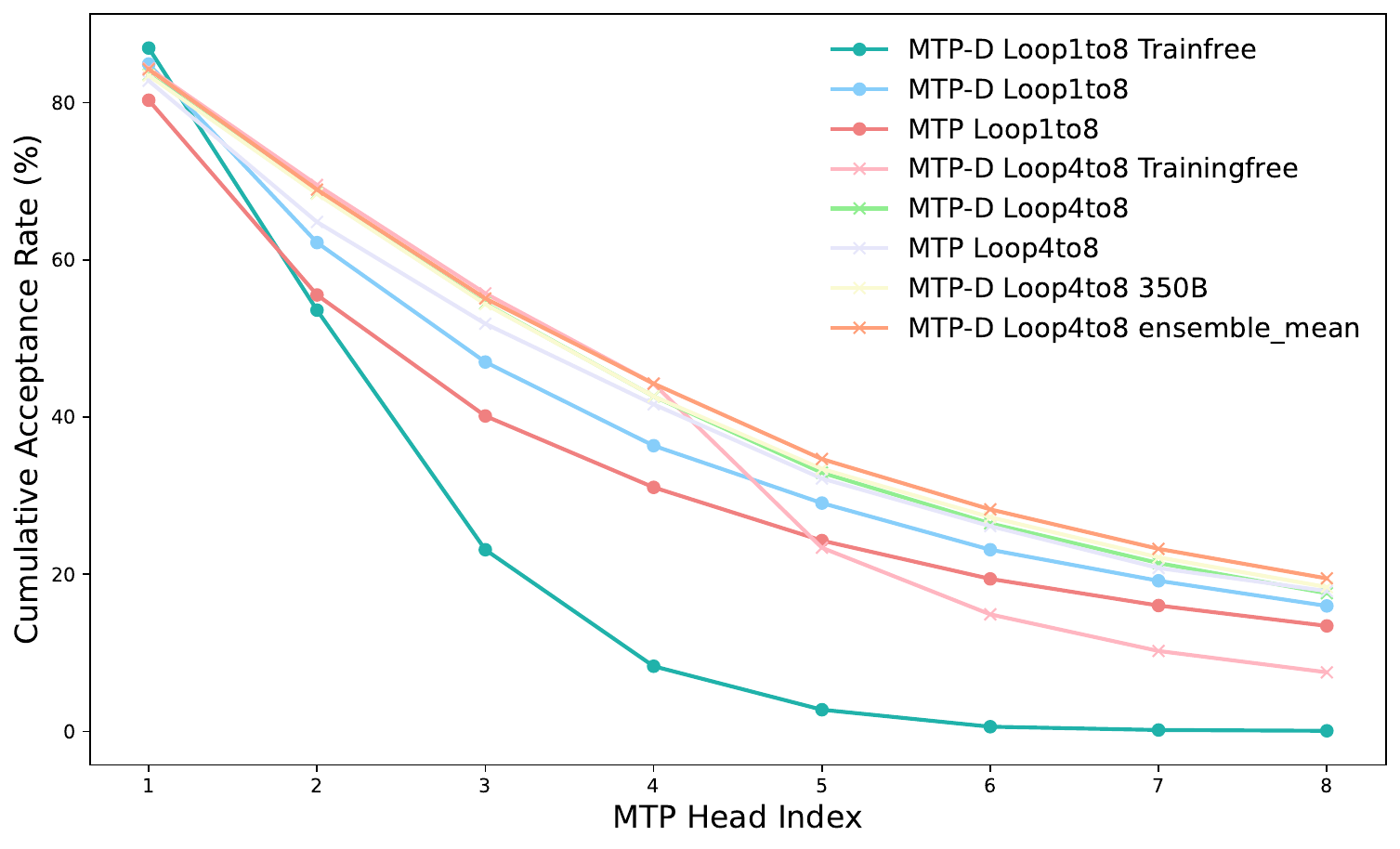}
    \subcaption{GSM8K.}
\end{subfigure}
\hfill
\begin{subfigure}{0.24\textwidth}
    \centering
    \includegraphics[width=\linewidth]{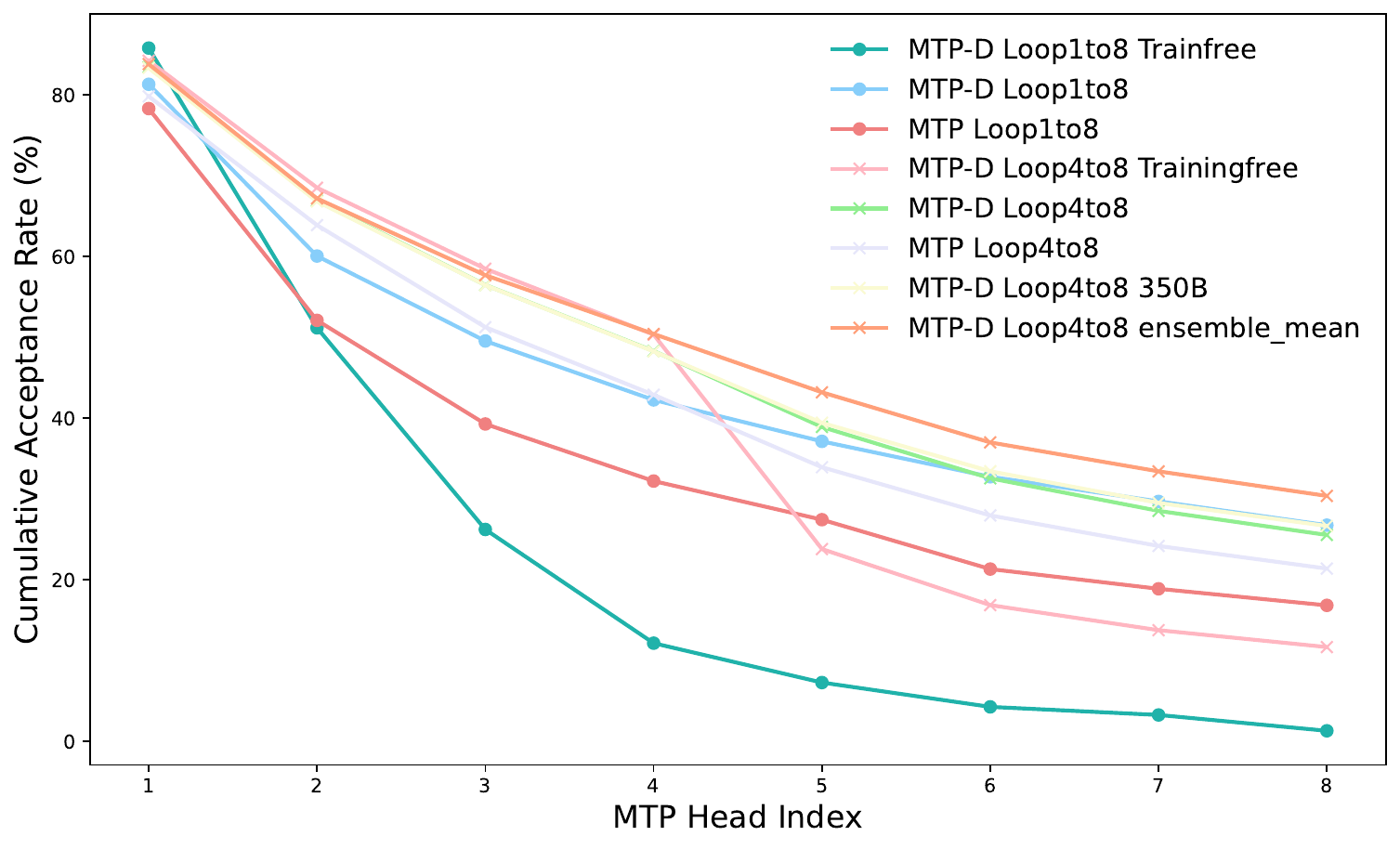}
    \subcaption{MATH.}
\end{subfigure}
\hfill
\begin{subfigure}{0.24\textwidth}
    \centering
    \includegraphics[width=\linewidth]{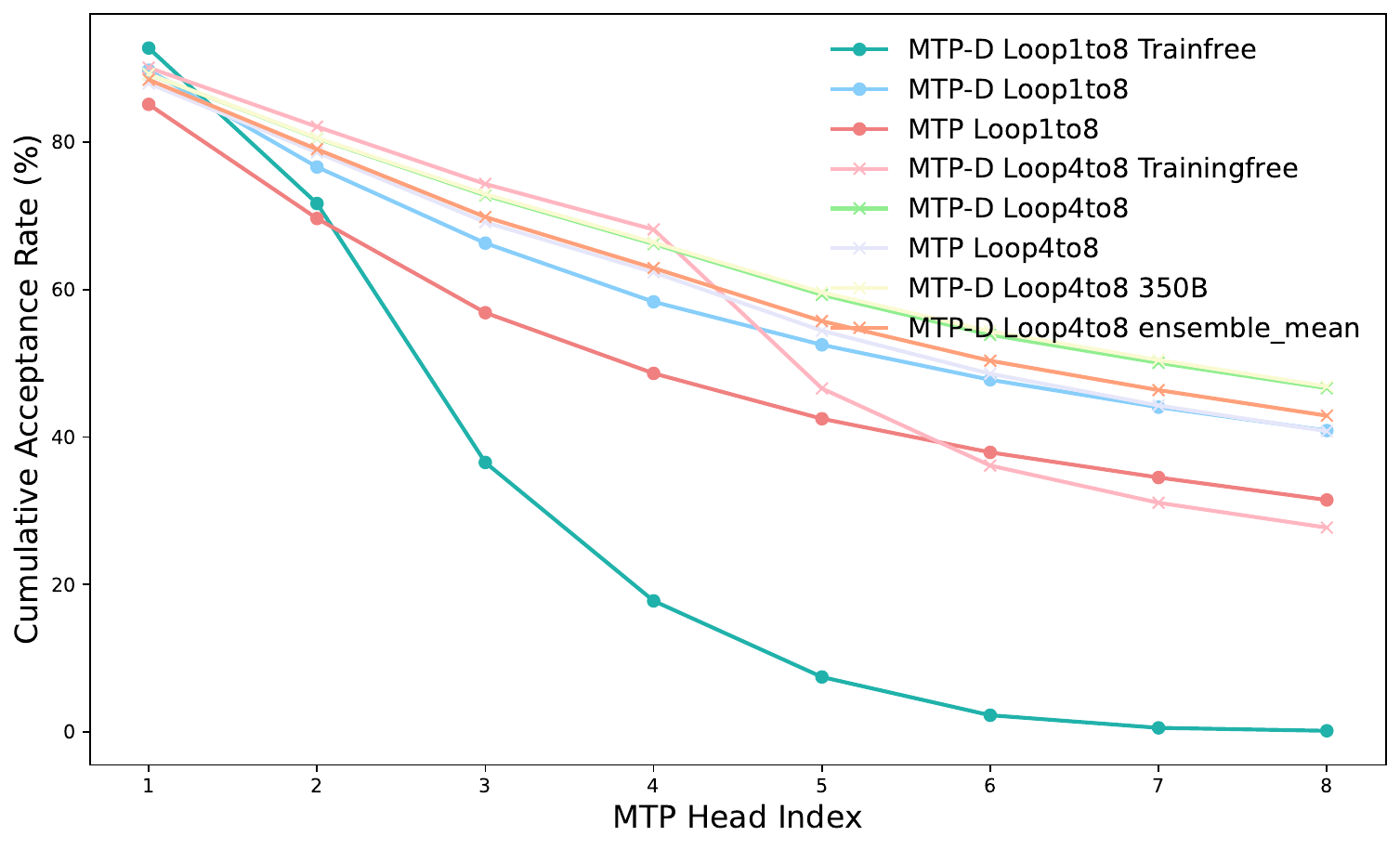}
    \subcaption{NaturalQuestions.}
\end{subfigure}
\hfill
\begin{subfigure}{0.24\textwidth}
    \centering
    \includegraphics[width=\linewidth]{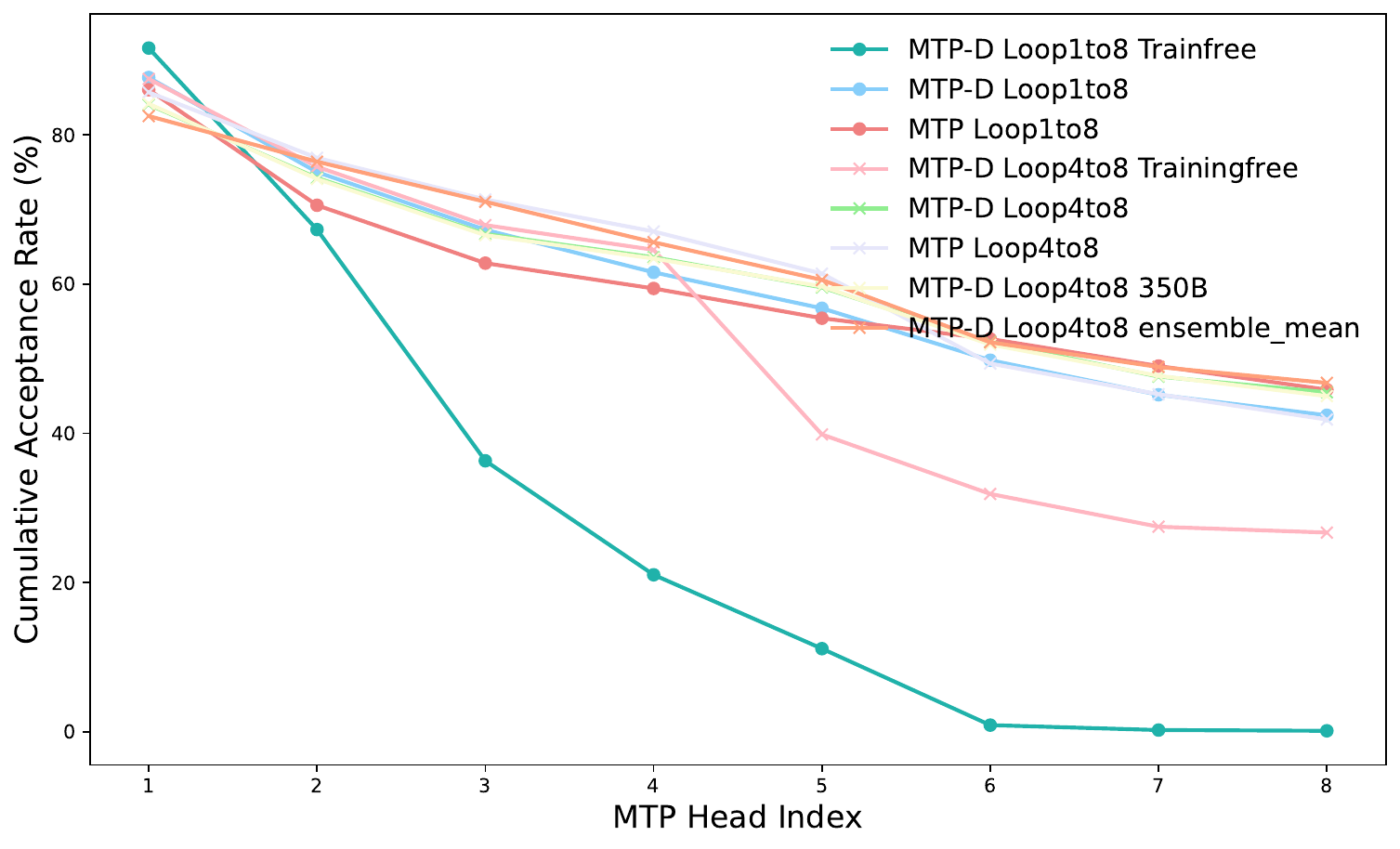}
    \subcaption{SimpleQA.}
\end{subfigure}
\hfill
\begin{subfigure}{0.24\textwidth}
    \centering
    \includegraphics[width=\linewidth]{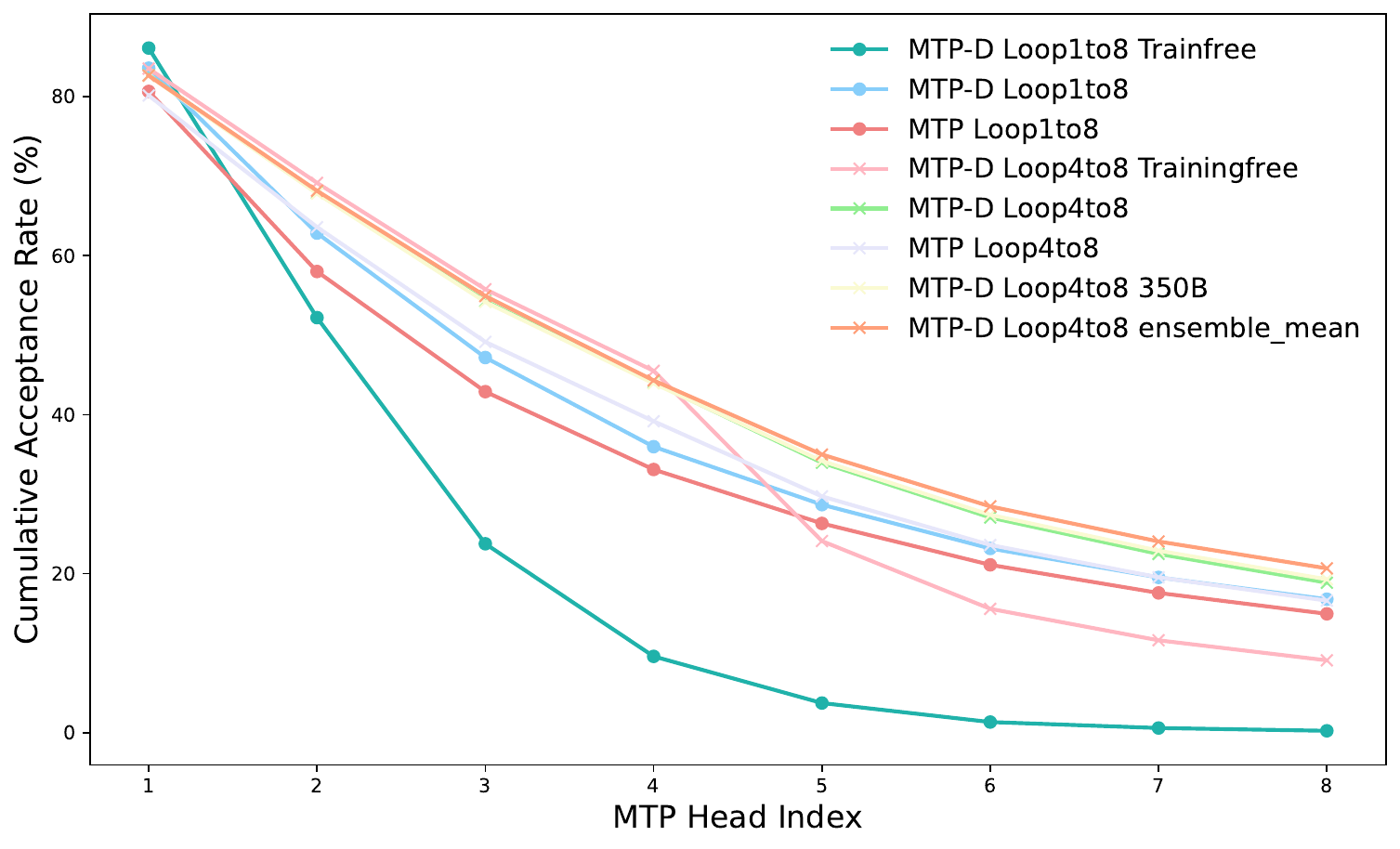}
    \subcaption{SuperGPQA.}
\end{subfigure}
\hfill
\begin{subfigure}{0.24\textwidth}
    \centering
    \includegraphics[width=\linewidth]{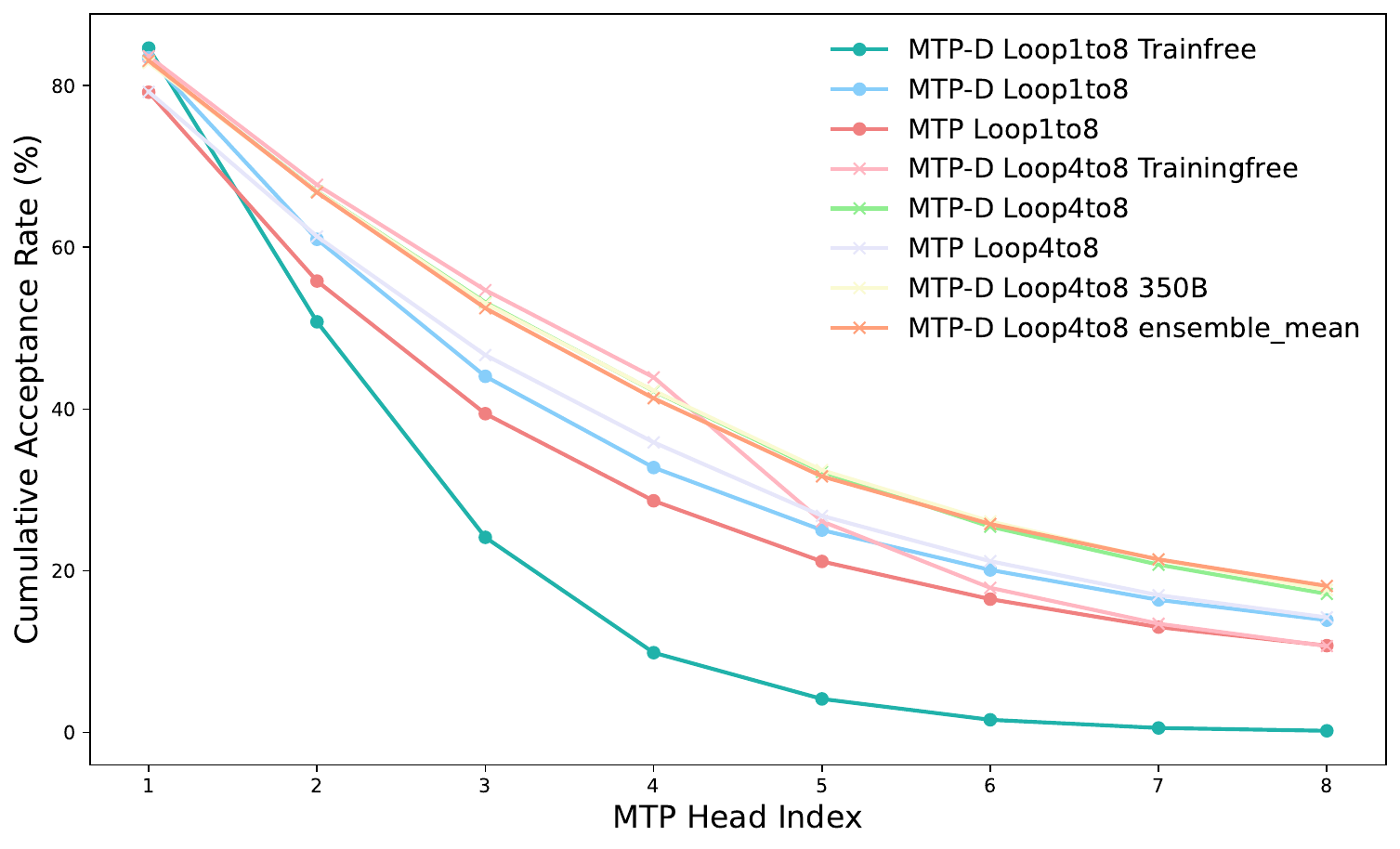}
    \subcaption{TriviaQA.}
\end{subfigure}
\hfill
\caption{Cumulative acceptance rates on multiple pretraining benchmarks with the looped MTP scaled up to 8.}
\label{fig:L-distill_loopto8}
\end{figure*}

\begin{figure*}[h]
\begin{subfigure}{0.24\textwidth}
    \centering
    \includegraphics[width=\linewidth]{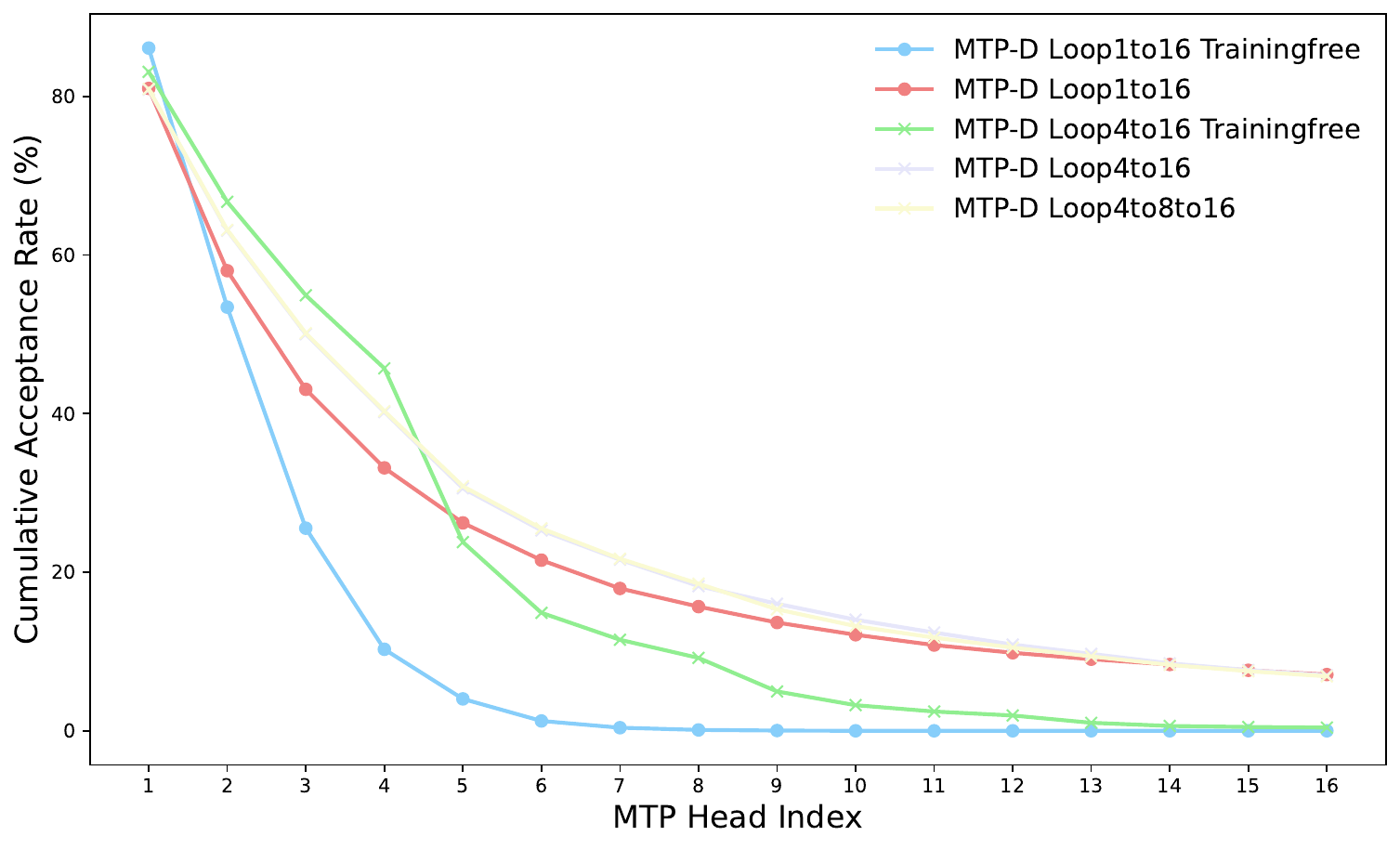}
    \subcaption{AGIEval-en.}
\end{subfigure}
\hfill
\begin{subfigure}{0.24\textwidth}
    \centering
    \includegraphics[width=\linewidth]{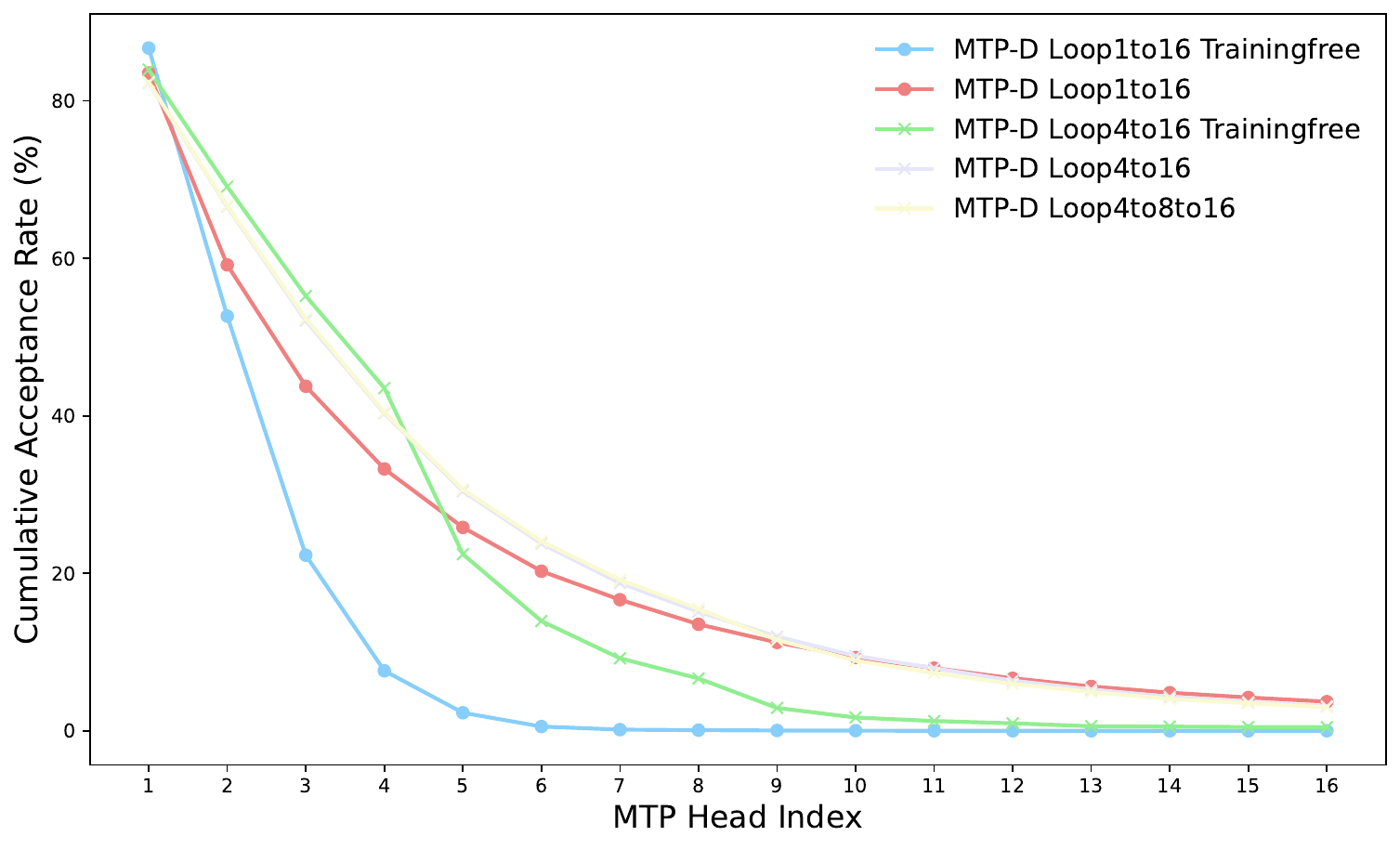}
    \subcaption{GSM8K.}
\end{subfigure}
\hfill
\begin{subfigure}{0.24\textwidth}
    \centering
    \includegraphics[width=\linewidth]{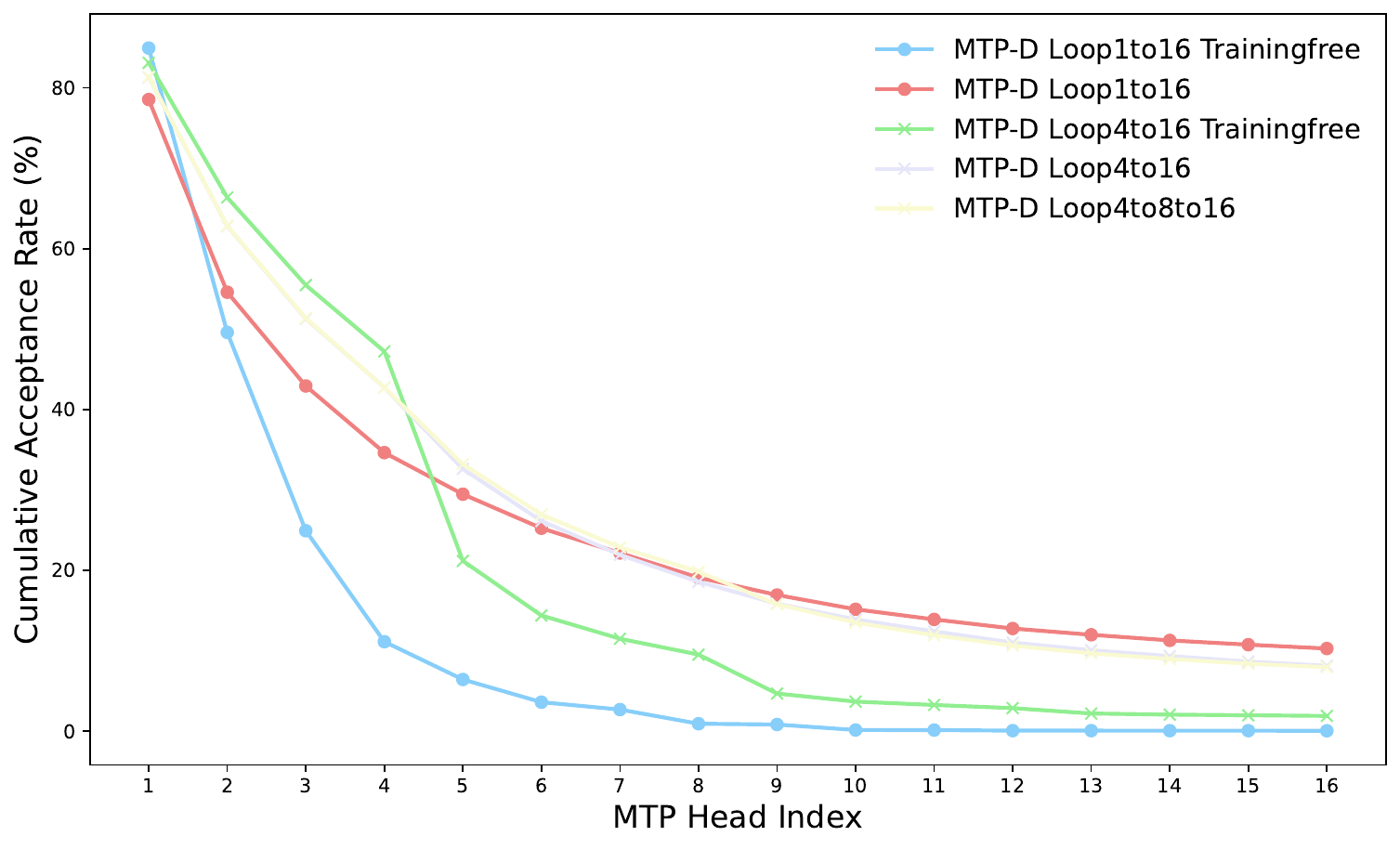}
    \subcaption{MATH.}
\end{subfigure}
\hfill
\begin{subfigure}{0.24\textwidth}
    \centering
    \includegraphics[width=\linewidth]{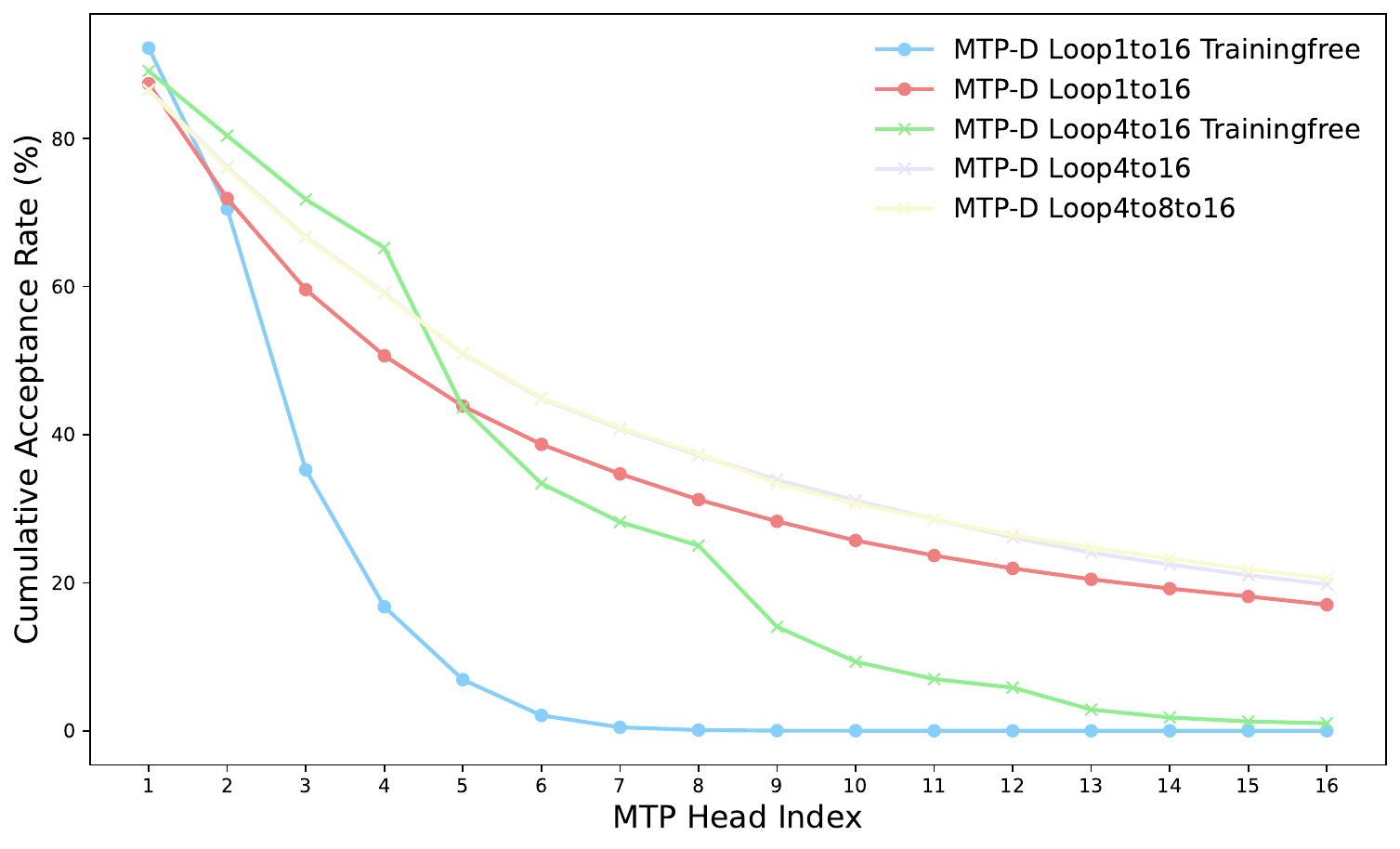}
    \subcaption{NaturalQuestions.}
\end{subfigure}
\hfill
\begin{subfigure}{0.24\textwidth}
    \centering
    \includegraphics[width=\linewidth]{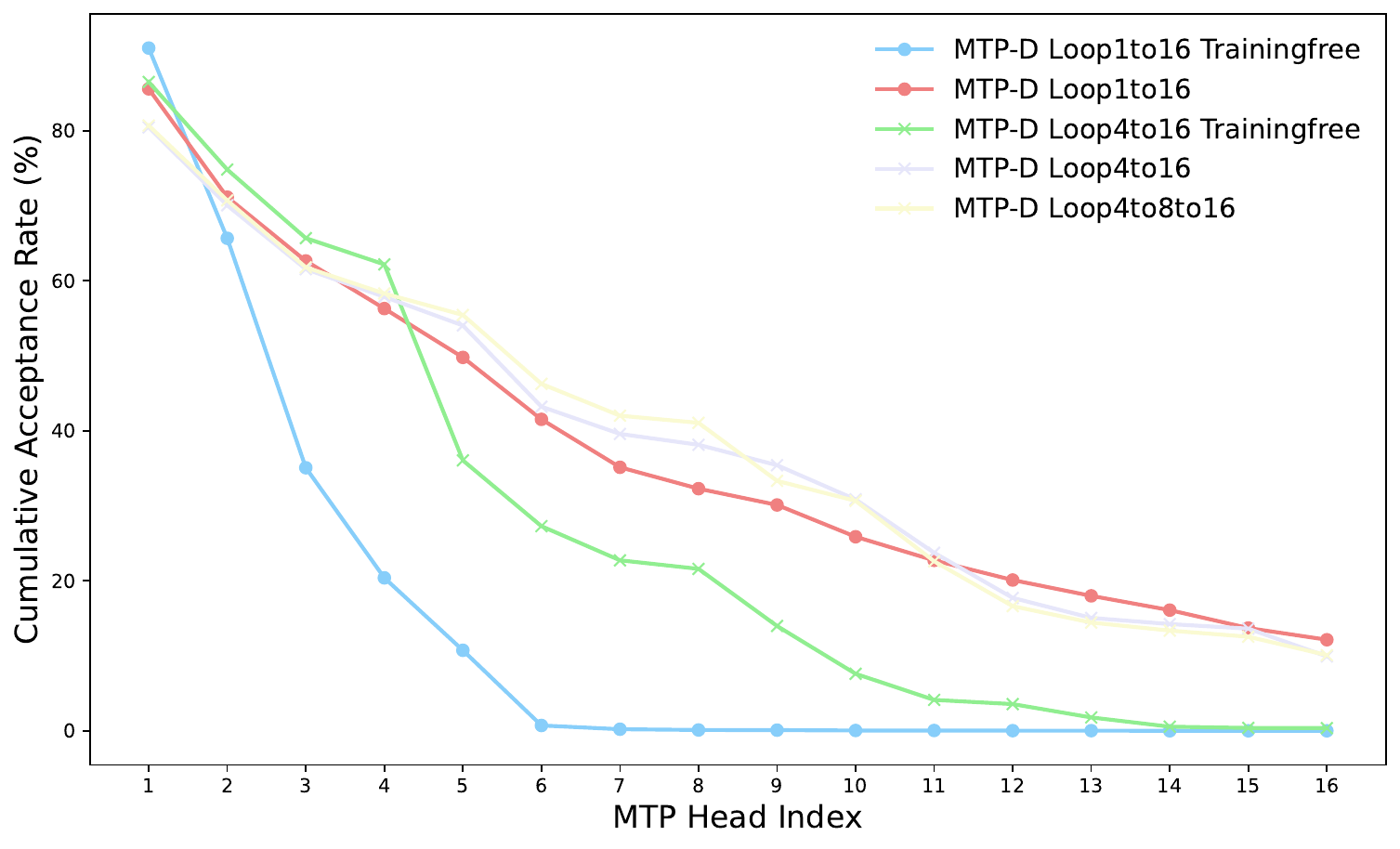}
    \subcaption{SimpleQA.}
\end{subfigure}
\hfill
\begin{subfigure}{0.24\textwidth}
    \centering
    \includegraphics[width=\linewidth]{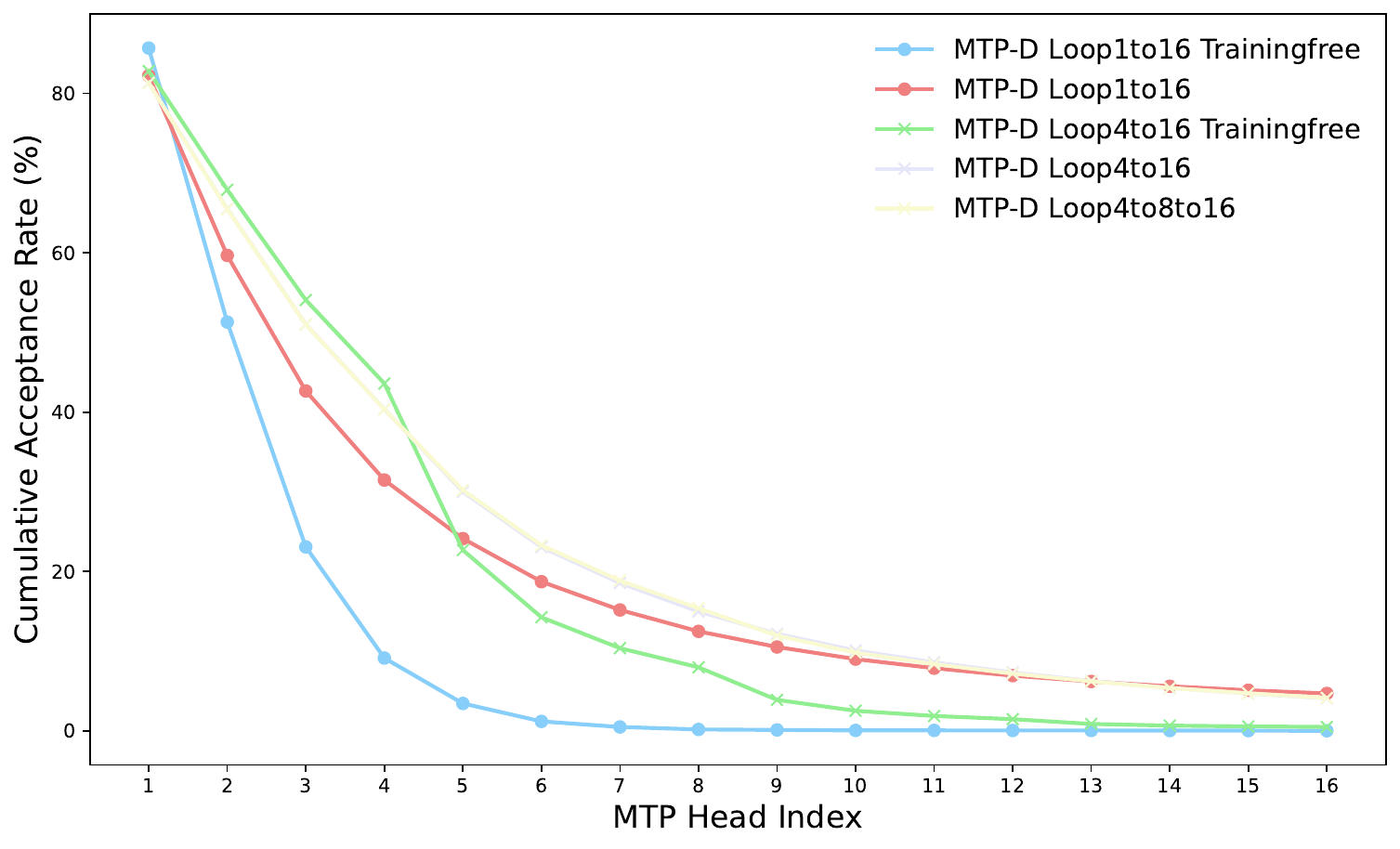}
    \subcaption{SuperGPQA.}
\end{subfigure}
\hfill
\begin{subfigure}{0.24\textwidth}
    \centering
    \includegraphics[width=\linewidth]{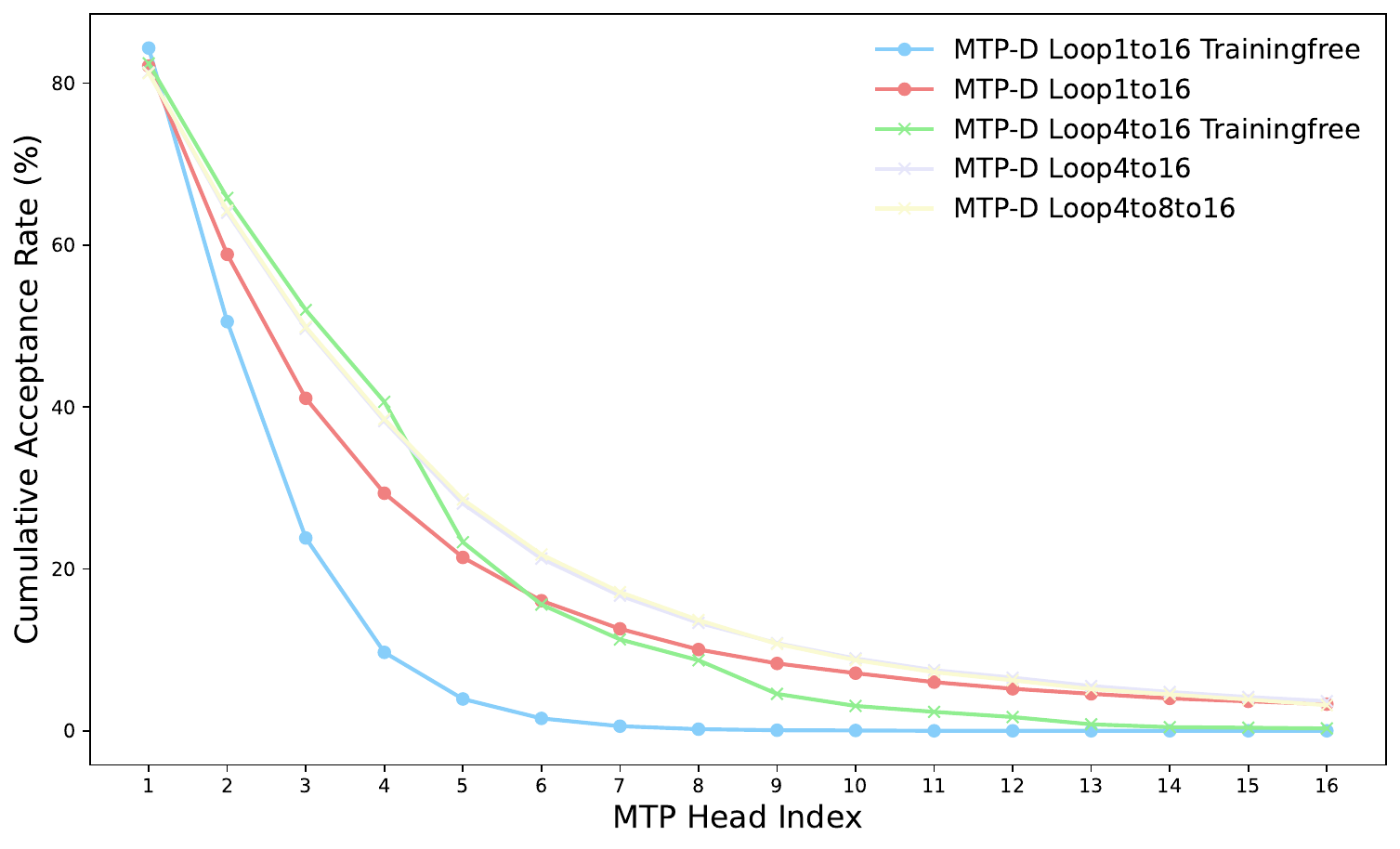}
    \subcaption{TriviaQA.}
\end{subfigure}
\hfill
\caption{Cumulative acceptance rates on multiple pretraining benchmarks with the looped MTP scaled up to 16.}
\label{fig:L-distill_loopto16}
\end{figure*}

Figures~\ref{fig:L-distill_loopto8} and \ref{fig:L-distill_loopto16} present CAR for loops up to 8 and 16, respectively, under continued pre-training. Together with Table~\ref{tab:results_loop_speedup}, these results are used to discuss several insights regarding looped MTP in the main text.

\begin{table*}[h]
\centering
\small
\setlength{\tabcolsep}{2.4pt}
\begin{tabular}{l l c c c c c c c c}
\toprule
& &
\multicolumn{1}{c}{\textbf{General}} &
\multicolumn{2}{c}{\textbf{Math}} &
\multicolumn{3}{c}{\textbf{Knowledge}} &
\multicolumn{1}{c}{\textbf{STEM}} &
 \\
\cmidrule(lr){3-3} 
\cmidrule(lr){4-5} 
\cmidrule(lr){6-8} 
\cmidrule(lr){9-9}
\textbf{Loop} & \textbf{Method} &
\shortstack{\textbf{AGIEval} \\ \textbf{en}} &
\shortstack{\textbf{GSM8K}} & 
\textbf{MATH} &
\shortstack{\textbf{Natural} \\ \textbf{Questions}} &
\shortstack{\textbf{Simple} \\ \textbf{QA}} &
\textbf{TriviaQA} &
\shortstack{\textbf{Super} \\ \textbf{GPQA}} &
\textbf{Mean}
\\
\midrule
\multirow{1}{*}{1} 
    & MTP-D & 1.128 & 1.142 & 1.188 & 1.212 & 1.196 & 1.114 & 1.016 & 1.142 \\
\midrule
\multirow{1}{*}{4} 
    & MTP-D & 1.940 & 2.107 & 2.330 & 2.856 & 2.594 & 2.782 & 1.994 & 2.372 \\
\midrule
\multirow{3}{*}{1 to 8} 
    & MTP-D$^*$ & 1.337 & 1.451 & 1.726 & 2.356 & 1.971 & 2.263 & 1.280 & 1.769 \\
    & MTP & 1.764 & 2.034 & 2.343 & 3.835 & 3.440 & 3.353 & 1.733 & 2.644 \\
    & MTP-D & 1.964 & 2.197 & 2.741 & 4.193 & 3.505 & 3.455 & 1.865 & \textbf{2.846} \\
\midrule
\multirow{5}{*}{4 to 8} 
    & MTP-D$^*$ & 1.851 & 2.058 & 2.358 & 3.552 & 2.885 & 3.192 & 1.884 & 2.540 \\
    & MTP & 1.870 & 2.230 & 2.547 & 3.962 & 3.368 & 3.324 & 1.951 & 2.751 \\
    & MTP-D & 2.071 & 2.365 & 2.800 & 4.573 & 3.591 & 3.985 & 1.980 & 3.052 \\
    & MTP-D 350B & 2.068 & 2.393 & 2.818 & 4.593 & 3.659 & 3.972 & 2.096 & 3.086 \\
    & MTP-D ensemble & 2.132 & 2.414 & 2.972 & 4.433 & 3.678 & 3.959 & 2.026 & \textbf{3.088} \\
\midrule
\multirow{2}{*}{1 to 16} 
    & MTP-D$^*$ & 1.062 & 1.156 & 1.443 & 2.452 & 1.753 & 2.687 & 1.050 & 1.657 \\
    & MTP-D & 1.535 & 1.698 & 2.155 & 3.586 & 2.839 & 3.157 & 1.571 & \textbf{2.363} \\
\midrule
\multirow{2}{*}{4 to 16} 
    & MTP-D$^*$ & 1.510 & 1.713 & 2.095 & 4.142 & 2.806 & 3.741 & 1.496 & 2.500 \\
    & MTP-D & 1.766 & 2.012 & 2.555 & 5.803 & 3.884 & 4.765 & 1.648 & \textbf{3.204} \\
\midrule
4 to 8 to 16 
    & MTP-D & 1.735 & 1.958 & 2.465 & 5.138 & 3.577 & 4.323 & 1.676 & 2.982 \\
\bottomrule
\end{tabular}
\caption{Speedup ratios of different loop strategies for MTP-D and MTP during continued pre-training. Except for MTP-D with 350B data sizes, all other continued pre-training experiments use a data size of 70B tokens. \textasteriskcentered{} indicates Training Free.}
\label{tab:results_loop_speedup}
\end{table*}
\begin{algorithm}[h]
\small
\caption{Main-Head-constrained Speculative Decoding}
\label{alg:mtp_spec_refined}
\SetAlgoLined
\DontPrintSemicolon
\SetKwInput{Input}{Input}
\SetKwInput{Output}{Output}
\SetKwFunction{Sample}{Sample}

\Input{Initial prompt sequence $\mathbf{x}$, Main model $f$, MTP heads $\{g_k\}_{k=1}^K$, Max new tokens $N$}
\Output{Generated sequence $\mathbf{x}$}

\BlankLine
$L_{prompt} \leftarrow |\mathbf{x}|$ \quad $L_{gen} \leftarrow 0$\;

$C_{\text{step}} \leftarrow 0$\;

\For{$j = 1$ \KwTo $K$}{
    $A_j \leftarrow 0$\;
    $C^{cmp}_j \leftarrow 0$\;
}

\BlankLine
\While{$L_{gen} < N$}{
    \tcp{Verification and Correction}
    $(\mathbf{P}, \mathbf{h}) \leftarrow f(\mathbf{x})$\;
    $is\_verified \leftarrow \text{True}$\;
    
    \If{$L_{gen} > 0$}{
        $C_{\text{step}} \leftarrow C_{\text{step}} + 1$\;
        $m \leftarrow 0$\;
        
        \For{$j = 1$ \KwTo $K$}{
            $i \leftarrow K - j + 1$\;
            $C^{cmp}_j \leftarrow C^{cmp}_j + 1$\;
            
            $\hat{x} \leftarrow \Sample(P_{|\mathbf{x}|-i})$\;
            
            \If{$\hat{x} \neq x_{|\mathbf{x}|-i+1}$}{
                $\mathbf{x} \leftarrow \mathbf{x}_{1:|\mathbf{x}|-i} \mathbin{\Vert} (\hat{x})$\;
                
                $\mathbf{h} \leftarrow \mathbf{h}_{1:|\mathbf{x}|-i}$\;
                
                $is\_verified \leftarrow 
                \text{False}$\;
                
                \textbf{break}\;
            }
            \Else{
                $m \leftarrow m + 1$\;
            }
        }
        
        \If{$m > 0$}{
            \For{$t = 1$ \KwTo $m$}{
                $A_t \leftarrow A_t + 1$\;
            }
        }
    }
    
    \If{$\text{EOS} \in \mathbf{x}$}{\textbf{break}}

    \BlankLine
    \tcp{Speculative Expansion}
    \If{$is\_verified$}{
        $x_{|\mathbf{x}|+1} \leftarrow \Sample(P_{|\mathbf{x}|})$\;
        
        $\mathbf{x} \leftarrow \mathbf{x} \mathbin{\Vert} (x_{|\mathbf{x}|+1})$\;
    }
    
    \For{$k = 1$ \KwTo $K$}{
        $(\hat{P}_{MTP}, \mathbf{h}) \leftarrow g_k(\mathbf{x}, \mathbf{h})$\;
        
        $x_{|\mathbf{x}|+1} \leftarrow \Sample(\hat{P}_{MTP})$\;
        
        $\mathbf{x} \leftarrow \mathbf{x} \mathbin{\Vert} (x_{|\mathbf{x}|+1})$\;
    }
    
    $L_{gen} \leftarrow |\mathbf{x}| - L_{prompt}$\;
}

\Return $\mathbf{x}$
\end{algorithm}
\end{document}